\documentclass{article}

\usepackage[preprint]{neurips_2022}

\usepackage[utf8]{inputenc} % allow utf-8 input
\usepackage[T1]{fontenc}    % use 8-bit T1 fonts
\usepackage[colorlinks=true, linkcolor=blue, citecolor=blue]{hyperref}
\usepackage{url}            % simple URL typesetting
\usepackage{booktabs}       % professional-quality tables
\usepackage{amsfonts}       % blackboard math symbols
\usepackage{nicefrac}       % compact symbols for 1/2, etc.
\usepackage{microtype}      % microtypography
\usepackage{enumerate}
\usepackage{comment}
\usepackage{enumitem}

\usepackage{graphicx}
\usepackage{subcaption}
\usepackage{epstopdf}
\bibpunct{(}{)}{;}{a}{,}{,}

\usepackage{xcolor}         % colors
\usepackage{algorithm, algorithmic}

\usepackage{framed}
\usepackage{amsmath}
\usepackage{mathtools}
\newcommand{\E}{\mathbb E}
\newcommand{\EE}{\mathbb E}

\newcommand{\RR}{\mathbb R}
\newcommand{\PP}{\mathbb{P}}

\newcommand{\cN}{\mathcal{N}}
\newcommand{\cD}{\mathcal{D}}
\newcommand{\lv}{\left \langle}
\newcommand{\rv}{\right \rangle}
\newcommand{\cO}{\mathcal{O}}
\newcommand{\cS}{\mathcal{S}}
\newcommand{\cX}{\mathcal{X}}
\newcommand{\cK}{\mathcal{K}}
\newcommand{\cR}{\mathcal{R}}
\newcommand{\1}{\textbf{1}}

\def\sg#1{\textcolor{blue}{[SG: #1]}}
\def\mw#1{\textcolor{red}{[MW: #1]}}
\def\xz#1{\textcolor{purple}{[XZ: #1]}}

\DeclareMathOperator*{\argmin}{\mathrm{argmin}}

\newtheorem{theorem}{Theorem}[section]

\newtheorem{lemma}[theorem]{Lemma}

\newtheorem{assumption}[theorem]{Assumption}

\newcommand{\QED}{$\hfill \square$}

\title{Decentralized Gossip-Based Stochastic Bilevel Optimization 
over Communication Networks
}

\author{
Shuoguang Yang
\\
IEDA, HKUST
\\
\texttt{yangsg@ust.hk}
\And
Xuezhou Zhang\\
Princeton University\\
\texttt{xz7392@princeton.edu}
\And
Mengdi Wang\\
Princeton University\\
\texttt{mengdiw@princeton.edu}
}

\begin{document}

\maketitle

\begin{abstract}
Bilevel optimization have gained growing interests, with numerous applications found in meta learning, minimax games, reinforcement learning, and nested composition optimization. 
This paper studies the problem of distributed bilevel optimization over a network where agents can only communicate with neighbors, including examples from multi-task, multi-agent learning and federated learning.
In this paper, we propose a gossip-based distributed bilevel learning algorithm that allows networked agents to solve both the inner and outer optimization problems in a single timescale and share information via network propagation. We show that our algorithm enjoys the $\cO(\frac{1}{K \epsilon^2})$ per-agent sample complexity for general nonconvex bilevel optimization and $\cO(\frac{1}{K \epsilon})$ for strongly convex objective, achieving a speedup that scales linearly with the network size. The sample complexities are optimal in both $\epsilon$ and $K$.
We test our algorithm on the examples of hyperparameter tuning and decentralized reinforcement learning. Simulated experiments confirmed that our algorithm achieves the state-of-the-art training efficiency and test accuracy.
% \xz{If we decide to down-weight the strongly convex setting we should also remove it from abstract.}
\end{abstract}

\section{Introduction}\label{sec:intro}
In recent years, stochastic bilevel optimization (SBO) has attracted increasing attention from the machine learning community. It has been found to provide favorable solutions in a variety of problems, such as meta learning and hyperparameter optimization \citep{franceschi2018bilevel,snell2017prototypical, bertinetto2018meta}, composition optimization \citep{wang2016stochastic}, two-player games \citep{von1952theory}, reinforcement learning and imitation learning \citep{arora2020provable, hong2020two}. While the majority of the above mentioned work focuses on algorithm designs in the classic centralized setting, such problems often arise in distributed/federated applications, where agents are not willing to share data but rather perform local updates and communicate with neighbors. Theories and algorithms for distributed stochastic bilevel optimization are less developed.

 Consider the \emph{decentralized} learning setting where the data are distributed over $K$ agents $\cK= \{ 1,2,\cdots, K\}$ over a communication network. Each agent can only communicate with its neighbors. One example is federated learning which often concerns with a single-server-multi-user system, where agents communicate with a central server to cooperatively solve a task \citep{lan2018random},\citep{ge2018minimax}. Another example is sensor network, where sensors are fully decentralized and can only communicate with nearby neighbors \citep{distributed_sensor} .
% Such systems possess advantages in privacy, security and robustness.  \mw{cite}. 
% %The computation/communication efficiency of the central server may become a bottleneck and slow down the convergence \citep{lian2018asynchronous}. 
% %Decentralized systems avoid such bottleneck and exhibit robustness. 
% \mw{do not compare decentralized against federated. they are just two settings and there is no good or bad}

\begin{figure}[t]
    \centering
      \vskip -1.5cm
        \includegraphics[width=0.35\linewidth]{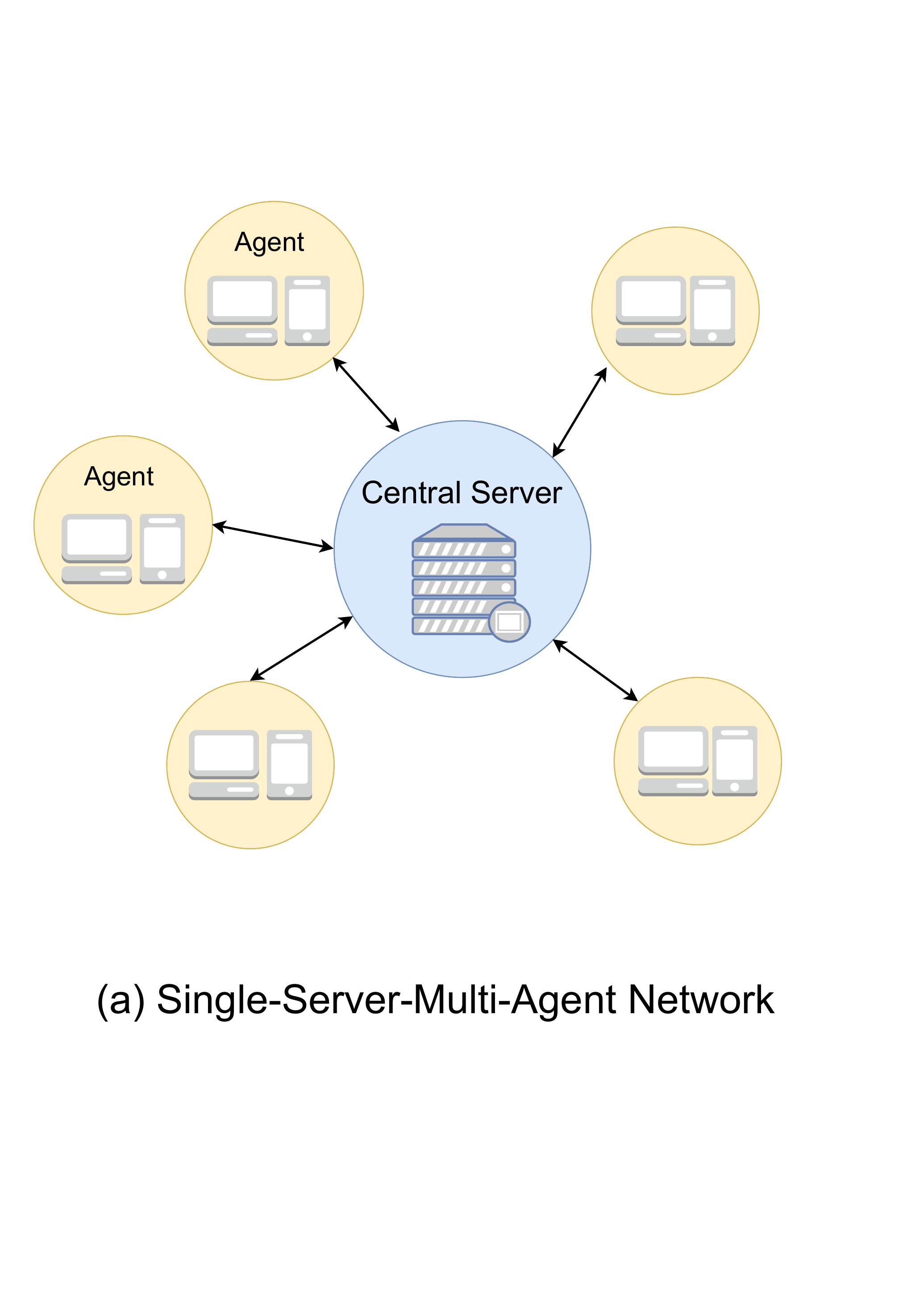} 
        \qquad\quad
  \includegraphics[width=0.35\linewidth]{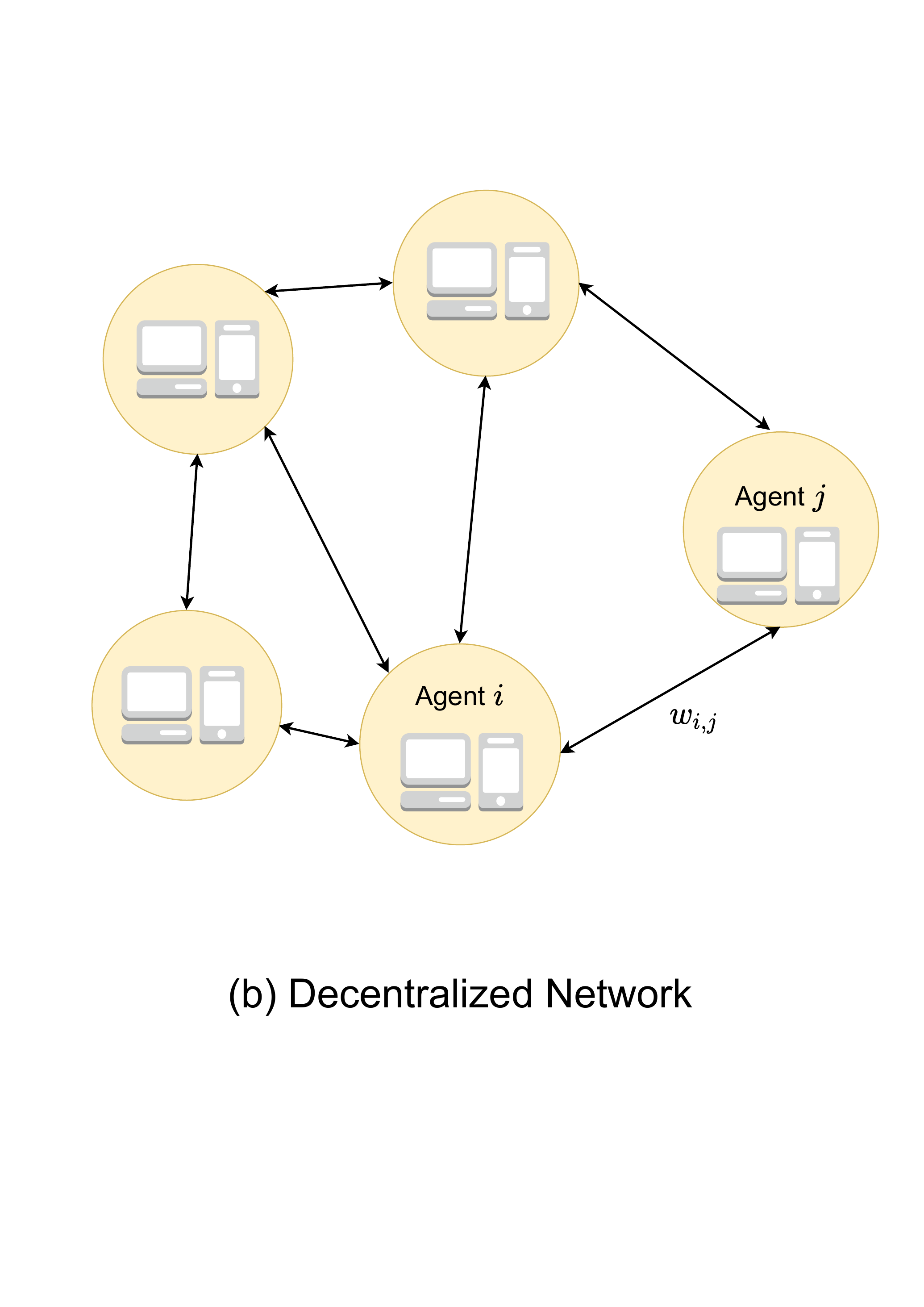}
  \vskip -1.5cm
   \caption{Illustration of Distributed Network Structures. In a centralized network, the agents may work cooperatively to solve a specific task by communicating with the central server, commonly seen in federated learning.  }
   %In federated learning, the agents may have  heterogeneous data distribution and or with its neighbors in centralized or decentralized network, respectively. 
%   \mw{Not precise. federated learning is only a}}
    \label{fig:network}
\end{figure}

% \xz{Write a sentence of why the network structure is a good assumption for applications. Maybe give an example where such a network emerges.}
We consider the following decentralized stochastic bilevel optimization problem, 
\begin{equation}\label{prob:1}
\min_{ x\in \RR^{d_x}}   F(x) =  \left \{ \frac{1}{K}\sum_{k=1}^K f^k (x, y^*(x) )  \right \}, \text{ s.t. }y^*(x) = \argmin_{y \in \RR^{d_y} } \left \{  \frac{1}{K}\sum_{k=1}^K g^k \big (x, y )  \right \}, 
\end{equation}
where $f^k(x,y) = \E_{\zeta^k} [f^k (x, y ;\zeta^k )  ]$
 and $g^k(x,y) =  \EE_{\xi^k}[g^k(x,y; \xi^k)]$ may vary across agents. The expectations $\EE_{\zeta^k}[\bullet]$  and $\EE_{\xi^k}[\bullet]$ are taken with respect to the random variables $\zeta^k$ and $\xi^k$, with heterogeneous distributions across agents. We consider the scenario where each $g^k(x,y)$ is strongly convex in $y$ and denote by $f(x,y) =\frac{1}{K} \sum_{k\in \cK} f^k(x,y)$ and $g(x,y) = \frac{1}{K} \sum_{k\in \cK}  g^k(x,y)$ for notational convenience. 
\subsection{Example applications of SBO}

SBO was first employed to formulate the resource allocation problem~\citep{bracken1973mathematical} and has since found its application in many classic operations research settings \citep{cramer1994problem, sobieszczanski1997multidisciplinary, livne1999integrated, tu2020two, tu2021two}, and more recently in machine learning problems \citep{franceschi2018bilevel,snell2017prototypical, bertinetto2018meta,wang2016stochastic}. In particular, we introduce two applications that have received a lot of recent attention, namely hyperparameter optimization and compositional optimization.

\paragraph{Hyperparameter Optimization}
The problem of hyper-parameter tuning \citep{okuno2021lp} often takes the following form:
 \begin{equation}\label{prob:hyperopt}
  \min_{x \in \RR^d} \left \{   \sum_{i \in \cD_{\text{val}}}  \ell_i (y^*(x)) \right \}, \mbox{ s.t. } y^*(x) \in \argmin_{y \in \RR^{d_y}} \left \{  \sum_{j \in  \cD_{\text{train}}} \ell_j(y) + \cR(x,y) \right \},
 \end{equation}
 where $\cD_{\text{train}}$ and $\cD_{\text{val}}$ are two datasets used for training and validation, respectively, $y \in \RR^{d_y}$  is a vector of unknown parameters to optimize, $\ell_j(y)$ is a convex loss over data $i$, and $x \in \RR^d$ is a vector of hyper-parameters for a strongly convex regularizer $\cR(x,y)$. For any hyper-parameter $x \in \RR^d$, the inner-level problem solves for the best parameter $y^*(x)$ over the training set $\cD_{\text{train}}$ under the regularized training loss $\ell_j(y)+\cR(x,y)$.
 The goal is to find the hyper-parameter $x^* \in \RR^d$ whose corresponding best response $y^*(x)$ yields the least loss over the validation set $\cD_{\text{val}}$. 
  In practice, tuning continuous hyperparameters is often done by grid search which is exponentially expensive. Efficient algorithm of SBO should find optimal parameters in time polynomial to dimension, rather than exponential. When the training and validation set are distributed across nodes, the problem becomes a distributed SBO.
 
\paragraph{Compositional Optimization} Let $g(x,\xi) : \RR^d \to \RR^{d_y}$ and $f(y,\zeta): \RR^{d_y} \to \RR$  be two stochastic mappings, stochastic compositional optimization (SCO) \citep{wang2017stochastic} takes the form
\begin{equation*}
\min_{ x \in \cX} \EE_\zeta[ f ( \EE_\xi [ h(x; \xi)]; \zeta)].
\end{equation*}
SCO applies to risk management \citep{yang2019multilevel,ruszczynski2021stochastic}, machine learning \citep{chen2021solving}, and reinforcement learning \citep{wang2017accelerating}. 
SCO was identified as a special case of SBO by \citep{chen2021tighter}. To see this, we take the inner optimization objective to be
 $g(x,y) = \EE_\xi [(y - h(x,\xi))^2] $. Thus $y^*(x)  = \argmin_y g(x,y)$ becomes  $y^*(x) = \EE_\xi[ h(x;\xi)]$, arriving at an instance of SBO.

\subsection{Challenges with Distributed SBO}\label{sec:challenges}

Despite the recent fast development of distributed single-level optimization, a proper method for Decentralized Stochastic Bilevel Optimization (DSBO) remains illusive. The major hurdle of solving DSBO lies in the absence of explicit knowledge of $y^*(x)$, so that an unbiased gradient of $\nabla f(x, y^*(x))$ is unavailable. Recently, by applying the implicit function theorem, \citep{couellan2016convergence,ghadimi2018approximation} showed that the gradient of non-distributed SBO can be expressed as 
 \begin{equation}\label{eq:gradient_BO}
 \nabla_x f(x,y^*(x)) - \nabla_{xy}^2 g(x,y^*(x)) [\nabla_{yy}^2 g(x,y^*(x)) ]^{-1}\nabla_y f(x,y^*(x)),
\end{equation}
providing a connection between SBO and classical stochastic optimization. 
Since then, various algorithms have been proposed to obtain sharp estimators for $y^*(x)$ and reduce the bias of the constructed gradients \citep{chen2021single,hong2020two,ji2021bilevel,yang2021provably}. These techniques result in nearly tight convergence analysis and give rise to algorithms widely used in applications. However, none of the prior algorithms can be applied to the distributed setting.
% \mw{specify algorithmic challenges}
% \mw{compound effect of consensus and SA}

%Here we brief explain the challenges within solving Prob.~\eqref{prob:1}. 
Solving Problem \eqref{prob:1} becomes challenging in the distributed setting in several aspects:
\begin{enumerate}[leftmargin=*,itemsep=0pt]
    \item Even in single-agent SBO, lack of $y^*(x)$ makes outer optimization nontrivial, and estimating $y^*(x)$ requires additional stochastic approximation, weighted averaging, and sophisticated calculation.
    \item Calculating outer gradient is highly nontrivial, even when we have an inner solution $y$. Note that 
\begin{eqnarray}\label{eq:biased_grad}
 \frac{1}{K} \sum_{k \in \cK} \left ( \nabla_{xy}^2 g^k(x,y) [ \nabla_{yy}^2 g^k(x,y) ]^{-1} \nabla_y f^k(x,y)  \right )  \neq \nabla_{xy}^2 g(x,y) [\nabla_{yy}^2 g(x,y) ]^{-1}\nabla_y f(x,y).
\end{eqnarray}
In other words, even the inner problem is solved, the outer gradient requires new mechanism to estimate.
\item Now we move to SBO over distributed networks. In network learning, communication among agents can be limited by the network stucture and communication protocol, so taking a simple average across agents may require multiple rounds of communication. 
%As a result, known methods for SBO or compositional optimization \citep{chen2021tighter,chen2021solving} cannot be applied to the distributed network setting.
\end{enumerate}

% When SBO meets distributed network optimization, new challenges arise. Most distributed optimization methods rely on the assumption that agents can access unbiased gradients. However, neither gradient estimate or $y^*(x)$ is accessible to local agents. Furthermore, communication among agents can be limited by the network topology, so taking a simple average across agents may require multiple rounds of communication. 

Because of the above hurdles, it remains unclear how to sharply estimate the outer gradient under a distributed network. 
As an attempt to tackle this problem, this paper study the convergence theory and sample complexity of gossip-based algorithm. In particular, we ask two theoretical questions: \begin{center}
\emph{
    (i) How does the sample complexity of DSBO scale with the optimality gap and network size?\\
    (ii) How is the efficiency of DSBO affected by the network structure?
}
\end{center}

In this paper, we develop a gossip-based stochastic approximation scheme where each agent  solves an optimization problem collaboratively by sampling stochastic first- and second-order information using its data and making gossip communications with its neighbors. 
In addition, we develop novel convergence analysis techniques to characterize the convergence behavior of our algorithm.
To the best of our knowledge, our work is the first to  formulate DSBO mathematically and propose an algorithm with theoretical convergence guarantees. Moreover, we show that our algorithm enjoys an $\cO(\frac{1}{K \epsilon^2})$ sample complexity for finding $\epsilon$-stationary points for nonconvex objectives and enjoys an $\cO(\frac{1}{K \epsilon})$ sample complexity for $\mu$-PL functions, subsuming strongly convex optimization. 
These results subsume the state-of-the-art results for non-federated  stochastic bilevel optimization \citep{chen2021tighter} and central-server regime \citep{tarzanagh2022fednest}, showing that nearly no degradation is induced by network consensus. 
%\xz{We should provide a formal definition of "Consensus" somewhere in the text. I know roughly what it is but am not crystal clear..} 
Further, the above results suggest that our algorithm exhibits the linear speed-up effect for decentralized settings; that is, the required per-agent sample complexities decrease linearly to the number of agents.

\section{Related Works}
Bilevel optimization was first formulated by~\citep{bracken1973mathematical} for solving resource allocation   problems. Later, a class of constrained-based algorithms was proposed by \citep{hansen1992new,shi2005extended},  which treats the inner-level optimality condition as constraints to the out-level problem. 
Recently,  \citep{couellan2016convergence} examined the finite-sum case for unconstrained strongly convex lower-lower problems and proposed a gradient-based algorithm that exhibits asymptotic convergence under certain step-sizes. 
For SBO, \citep{ghadimi2018approximation} developed a double-loop algorithm and established the first known complexity results. Subsequently, various methods  have been employed to improve the sample complexity, including two-timescale stochastic approximation~\citep{hong2020two}, acceleration  \citep{chen2021single}, momentum~\citep{khanduri2021near},  and variance reduction
\citep{guo2021randomized,ji2021bilevel,yang2021provably}.
%From the optimization perspective, it is worth emphasizing that bilevel optimization can be treated as a special case of mathematical programs with equilibrium constraints \citep{luo96mathematical}.

%  The two-level SCO was originally proposed by \citep{Erm76}. Under a weighted-average stochastic approximation scheme, Wang et al. \citep{wang2017stochastic} developed a two-timescale algorithm that achieves the first known $\cO(1/\epsilon^{4})$ and $\cO(1/\epsilon^{1.5})$ sample complexities for nonconvex and strongly convex objectives, respectively. These convergence rates were further improved to $\cO(1/\epsilon^{9/4})$ and $\cO(1/\epsilon^{5/4})$ by using acceleration \citep{wang2017accelerating}. The generalization to  multilevel compositional setting was then established by \citep{yang2019multilevel}. Later, \citep{ghadimi2020single} developed a single-timescale algorithm that achieves the first provable $\cO(1/\epsilon^2)$ sample complexity for nonconvex two-level problems, while its multi-level generalization was further developed by \citep{balasubramanian2020stochastic,ruszczynski2021stochastic}. Extensive attempts have also been made to improve the convergence rates, such as variance-reduction \citep{zhang2021multilevel} and  linearization \citep{chen2021solving}. 
%The finite-sum regime has also been considered by finite-sum \citep{huo2018accelerated,lian2017finite,lin2020improved}, while \citep{zhang2020optimal} attacked the convex regime and obtain the optimal sample complexity by using Fenchel conjugates. 

Distributed optimization was developed to  handle real-world large-scale datasets~\citep{dekel2012optimal,feyzmahdavian2016asynchronous} and graph estimation \citep{wang2015distributed}. Centralized and decentralized systems are two important problems that have drawn significant attention. A centralized system considers the network topology where there exists a central agent that communicates with the remaining agents \citep{lan2018random}, while in a decentralized system \citep{gao2020periodic,koloskova2019decentralized,lan2020communication,mcmahan2017communication}, each agent can only communicate with its neighbors by using gossip \citep{lian2017can} or  gradient tracking \citep{pu2021distributed} communication strategies. Variance reduction approaches \citep{xin2020variance,xin2021improved,pmlr-v54-lian17a} have also been applied to improve the convergence rate of decentralized optimization. Random projection schemes have been studied to handle large sets of constraints \citep{wang2016stochastica,7178639}.
All of the above trials were made on vanilla stochastic optimization problems.
We are the first to study the decentralized bilevel optimization problem to our knowledge. 
% \xz{optional: instead of this line, we could add a paragraph discussing the two concurrent work and how our work subsumes both of them. I think it will strengthen the paper.}
Table \ref{table:1} compares this work with prior arts under different settings. 
 \begin{table} 
\begin{center}
\begin{tabular}{ |c|c|c| c  | c |  c |  c | } 
 \hline
\multicolumn{7}{|c|}{  Stochastic Bilevel Optimization }
 \\
 \hline 
 & BSA  &   stocBio & ALSET &  FEDNEST &  \multicolumn{2}{|c|}{ This Work} \\  \hline
%Agents & \multicolumn{3}{c}{Single-Agent} & \multicolumn{3}{|c|}{Multi-Agent}  \\ \hline 
Network & \multicolumn{3}{c}{Single-Agent}  & \multicolumn{2}{|c|}{ Central-Server} & Decentralized \\ \hline 
 Samples in $\zeta$  & $\cO(\epsilon^{-2} )$  &  $\cO(\epsilon^{-2})$  & $\cO(\epsilon^{-2})$ &  $\cO(\epsilon^{-2})$ &   $\cO(\tfrac{1}{K\epsilon^2} ) $  & $\cO(\tfrac{1}{K\epsilon^2} + \tfrac{K}{(1- \rho)^2\epsilon})$    \\   \hline 
  Samples in $\xi$  & $\cO(\epsilon^{-3} )$ &  $\cO(\epsilon^{-2})$  &  $\cO(\epsilon^{-2})$ &  $\cO(\epsilon^{-2})$ & $\cO(\tfrac{1}{K\epsilon^2} ) $ & $\cO(\tfrac{1}{K\epsilon^2} + \tfrac{K}{(1- \rho)^2\epsilon})$  \\   \hline 
 \end{tabular}
 \vskip 0.5cm
 \begin{tabular}{ |c|c|c| c | c | c |  c | } 
 \hline
\multicolumn{7}{|c|}{  Stochastic Compositional Optimization}
 \\
 \hline 
 & SCGD  &  NASA & ALSET & FEDNEST  &  \multicolumn{2}{|c|}{ This Work}   \\  \hline
%Agents & \multicolumn{3}{c}{Single-Agent} & \multicolumn{3}{|c|}{Multi-Agent}  \\ \hline 
Network & \multicolumn{3}{c}{Single-Agent}   & \multicolumn{2}{|c|}{ Central-Server} & Decentralized \\ \hline 
 Samples in $\zeta$  & $\cO(\epsilon^{-4} )$ & $\cO(\epsilon^{-2} )$ &  $\cO(\epsilon^{-2})$  &   $\cO(\epsilon^{-2})$  & $\cO( \frac{1}{K\epsilon^2})$ & $\cO(\tfrac{1}{K\epsilon^2} + \tfrac{K}{(1- \rho)^2 \epsilon})$   \\   \hline 
  Samples in $\xi$  & $\cO(\epsilon^{-4} )$   & $\cO(\epsilon^{-2})$  & $\cO(\epsilon^{-2})$  &  $\cO(\epsilon^{-2})$  &  $\cO( \frac{1}{K\epsilon^2})$ & $\cO(\tfrac{1}{K\epsilon^2} + \tfrac{K}{(1- \rho)^2 \epsilon})$   \\   \hline 
\end{tabular}

\end{center}
 \caption{Summary of per-agent sample complexities for nonconvex stochastic bilevel and compositional optimization. $(1-\rho)$ denotes the spectral gap of the network adjacency matrix. BSA \citep{ghadimi2018approximation}, ALSET \citep{chen2021single}, stocBio \citep{ji2021bilevel}, FEDNEST \citep{tarzanagh2022fednest}, SCGD \citep{wang2016stochastic}, NASA \citep{ghadimi2020single}. }
 \label{table:1}
\end{table}

\section{Problem Setup}

%We start by formally specifying the sampling environment that generates the first- and second-order stochastic information.  In this paper, we assume access to the following black-box Sampling Oracle ($\cS\cO$):

\begin{assumption}[Sampling Oracle $\mathcal{SO}$]\label{assumption:SO}
Agent $k$ may query the sampler and receive an independent locally sampled unbiased first- and second-order information $\nabla_x f^k(x,y; \zeta^k)$, $\nabla_y f^k(x,y; \xi^k)$, $\nabla_y g^k(x,y;\xi^k)$, $\nabla_{yy}^2 g^k(x,y;\xi^k) $, and  $ \nabla_{xy}^2 g^k(x,y; \xi^k)$.
%\sg{shorten, combine with bounded second moments Ass. 5.1} 
\begin{comment}
\begin{enumerate}
\item  local inner gradient sample $\nabla_y g^k(x,y ;\xi^k) $ such that 
$
\EE_{\xi^k}[ \nabla_y g^k( x,y  ;\xi^k) ] = \nabla_y g^k( x,y ). 

\item local inner Hessian n $\nabla_{xy}^2  g^k(x,y;\xi^k) \in \RR^{d \times d_y}$ and $\nabla_{yy}^2  g^k(x,y;\xi^k) \in \RR^{d_y \times d_y}$, only observable to agent $k$, such that 
$$
\EE_{\xi^k}[ \nabla_{xy}^2  g^k( x,y  ;\xi^k)] =  \nabla_{xy}^2 g^k( x,y  ), \text{ and }\EE_{\xi^k}[ \nabla_{yy}^2  g^k( x,y ;\xi^k)] =  \nabla_{yy}^2 g^k( x,y ).
$$
\item the $\cS\cO$ independently returns  unbiased stochastic first-order information $\nabla_x  f^k(x,y; \zeta^k ) \in \RR^{d}$ and $\nabla_y  f^k( x,y ; \zeta^k ) \in \RR^{d_y}$, only observable to agent $k$, such that 
$$
\EE_{\zeta^k} [ \nabla_x  f^k( x,y  ; \zeta^k )] =  \nabla_x f^k( x,y  ), \text{ and }\EE_{\zeta^k} [ \nabla_y  f^k( x,y  ; \zeta^k )] =  \nabla_y f^k( x,y ).
$$
\end{enumerate}
\end{comment}
\end{assumption}

\begin{assumption}[Gossip Protocol] \label{assumption:W}
The network gossip protocol is specified by a $K\times K$ symmetric matrix $W$ with nonnegative entries. Each agent $k$ may receive information from its neighbors, e.g., $z_j, j\in\mathcal{N}_k$, and aggregate them by a weighted sum $\sum_{j\in\mathcal{N}_k} w_{k,j} z_j$. Further, 
matrix $W$ satisfies
\begin{enumerate}
%\item The entries are nonnegative such that $w_{i,j} \geq 0$ for all $i,j \in [K]$. 
%\item $W$ is symmetric such that $w_{i,j} = w_{i,j}$ for all $i,j \in [K]$. 
\item[(i)] $W$ is doubly stochastic such that $\sum_{i } w_{i,j} = 1$ and $\sum_{j} w_{i,j} = 1$ for all $i,j \in [K]$.
\item[(ii)] There exists a constant $\rho \in (0,1)$ such that $\| W - \frac{1}{K} \1 \1^\top \|_2^2 = \rho $, where $\| A\|_2$ denotes the spectral norm of $A \in \RR^{K \times K}$.
%$\max \{ | \lambda_2 (W^\top W) |, | \lambda_n(W^\top W)| \} \leq \rho$ where $\lambda_k(A)$ denotes the $k$-th largest eigenvalue value of $A \in \RR^{K \times K}$.
\end{enumerate}
\end{assumption}
These assumptions on the adjacency matrix are crucial to ensure the convergence of decentralized algorithms, and are commonly made in the decentralized optimization literature \citep{lian2017can}.

% \begin{assumption}[Smoothness and Convexity]\label{assumption:smooth}
% We impose the following smoothness assumptions:
% \begin{enumerate}
% \item[(i)] $\nabla_x f^k(x,y)$ and $\nabla_y f^k(\bullet,y)$ are $L_f$-Lipschitz continuous in $(x,y)$.   
% \item[(ii)] $\nabla_y g^k(x,y;\xi^k)$ is $L_g$-Lipschitz continuous in $y$.  
% \item[(iii)]$\nabla_y g^k(x,y)$, $ \nabla_{xy}^2 g^k(x,y)$, and $\nabla_{yy}^2 g^k(x,y)$ are $L_g$-Lipschitz continuous in $(x,y)$. 
% \item[(iv)] $g^k(x,y;\xi^k)$ is $\mu_g$-strongly convex in $y$. 
% % \item The stochastic information returned by the $\cS\cO$ has bounded second-order moments.  
% \end{enumerate}
% \end{assumption}
% We defer the detailed assumptions to Section \ref{app:ass} of the supplement. 
\begin{assumption}\label{assumption:1}
Let $C_f, L_f$ be positive scalars. The outer level functions $\{ f^k \}_{k \in \cK} $ satisfy the followings.  
\begin{enumerate}
\item[(i)] There exists at least one optimal solution to Prob.~\eqref{prob:1}.
\item[(ii)] Both $\nabla_x f^k(x,y)$ and $\nabla_y f^k(x, y)$ are $L_f$-Lipschitz continuous in $(x,y)$ such that for all $x_1,x_2 \in \RR^{d_x}$ and $y_1,y_2 \in \RR^{d_y}$, 
\begin{align*}
   & \|  \nabla_x f^k(x_1,y_1) - \nabla_x f^k(x_2,y_2) \| \leq L_f (\| x_1 - x_2\| + \| y_1 - y_2 \|),
    \\
    \text{ and }  &  \|  \nabla_y f^k(x_1,y_1) - \nabla_y f^k(x_2,y_2)  \| \leq L_f (\| x_1 - x_2\| + \| y_1 - y_2 \|).
\end{align*}

 \item[(iii)] For all $x \in \RR^{d_x}$ and $y \in \RR^{d_y}$, 
 $$\EE [ \| \nabla_y f^k(x,y; \zeta^k )\|^2] \leq C_f^2 \text{ and }\EE [ \| \nabla_x f^k(x,y; \zeta^k )\|^2] \leq~C_f^2.$$
\end{enumerate}
\end{assumption}

\begin{assumption}\label{assumption:2}
Let $C_g, L_g, \widetilde L_g, \mu_g,\kappa_g $ be positive scalars.
The inner level functions $\{ g^k\}_{ k \in \cK}$ satisfy the followings. 
\begin{enumerate}
\item[(i)] For all $x \in \RR^{d_x}$,  $g(x,y)$ is $\mu_g$-strongly convex in $y$. 
\item[(ii)] For all $x \in \RR^{d_x}$ and $ y \in \RR^{d_y}$, $g^k(x,y)$ is twice continuously differentiable in $(x,y)$. 
\item[(iii)] 
$\nabla_y g^k(x,y)$, $ \nabla_{xy}^2 g^k(x,y)$, and $\nabla_{yy}^2 g^k(x,y)$ are Lipschitz continuous in $(x,y)$ such that for all $x_1,x_2 \in \RR^{d_x}$ and $y_1,y_2 \in \RR^{d_y}$, 
\begin{align*}
\| \nabla_{y} g^k(x_1,y_1) - \nabla_{y} g^k(x_2,y_2)\| & \leq L_g (\| x_1 - x_2\| + \| y_1 - y_2\| ),\\
\| \nabla_{xy}^2 g^k(x_1,y_1) - \nabla_{xy}^2 g^k(x_2,y_2)\|_F & \leq \widetilde L_g (\| x_1 - x_2\| + \| y_1 - y_2\| ),\\
   \text{ and } \| \nabla_{yy}^2 g^k(x_1,y_1) - \nabla_{yy}^2 g^k(x_2,y_2)\|_F & \leq \widetilde L_g (\| x_1 - x_2\| + \| y_1 - y_2\| ).
\end{align*}

\item[(iv)] For all  $x \in \RR^{d_x}, y \in \RR^{d_y} $,  $\nabla_{y} g^k(x, y; \xi^k )$,  $\nabla_{yy}^2 g^k(x, y; \xi^k )$,  and $\nabla_{xy}^2 g^k( x, y; \xi^k)$ have bounded second-order moments such that 
$
\EE[ \| \nabla_y g^k(x, y; \xi^k) \|^2 ] \leq C_g^2
$, $\EE[ \|  \nabla_{yy}^2 g^k(x, y; \xi^k ) \|_2^2] \leq L_g^2$,  and $\EE[ \| \nabla_{xy}^2 g^k(x, y; \xi^k) \|_2^2 ] \leq L_g^2
$
.

\item[(v)] For all  $x \in \RR^{d_x}, y \in \RR^{d_y} $,  $I - \frac{1}{L_g}\nabla_{yy}^2 g^k(x, y; \xi^k )$ has bounded second moment such that 
$
\EE[ \| I - \frac{1}{L_g}\nabla_{yy}^2 g^k(x, y; \xi^k )\|_2^2] \leq (1-\kappa_g )^2
$, where $0< \kappa_g \leq   \frac{\mu_g}{L_g} \leq 1$.
\end{enumerate}
%\xz{do the $L_g$ in (iii), (iv), (v) mean to be the same constant?}
\end{assumption}

Here we point out that the above assumptions allow the heterogeneity between functions $f^k$'s and $g^k$'s over the agents, and the smoothness and boundedness conditions are commonly adopted in SBO \citep{hong2020two,chen2021single}. 

%The oracle stated in Assumption \ref{assumption:2} (v) can be implemented via the HIA algorithm \citep{ghadimi2018approximation} and biase geometrically diminishing in the stochastic samples $b$.
\begin{algorithm}[t!]
\caption{Gossip-Based Decentralized Stochastic Bilevel Optimization}\label{alg:1}
\begin{algorithmic}[1]
\REQUIRE Step-sizes $\{ \alpha_t \}$,  $ \{ \beta_t \}$, $\{ \gamma_t\}$, total iterations $T$,  adjacency matrix $W$, smoothness constant~$L_g$, hessian sampling parameter $b$
\\
$x_0^k = \bf{0}$, $y_0^k = \bf{0}$, $u_0^k = \bf{0}$,  $s_0^k = \bf{0}$, $h_0^t = \bf{0}$, $v_{0,i}^k = \mu_g \bf{I}$ for $i=1,2,\cdots,b$ 
\FOR{$t=0, 1, \cdots, T-1$}
	\FOR{$k=1,\cdots, K$}
	\STATE \textcolor{blue}{Local sampling}: Query $\cS\cO$ at $(x_t^k,y_t^k)$ to obtain $\nabla_x f^k(x_t^k,y_t^k;\zeta_t^k)$, 
	$\nabla_y f^k (x_t^k,y_t^k;\zeta_t^k)$, 
	\\
	\hspace{2.3cm} $\nabla_y g^k (x_t^k,y_t^k;\xi_t^k )$, $\nabla_{xy}^2  g^k(x_t^k,y_t^k;\xi_t^k)$,   and  $\{ \nabla_{yy}^2 g^k(x_t^k,y_t^k;\xi_{t,i}^k)  \}_{i=1}^b$.
	\STATE \textcolor{blue}{Outer loop update}: $x_{t+1}^k = \sum_{j \in \cN_k}w_{k,j} x_t^j  - \alpha_t  \left ( s_t^k -  u_t^k q_t^k  h_t^k \right ) $.
	\STATE 
	%Query the $\cS\cO$ at $(x_t^k,y_t^k)$ to obtain $\nabla_y g^k(x_t^k,y_t^k,\xi_t^k)$ and $\nabla_{xy}^2g^k(x_t^k,y_t^k;\xi_t^k)$.
	%Query the $\cS\cO$ at $(x_t^k,y_t^k)$ to obtain stochastic gradient $\nabla_y g^k (x_t^k,y_t^k,\xi_t^k )$. 
	\textcolor{blue}{Inner loop update}:  
$
	y_{t+1}^k  =  \sum_{j \in \cN_k} w_{k,j} y_t^j   - \gamma_t \nabla_y g^k (x_t^k,y_t^k;\xi_t^k ).
$
\STATE 
\textcolor{blue}{Estimate $\nabla_x f(x_t,y_t)$:}
$
s_{t+1}^k   =(1-\beta_t) \sum_{j \in \cN_k} w_{k,j} s_t^j + \beta_t  \nabla_x f^k(x_t^k, y_t^k ; \zeta_t^k )$.
\STATE 
\textcolor{blue}{Estimate $\nabla_y f(x_t,y_t)$:}
$
h_{t+1}^k  =(1-\beta_t) \sum_{j \in \cN_k} w_{k,j} h_t^j + \beta_t  \nabla_y f^k(x_t^k, y_t^k ;  \zeta_t^k).  
$

\STATE \textcolor{blue}{Estimate $\nabla_{xy}^2 g(x_t,y_t)$:} 
$
u_{t+1}^k   =(1-\beta_t) \sum_{j \in \cN_k} w_{k,j} u_t^j + \beta_t \nabla_{xy}^2 g^k(x_t^k,  y_t^k; \xi_t^k).
$
\STATE 
\textcolor{blue}{Estimate $[\nabla_{yy}^2  g(x_t,y_t)]^{-1}$:} 
Set 
$Q_{t+1,0}^k = \mathbf{I}$
\FOR{ $i= 1,\cdots,b$}
\STATE $v_{t+1,i}^k  =(1-\beta_t) \sum_{j \in \cN_k} w_{k,j} v_{t,i}^j + \beta_t  \nabla_{yy}^2 g^k(x_t^k,y_t^k;\xi_{t,i}^k),$ 
%\xz{Where do you get $v_{t,i}^j$ when updating agent $k$?}
\STATE $Q_{t+1,i}^k = \mathbf{I} + ( \mathbf{I} - \frac{1}{L_g} v_{t+1,i}^k)Q_{t+1,i-1}^k$
\ENDFOR
\STATE Set $q_{t+1}^k = \frac{1}{L_g} Q_{t+1,b}^k$
% Set $q_t^k = \frac{b_t}{L_g}\prod_{i=1}^{p} (I - \frac{1}{L_g} \nabla_{yy}^2 g^k(x_t^k,y_t^k;\xi_t^k)) $, 
% \\
% \hspace{0.8cm} \text{ where } $p \sim \text{Unifom}[0,\cdots, b_t-1]$.
% % by HIA algorithm of \citep{ghadimi2018approximation}:
% % \begin{equation}
% %     q_t^k = \prod_{i=1}^{b_t} (I - [\nabla_{yy}^2  g^k(x_t^k,y_t^k;\xi_t^k)]^{-1})
% % \end{equation}
% Update
% $
% v_{t+1}^k  =(1-\beta_t) \sum_{j \in \cN_k} w_{k,j} v_t^j + \beta_t  q_t^k .
% $

%  		\STATE \textcolor{blue}{Gradient update:}   $z_{t+1}^k =  s_{t+1}^k -  u_{t+1}^k v_{t+1}^k h_{t+1}^k $.
 	\ENDFOR
 \ENDFOR
 \ENSURE $\bar{x}_t=\frac{1}{K} \sum_{k \in\cK} x_t^k$
\end{algorithmic}
\end{algorithm}

\section{Algorithm}

As discussed in Section~\ref{sec:challenges}, the key challenge in solving DSBO is that each agent has only access to its own data but is required to construct estimators for the gradients and Hessian averaged across all agents. It is particularly challenging to construct such estimators when limited by the network's communication protocol. 

To overcome this issue, we propose a gossip-based 
DSBO algorithm where each agent $k \in \cK$ iteratively updates a solution pair $(x_t^k,y_t^k)$ by using the combination of gossip communications and weighted-average stochastic approximation, where $y_t^k$ serves as an estimator of the best response $y_t^* := y^*(\bar x_t)$ with  $\bar x_t : = \frac{1}{K}\sum_{k\in \cK} x_t^k$.
The full algorithm is given in Alg.~\ref{alg:1}.

Here we briefly explain the idea of our DSBO algorithm. Suppose agent $k$ would like to estimate $\nabla_x f(x_t^k,y_t^k)$ by  $s_t^k$, under Alg.~\ref{alg:1} Step 6, it would query the stochastic first-order information using its own data,  make gossip communications with neighbors, and update its estimators by taking the weighted average among its previous estimate $ s_{t-1}^{k} $, neighbors $j$'s estimate $s_{t-1}^{j}$, and the newly sampled gradient $\nabla_x f^k(x_t^k,y_t^k;\zeta_t^k)$. Roughly speaking, this procedure can be viewed as taking weighted average among gradients sampled by all agents over the network, 
except that the effect of consensus should also been taken into account.

The updates for $\nabla_y f$ and  $\nabla_{xy}^2 g$ are conducted in a similar manner (Steps 7 \& 8), but it takes extra efforts to evaluate $[\nabla_{yy}^2  g ]^{-1}$, due to the absence of its unbiased estimator. 
{To be specific, we note that $\frac{1}{K}\sum_{k\in \cK} [\nabla_{yy}^2  g^k]^{-1} \neq [\frac{1}{K} \sum_{k \in \cK} \nabla_{yy}^2 g^k]^{-1}$, making the unbiased estimator of the desired term unavailable even if each agent has an unbiased estimator for $[\nabla_{yy}^2  g^k]^{-1}$. This is indeed a unique challenge for decentralized bilevel optimization. 
To overcome this issue, we propose a \emph{novel} approach that each agent $k$ constructs $b$ independent estimators $\{ v_{t,j}^k \}_{j=1}^b$ for $\nabla_{yy}^2 g(x_t^k,y_t^k)$ via consensus and stochastic approximation. We then estimate $ \nabla_{yy}^2 g(x_t^k,y_t^k) $ by utilizing the following approximation.
$$
   \Big  (\frac{1}{L_g} \nabla_{yy}^2 g(x_t^k,y_t^k)  \Big )^{-1} \approx \mathbf{I} +  \sum_{j=1}^b  \Big (\mathbf{I} - \frac{1}{L_g} \nabla_{yy}^2 g(x_t^k,y_t^k) \Big )^j \approx \mathbf{I} +  \sum_{i=1}^b \prod_{j=1}^i \Big (\mathbf{I}  - \frac{1}{L_g} v_{t,j}^k   \Big ). 
$$
We provide details in Steps 4 \& 9. 

% To solve this issue, agent $k$ uses a stochastic power iteration (similar to the HIA algorithm of \cite{ghadimi2020single}) to construct a \emph{nearly unbiased} estimator $q_t^k$ using $b_t$ stochastic Hessian $\{ \nabla_{yy}g^k(x_t^k,y_t^k; \xi_i^k) \}_{i= 1}^{b_t} $ sampled from its own data, whose bias decays to zero geometrically. With such $q_t^k$, agent $k$ can estimates 
% $\nabla_x f^k$ via consensus and stochastic approximation (Step 9). 

Finally, by utilizing the estimators obtained in the above procedure, each agent would compute the gradient \eqref{eq:gradient_BO} and update the outer solution $x_t^k$ by using the combination of gossip communication and stochastic gradient descent (Step 3). The inner loop solution $y_t^k$ can be updated by (Step 4) in a similar way.

% To obtain an estimator $s_t^k$ for $\nabla_x f(x_t^k,y_t^k)$ and address the challenges mentioned in Section~\ref{sec:challenges}, agent $k$ would first query the $\cS\cO$ for a sampled gradient $\nabla_x f^k(x_t^k,y_t^k;\zeta_t^k)$. However, such information is only unbiased estimator of $\nabla_x f^k(x_t^k,y_t^k)$ instead of $\nabla_x f(x_t^k,y_t^k)$ across all agents. Then, agent $k$ would make gossip communications with its neighbors to obtain their estimates for $\nabla_x f$, and updates its own estimation by taking weighted average of its previous estimate $ s_{t-1}^{k} $, neighbors $j \in \cN_k$ 's estimate $s_{t-1}^{j}$, and the newly sampled gradient $\nabla_x f^k(x_t^k,y_t^k;\zeta_t^k)$, provided in Step (6) of Alg.~\ref{alg:1}. 

% The estimates for $\nabla_y f^k$ and  $\nabla_{xy}^2 g^k$ can be acquired in a similar manner but it takes extra effort to evaluate $[\nabla_{yy}^2  g^k ]^{-1}$, due to the absent of its unbiased estimator. 

% and  $\bar y_t : = \frac{1}{K}\sum_{k\in \cK} y_t^k$. 
%We consider a decentralized network and denote  the adjacent matrix by $W$. We adopt $w_{i,j}$ to represent the $(i,j)$-th component of  $W$. 

We highlight the following key features of our DSBO algorithm:
(1) each agent would only communicate their estimates instead of the raw data in the gossip-communication process, which preserves data privacy. 
(2) agent $k$ would make $\cO(|\cN_k|)$ communications with its neighbors in each round, which is much smaller than the total number of agents in a naive approach. %\textcolor{blue}{(3) each agent directly estimates the hessian and reduces the per-iteration sample complexity from $\cO( \log(T))$ to $\cO(1)$, compared with those estimating the inverse hessian via stochastic power iteration (similar to the HIA algorithm of \cite{ghadimi2020single}).}
(3) the algorithm is robust to contingencies in the network. If some communication channel fails, the rest of the agents can still jointly learn provided that the network is still connected. In contrast, a single-center-multi-user network would fail completely in case of a center failure.

\section{Theory}
In this section, we analyze the performance of our DSBO algorithm for  both nonconvex and $\mu$-PL objectives, and derive the convergence rates for both cases. 

%\xz{all assumptions move to sec 3?}

\subsection{Nonconvex Objectives}\label{sec:nonconvex}
We first consider the scenario where overall objective function $F(x)$ is nonconvex. For nonconvex objectives, given the total number of iterations $T$, we employ the step-sizes in a constant form  that 
\begin{equation}\label{eq:stepsize}
\alpha_t = C_0 \sqrt{\tfrac{K}{T}} \text{ and }   \beta_t = \gamma_t = \sqrt{\tfrac{K}{T}}, 
%  b_t = 2 | \lceil \log_{\tilde c} T \rceil |, 
\text{ for all }t=0,1,\cdots, T,
\end{equation} 
where $C_0 > 0 $ is a small constant. 

\textbf{Compounded effect of consensus and SBO:}
As discussed earlier, to derive the convergence rate of SBO under decentralized federated setting, the key step is to quantify the compounded effect between the consensus errors induced by the network structure and the biases induced by estimating gradient within~\eqref{eq:gradient_BO}. Distinct from the central-server or non-federated regimes, the consensus errors induced by the decentralized network structure has to be handled carefully. 
%To address this issue, we decompose the averaged estimation error across all agents into a stochastic recursion, where the bias terms are induced by consensus errors and variance terms are incurred from the stochastic sampling environment. 
We conduct careful analysis to derive the contraction of consensus errors, and further show that both bias and variance of the averaged estimator diminish to zero, establishing a nontrivial convergence argument to the desired gradient and Hessian. In particular, the estimators preserve a concentration property so that their variances decrease proportionally to $1/K$, suggesting that the network consensus effect would not degrade the concentration of the generated stochastic samples. To achieve the best possible convergence rate, we carefully set the algorithm parameters, including step-sizes $\alpha_t, \gamma_t$ and averaging weights $\beta_t$, to control the above consensus errors and biases.

We provide the convergence rate of Alg.~\ref{alg:1} in the following theorem and defer the detailed proof to  Section \ref{app:proof_nonconvex} of the supplementary material. 
%\xz{currently there is no proof sketch.}

%%%%%%%%%%%%%
%%%%%%%%%%%%%%
\begin{theorem}\label{thm:nonconvex}
Suppose Assumptions \ref{assumption:SO}, \ref{assumption:W}, \ref{assumption:1}, and \ref{assumption:2} hold. Letting $\bar x_t = \frac{1}{K}\sum_{k \in \cK} x_t^k$, then
% Let $\{ x_t\}_{t=0}^{T-1}$ be the sequence generated by Alg.~\ref{alg:1}, with step-sizes being $\alpha_t = C_0 \sqrt{ \frac{K}{T}}$ and $\beta_t = \gamma_t = \sqrt{ \frac{K}{T}}$ for $t=0,1,\cdots, T-1$ and $C_0 \leq \tilde C_0$, and . Then we have that 
    \begin{equation*}
\begin{split}
&\frac{1}{T}\sum_{t=0}^{T-1} \EE[ \| \nabla F(\bar x_t)\|^2]  \leq  \cO \left (\frac{1}{ \sqrt{ KT} } \right )  + \cO \left ( \frac{K}{T(1- \rho)^2 } \right ) . 
  \end{split}
\end{equation*}
\end{theorem}
More details and proof are deferred to Section \ref{app:proof_nonconvex} of the supplement.

\textbf{Effect of consensus:}
In this result, the $\cO(\tfrac{K}{T(1- \rho)^2})$ term represents the errors induced by consensus of network. Despite the dependency with the network structure, such term diminishes to zero in the order of $\cO(1/T)$, becoming a small order term when $T$ is large. Consequently, given the network topology, the asymptotic convergence behavior of DSBO is independent of the network structure, answering the question (ii) raised in Section~\ref{sec:intro}. 
\\
\textbf{Linear speedup:} Because each agent queries $\cO(1)$ stochastic samples per round, clearly the required iteration and per-agent sample complexities for finding an $\epsilon$-stationary point such that 
$
\frac{1}{T}\sum_{t=0}^{T-1} \EE[ \| \nabla F(\bar x_t)\|^2]  \leq  \epsilon
$
are both $\cO(\frac{1}{K \epsilon^2})$. Such result implies that our algorithm achieves a linear speed-up effect proportionate to the number of agents $K$. In other words, in the presence of more agents, each agent needs to query fewer stochastic samples to achieve a specified accuracy. 
Meanwhile, our rate also matches the best-known $\cO(1/K\epsilon^2)$ iteration and per-agent sample complexities under the decentralized vanilla stochastic gradient descent settings \citep{lian2018asynchronous}.  To our knowledge, it is the first time  such a   result has been established for DSBO problems. 
\\
\textbf{Single-center-multi-user-federated SBO}: We point out that a simplified version of our algorithm solves single-center-multi-user-federated SBO, where the central server directly collects information from each agent and calculates the gradient by employing the weighted-average stochastic approximation scheme among the collected information. In such scenario, the agents no longer  make gossip communication with neighbors but synchronously receives a common solution from the central server, so that the consensus effect would disappear. In Theorem \ref{thm:single_center}, we show that a variant of our algorithm can indeed achieve the same $\cO(\frac{1}{\sqrt{KT}})$ convergence rate in this setting.

\subsection{$\mu$-PL Objectives}\label{sec:PL}
%\xz{move to appendix if we run out of space.}

%In Section \ref{sec:nonconvex}, we have studied the convergence behavior of Algorithm~\ref{alg:1} for nonconvex objectives, and have established the $\cO(1/K\epsilon^2)$ iteration and per-node sample complexities for finding an $\epsilon$-stationary point such that $\EE[ \| \nabla F(\bar x_t) \|^2] \leq \epsilon$. It is  natural to ask whether we can improve the convergence rate for strongly convex objectives. 

Next we study the case where the objective function satisfies the following $\mu$-PL condition. 
\begin{assumption}\label{assumption:PL}
There exists a constant $\mu >0$ such that the objective satisfies the PL condition: 
\begin{equation*}\label{def:PL}
    2 \mu (F(x ) - F^*) \leq \| \nabla F(x)\|^2.
\end{equation*}
\end{assumption}

\begin{comment}
To investigate the convergence behavior of $\mu$-PL functions, we first present the following result to characterize the contraction of the objective gap $F(\bar x_t) - F^*$. 

\sg{move to appendix}

In the above result, we decompose the objective suboptimality gap $F(\bar x_{t+1}) - F^*$ incurred in each iteration $t$ in terms of the estimation errors. We further observe that when $(1-\alpha_t \mu) < 1$, the above result forms a contractive  recursion, which allows Alg.~\ref{alg:1} to achieve a faster  convergence rate. 

\end{comment}

Note that the class of 
strongly convex functions is a special case of $\mu$-PL functions. 
To utilize the $\mu$-PL property and achieve fast convergence, unlike the nonconvex case \eqref{eq:stepsize} where the step-sizes are set as constants depending on the total number of iterations $T$,  we employ  step-sizes in a diminishing form that 
\begin{equation}\label{eq:stepsize_PL}
\alpha_t = \frac{2}{\mu(C_1 + t)} \text{ and }  \beta_t =  \gamma_t = \frac{C_1}{C_1 + t}, 
%b_t = 2| \lceil \log_{\tilde c} t \rceil | \ \ 
\text{ for } t \geq 1,
\end{equation}
where $C_1 >0$ is  a large constant. 
By following an analysis routine similar to that in the nonconvex scenario, in the next result, we derive the convergence rate of Alg.~\ref{alg:1}  for $\mu$-PL objectives. 

\begin{theorem}\label{thm:PL}
Suppose Assumptions \ref{assumption:SO}, \ref{assumption:W}, \ref{assumption:1}, and \ref{assumption:2} hold and the function satisfies the $\mu$-PL Assumption~\ref{assumption:PL}.  Letting $\bar x_T = \frac{1}{K}\sum_{k \in \cK} x_T^k$, then
%There exists a constant $C_1 >0$ such that the following holds for $T \geq 1$.
% Let $\{ \bar x_t\}_{t=1}^T $ be the solutions sequence generated by Algorithm \ref{alg:1} with the step-sizes chosen as  $\alpha_t  = \frac{2}{\mu(t+1)}$ and $\beta_t = \gamma_t =  \frac{C}{t+1}$ for $C >0 $ sufficiently large. Then we have 
\begin{equation*}
\EE[ F(\bar x_T) ] - F^* \leq \cO \left (\frac{1}{KT} \right ) + \cO \left (\frac{\ln T}{T^2(1-\rho)^2} \right ).
\end{equation*}
Both iteration and per-agent sample complexities for finding an $\epsilon$-optimal point $\EE[F(\bar x_T)] - F^* \leq \epsilon$ are $\cO(\frac{1}{K\epsilon})$.
\end{theorem}

Details and proof are deferred to  Section \ref{app:proof_of_thm_PL} of the supplement.
This result shows that our algorithm achieves a faster convergence rate for functions satisfying the $\mu$-PL condition in terms of both iteration and sample complexities. 
First, similar as the nonconvex scenario, the consensus error decays in the order of $\tilde \cO(\frac{1}{T^2(1-\rho)^2})$. Dominated by $\cO(\frac{1}{KT})$, such consensus decaying order indicates that the network structure would not affect Alg.~\ref{alg:1}'s asymptotic convergence behavior under $\mu$-PL objectives. 
Meanwhile, the above result implies that Alg.~\ref{alg:1} speeds up  linearly  with the number of agents and matches the optimal $\cO(1/\epsilon)$ sample complexity for single-sever vanilla strongly-convex stochastic optimization~\citep{rakhlin2011making}. As a result, our algorithm achieves the \emph{optimal} sample complexities for decentralized stochastic bilevel optimization, establishing the benchmark. 

%\subsection{Special cases and Discussions}

\section{Numerical Experiments}\label{sec:numerics}
In this section, we validate the practical performance of our algorithm in two examples: hyper-parameter optimization and policy evaluation in Markov Decision Processes (MDP), on artificially constructed decentralized ring networks. 
We run the experiments on a single server desktop computer. We provide the details of federate hyper-parameter optimization and federated policy evaluation.

%\xz{are we putting mdp result in appendix or main text? currently it's in the main text.}

\textbf{Hyper-parameter Optimization}
We consider federated hyper-parameter optimization \eqref{prob:hyperopt} for a handwriting recognition problem over the Australia handwriting dataset \citep{chang2011libsvm} consisting of data points $(w_i,z_i)$, where $w_i \in \RR^{14}$ is the feature and $z_i \in \{ 0,1\}$ represents whether this data point belongs to category ``1'' or not.  
%\mw{intro dataset here} 
In our experiment, we consider the  sigmoid loss function that $l_i(z) = 1/(1+ \exp(-z))$ and a strongly convex regularizer 
 $\cR(x,y)  = \sum_{i=1}^d \frac{x_i }{2} \| y_i \|^2 $.
 We consider a ring network of $K$ agents where each agent $i$ preserves two neighbors $(i-1)$ and $(i+1)$ and conduct gossip communication strategy with adjacency matrix $w_{i,j} = \frac{1}{3}$ for $ j  \in \{ i-1,i,i+1 \}$. 
 
 Before testing Alg.~\ref{alg:1}, we first randomly split the dataset for training and validation, and then allocates 
both training and validation dataset over $K$ agents. We then run  Algorithm~\ref{alg:1} for $T=20000$ iterations, with $b = 200$, $\alpha_t = 0.1\sqrt{K/T}$, and $\beta_t = \gamma_t = 10\sqrt{K/T}$. 

% Australia  handwriting dataset \citep{chang2011libsvm}  consisting of data points $(w_i,z_i)$, where $w_i \in \RR^{14}$ is the feature and $z_i \in \{ 0,1\}$ represents whether this data point belongs to category ``1'' or not. 

% Given the size of ``0'' label data $n_{\text{train0}}$ within the training set, let $m$ be the amount of data allocated to each agent, we split that data heterogeneously such that the first $\lfloor \frac{n_{\text{train0}}}{m} \rfloor $ agents contain ``0'' label alone, the agent $\lfloor \frac{n_{\text{train0}}}{m} \rfloor +1$ contains mixture of ``0'' and ``1'' labels, and the rest agents contain 
% ``1'' label only. The validation data are allocated in a similar way. 
% \sg{remove this details, don't need to discuss too much}

For benchmark comparison, we implement a baseline algorithm Decentralized Bilevel Stochastic Approximation (DBSA), a naive extension of the double-loop BSA algorithm \citep{ghadimi2018approximation} in the decentralized setting,  formally stated in Section~\ref{app:hyper_opt} of the supplementary materials.

We first consider $K = 5$, test Alg.~\ref{alg:1} for $5 \times 10^4$ iterations, and compare its performance with DBSA.  We report the validation loss against total samples  in Figure~\ref{fig:1} and observe that DSBO exhibits superior performance to DBSA. In particular, Further, we observe that our algorithm outperforms the baseline algorithm DBSA that it requires 
fewer samples for DSBO to achieve a certain accuracy.

%than DBSA. 
%Further, our DSBO algorithm is able to find a solution whose validation loss is only 0.308 while DBSA can only find solutions that have validation loss around 0.326. This suggests DBSA may not converge to the optimal solution even if sufficiently large number of iterations are conducted, because the naive stochastic gradient employed by DBSA is not necessarily an unbiased estimator of the overall gradient, as illustrated in  \eqref{eq:biased_grad}. 

To investigate the efficiency of Alg.~\ref{alg:1} to the network structure, we test Alg.~\ref{alg:1} 
over $K = 5,10,20$, and report the details of training and validation loss in Figure~\ref{fig:1}. Further, comparing the performances of Alg.~\ref{alg:1} over different agents $K=5,10,20$, we observe that Alg.~\ref{alg:1} converges faster by using more  agents. 
This observation suggests that Alg.~\ref{alg:1} exhibits a speed-up effect by using more agents. 
%We provide additional numerical results regarding prediction accuracy in Section \ref{app:hyper_opt} of the supplement.

\begin{figure}[t]
\begin{minipage}{0.5\textwidth}\label{fig:source_v_trans}
    \centering
    \includegraphics[width=0.7\linewidth]{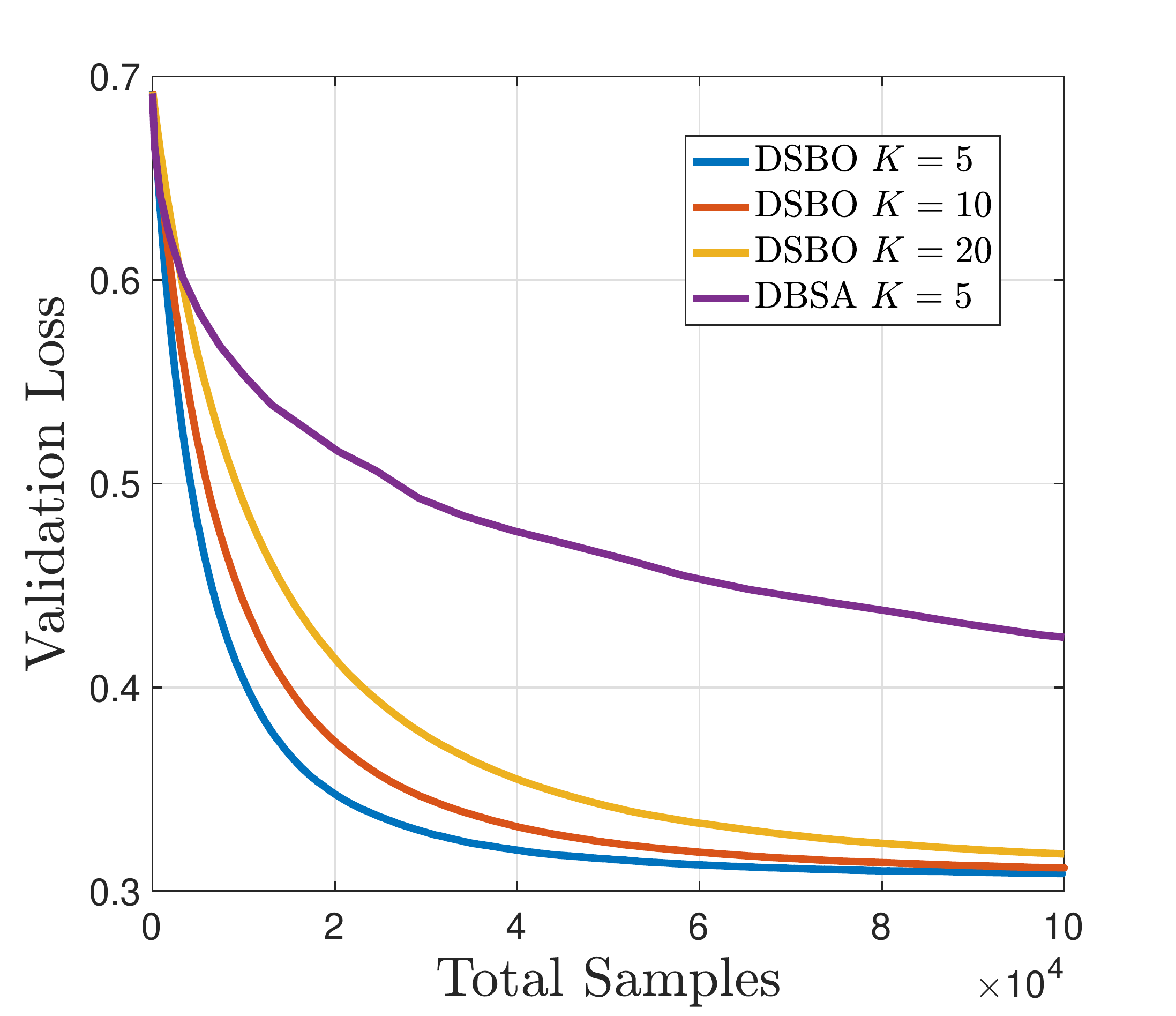}\\
    (a)
\end{minipage}
\begin{minipage}{0.5\textwidth}
    \centering
    \includegraphics[width=0.7\linewidth]{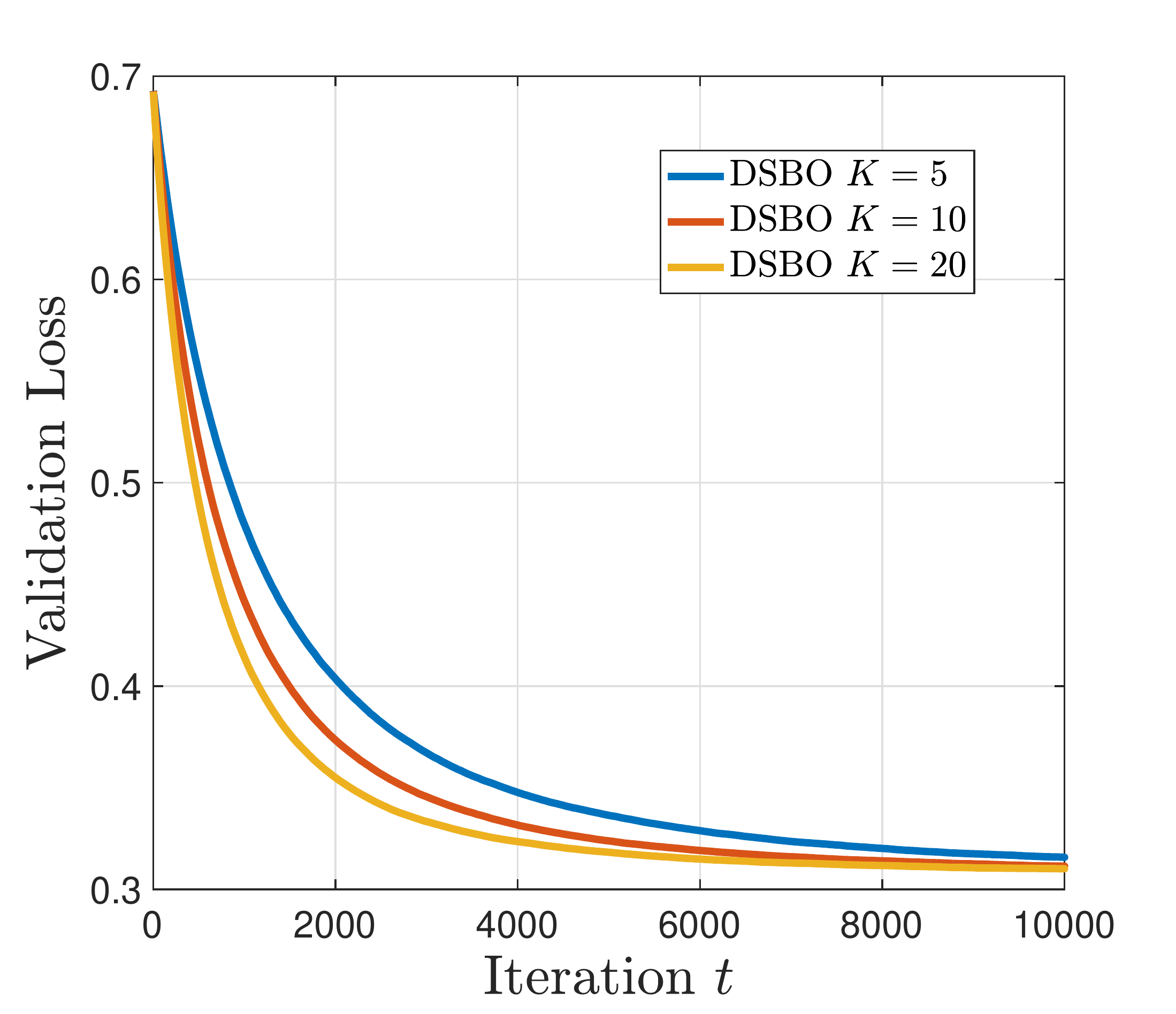}\\
    (b)
\end{minipage}
\caption{(a) Empirical averaged training loss against total samples  for DSBO $K=5,10,20$ and DSGD $K = 5$. 
(b) Empirical averaged validation loss against iteration for DSBO $K=5,10,20$. All figures are generated through 10 independent simulations over australia handwriting dataset.
}
% Training and validation loss for handwriting recognization over australia dataset} \mw{spacing, expand caption}  
\label{fig:1}
\end{figure}

\paragraph{Distributed policy evaluation for reinforcement learning}
 We consider a multi-agent MDP problem arising in reinforcement learning. Let $\cS$ be the state space, for any state $s \in \cS$, we denote by $V(s)$ the value function. 
we consider the scenario where the value function can be approximated by a linear function such that $V(s) = \phi_s^\top x^*$, where $\phi_s \in \RR^m$ is a feature and $x^* \in \RR^m$ is an unknown parameter.   
To obtain the optimal $x^*$, we consider the following regularized Bellman minimization problem
\begin{equation*}
    \min_{x} \ F(x) = \tfrac{1}{2|\cS|} \sum_{s \in \cS} \big (\phi_s^\top x - \EE_{s'}[r(s,s') + \gamma \phi_{s'}^\top x \mid s ] \big )^2 + \tfrac{\lambda \|x \|^2}{2},
\end{equation*}
where $r(s,s')$ is the random reward incurred from transition $s$ to $s'$, $\gamma \in (0, 1)$ is the discount factor, $\lambda$ is the coefficient for the $l_2$-regularizer, and the expectation is taken over all random transitions from $s$ to $s'$. 

In the federated learning setting, we consider a ring network of $K$ agents. Here each agent $k$ has access to its own data with heterogeneous random reward function $r^k$ and can only communicate with its two neighbors $k+1$ and $k-1$. 
We denote by $y_s^*(x) = \phi_s^\top x - \EE_{s'}[\frac{1}{K} \sum_{k\in\cK}r^k(s,s') + \gamma \phi_{s'}^\top x \mid s ] $ where $r^k(s,s')$ is the random reward function for agent $k$, the above problem can be recast as a bilevel optimization problem 
\begin{equation*}
    \min_{x \in \RR^d }f(x,y^*(x)) = \sum_{ s \in \cS}(  \phi_{s}^\top x - y_s^*(x) )^2  . 
\end{equation*}
As pointed out by \citep{wang2016accelerating}, the above problem is $\lambda$-strongly convex. 
%\xz{maybe swap $x$ and $y$ to match the order in previous sections? i.e. x denotes outer and y denotes inner variables.}

In our experiments, we simulate an environment with state space $|\cS| = 100$ and set the regularizer parameter $\lambda = 1$. 
We test the performance of Alg.\ref{alg:1} over three setups with $K= 5,10,20$ and conduct 10 independent simulations for each $K$. We implement a baseline double-loop algorithm DSGD that first estimates $y_s^*(x_t)$ with $t$ samples in iteration $t$ and then optimizes the solution $x_t$. We defer the implementation details of the environment and  above algorithms to Section \ref{app:MDP} of the supplement.  

%\mw{grammar: we dont say total agents, total time. say total number of agents}

%For benchmark comparison, we generate 
We first consider $K = 5$, run Alg.~\ref{alg:1} for $10^4$ iterations and compare its performance with DSGD. We plot the empirical averaged mean square error $\|\bar  x_t - x^*\|^2$ against total samples  generated by \emph{all} agents in Figure \ref{fig:1}. This empirical result suggests that Alg.~\ref{alg:1} outperforms the DSGD. 
To investigate the convergence rate of DSBO, we compare the performance of DSBO over all three setups $K=5,10,20$ and plot the trajectory of averaged log-error $ \log( \| \bar x_t - x^* \|^2)$ averaged, with one straight line of slope -1 provided for comparison. We observe that over all four setups, the slopes of $\log(\| \bar x_t - x^*\|^2)$ are close to -1,  matching our theoretical claim in Theorem~\ref{thm:PL} that Alg.~\ref{alg:1} converges at the rate of $\cO(1/t)$ for strongly convex objectives. 

In the above experiment, we also note that Alg.~\ref{alg:1} converges faster when using more agents.
To further demonstrate the linear speedup effect, we compute the total generated samples for finding an $\epsilon$-optimal solution $\|\bar x_t - x^* \|^2 \leq \epsilon $ and plot the 75\% confidence region of log-sample against number of agents $K = 5,10,20$ in Figure~\ref{fig:MDP1}. We observe that it takes a roughly same amount of samples to find a $10^{-6}$-optimal solution despite different number of agents are involved. This suggests that the per-node sample complexity decreases linearly to $K$ and validates the linear speedup claim in Theorem~\ref{thm:PL}. We provide additional numerical results for other optimality level $\epsilon$ in Section~\ref{app:MDP} of the supplementary material to further demonstrate the linear speedup effect.

\begin{figure}[t]
\begin{minipage}{0.32\textwidth}
    \centering
        \includegraphics[width=1\linewidth]{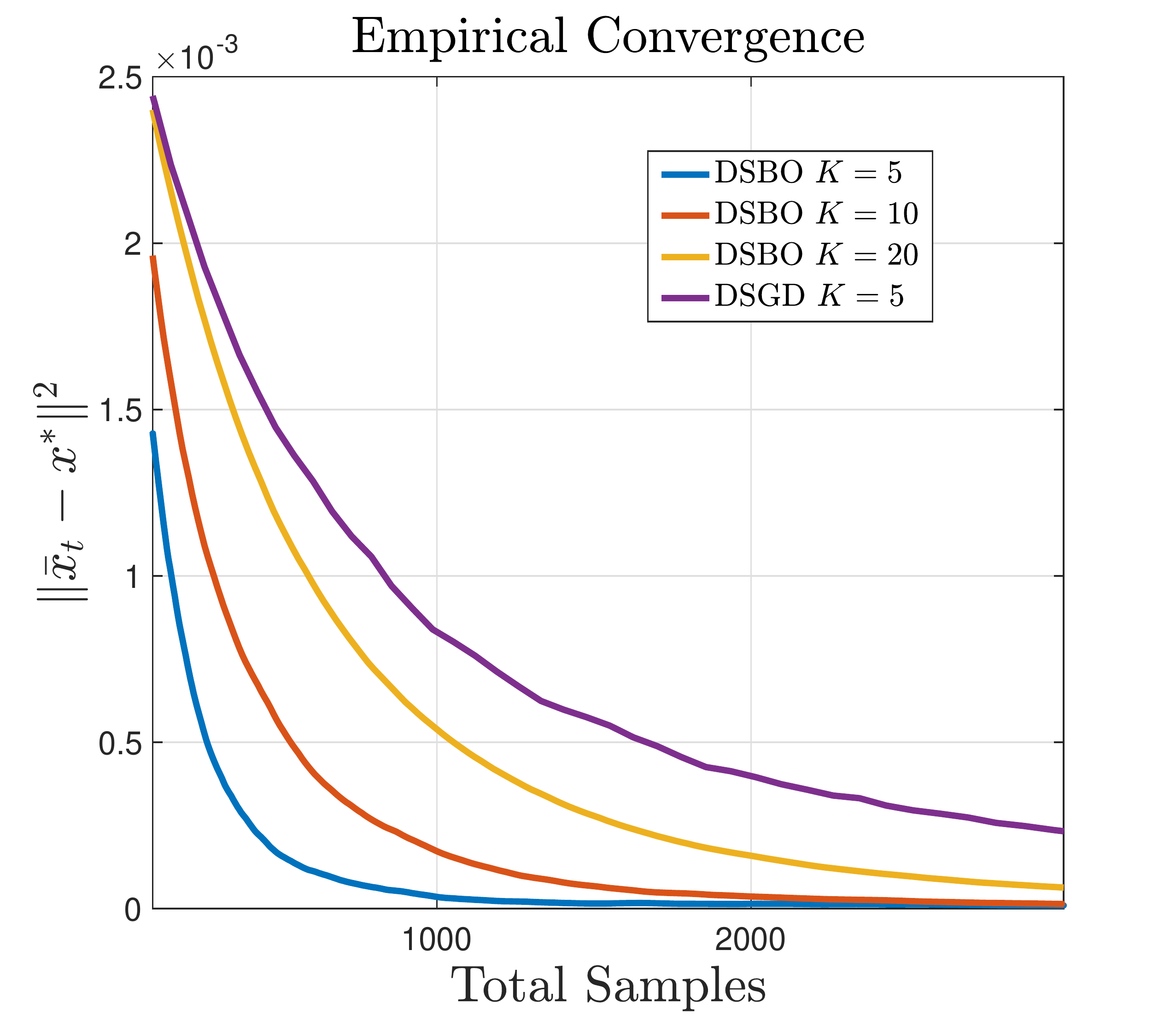}
        (a)
        \end{minipage}  
        \begin{minipage}{0.32\textwidth}
        \centering
   \includegraphics[width=1\linewidth]{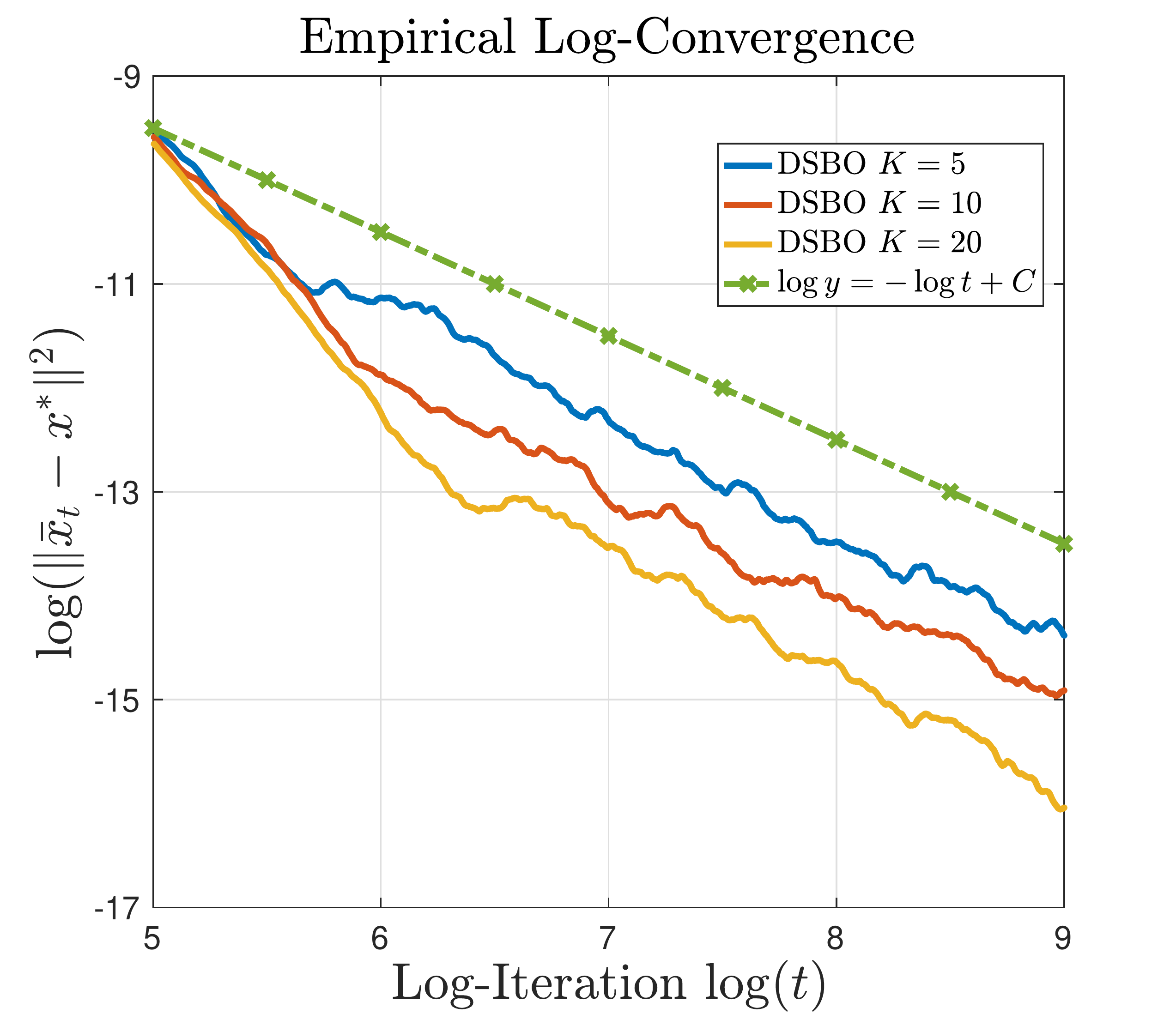} 
   (b)
   \end{minipage}
                  \begin{minipage}{0.32\textwidth}
        \centering
   \includegraphics[width=1\linewidth]{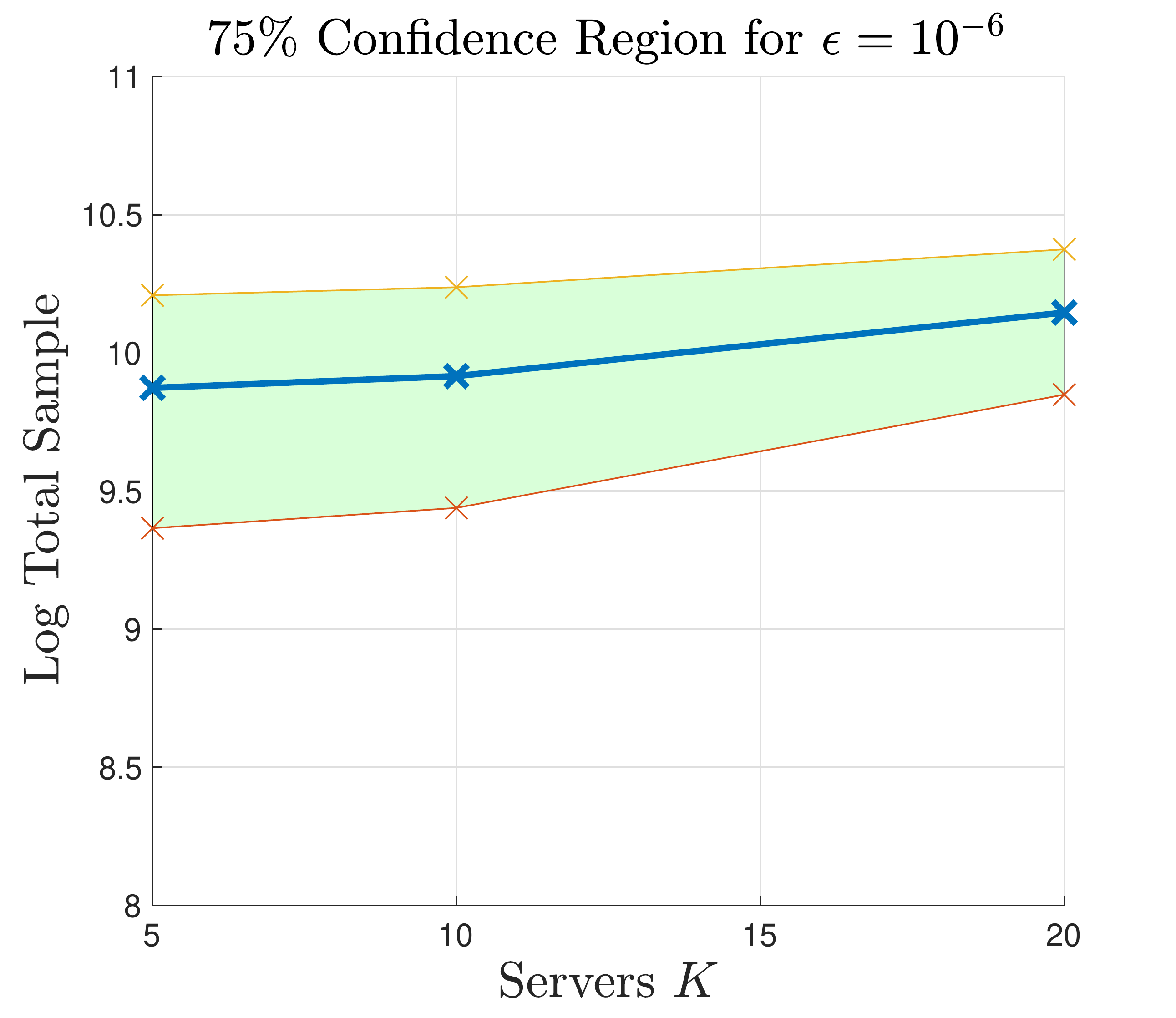}
   (c)
   \end{minipage}
    \caption{(a) Empirical averaged MSE$\| \bar x_t - x^*\|^2$ against total samples  for DSBO $K=5,10,20$ and DSGD $K = 5$.
    (b) Empirical averaged log-MSE $\log(\| \bar x_t - x^*\|^2)$ against log-iteration $\log(t)$ for DSBO $K=5,10,20$.
    (c) 75\% confidence region of log- total samples for achieving $\|\bar x_t - x^* \|^2 \leq \epsilon$, with varying network sizes $K = 5,10,20$. All figures are generated through 10 independent simulations. 
    }
    %\mw{log-sample confusing. total or per-agent?}}
    \label{fig:MDP1}
\end{figure}

%\sg{delete iteration 3. }
\section{Conclusion}
In this paper, we propose a novel formulation of decentralized stochastic bilevel optimization. We develop a gossip-based stochastic approximation scheme to solve this problem in various settings. We show that our proposed algorithm finds a stationary point at a  rate of $\cO(\tfrac{1}{\sqrt{KT}})$ for nonconvex objectives, and converges to the optimal solution at a  rate of $\cO(\tfrac{1}{KT})$ for strongly convex objectives. Numerical experiments on hyper-parameter optimization and  multi-agent federated MDP demonstrate the practical efficiency of our algorithm, exhibit the effect of speed-up in decentralized setting, and validate our theoretical claims.
In our future work, we wish to develop algorithms that achieve lower iteration complexities and enjoy lower communication cost.

\newpage 
\bibliography{SCGD,bib}

\newpage 

\appendix 
\section*{Supplementary Material}

% \textcolor{red}{Note the reviewers: After the submission deadline, we revised Steps 4 and 9 of Algorithm~\ref{alg:1}. In this revised manuscript, we construct an estimator $\bar v_t$ for the hessian $\nabla_{yy}^2 g(\bar x_t, y_t^*)$ and adopt $[\bar v_t]^{-1}$ as the estimator for $[\nabla_{yy}^2 g(\bar x_t, y_t^*)]^{-1}$, instead of directly estimating $[\nabla_{yy}^2 g(\bar x_t, y_t^*)]^{-1}$ using gossip communication and stochastic approximation in the prior version. The revised scheme allows us better handle the heterogeneity of hessians sampled by each agent, in a decentralized network.
% }
\section*{Outline}

\begin{itemize} 
    \item Section \ref{app:A}:  Notation, assumptions, and supporting lemmas
    \begin{itemize}
       % \item \ref{app:ass} Detailed assumptions
        \item \ref{app:lemmas} Technical lemmas for Lipschitz properties and hessian inverse estimation
    \end{itemize}
    \item Section \ref{app:B}: Proof of results for nonconvex objectives 
    \begin{itemize} 
    \item \ref{lemma:boundedness}-\ref{app:lemma_ht} Technical Lemmas for  consensus and estimation errors
    \item \ref{app:proof_nonconvex} Proof of Theorem \ref{thm:nonconvex}
    \end{itemize}
    \item Section \ref{app:C}: Proof of results for $\mu$-PL objectives
    \begin{itemize}
    \item \ref{app:main_PL} Technical Lemma for convergence properties of $\mu$-PL functions
        \item \ref{app:proof_of_thm_PL} Proof of Theorem~\ref{thm:PL}
    \end{itemize}
\end{itemize}

\section{Notation and Technical Lemmas} \label{app:A}
% \subsection{Notations} \mw{move these notations to the place where they are used for the first time}
For notational convenience, we denote by $\bar u_t = \frac{1}{K} \sum_{k=1}^K u_t^k$ the averaged estimates of $\nabla_{xy}^2 g(\bar x_t, y^*_t)$ over $u_t^k$'s within the network.  We denote by $\bar v_{t,j} = \frac{1}{K} \sum_{k=1}^K v_{t,j}^k \in \RR^{d_y \times d_y}$, $ \bar s_t = \frac{1}{K} \sum_{k=1}^K s_t^k$,  and $ \bar h_t = \frac{1}{K} \sum_{k=1}^K h_t^k$.   We denote by $\bar z_k = \sum_{k =1}^K z_t^k$. We denote by $\| a\| = \| a\|_2$ for a vector $a$ and denote by $\| A\| = \| A\|_2$ for a matrix $A$. We denote by $\|A \|_F $ the Frobenius norm for a matrix $A$ and  denote by $\lv A, B \rv_F = \sum_{i,j} A_{ij} B_{ij}$ the Frobenius inner product for two matrices $A$ and $B$. 
% We denote by 
% $$
% \overline{\nabla} f(x,y) = \nabla_x f(x,y) - \nabla_{xy}g(x,y) [\nabla_{yy}g(x,y)]^{-1} \nabla f_y(x,y).
% $$
For any $\bar x_t \in \RR^{d_x}$, we denote by $y_t^* = y^*(\bar x_t)$. 
For notational convenience, we drop the sub-scripts $\xi^k,\zeta^k$ within the expectations $\EE_{\xi^k}[\cdot]$ and $\EE_{\zeta^k}[\cdot]$.

In this paper, we denote by $L_q  = \frac{1}{\kappa_g L_g} \geq  \mu_g^{-1}$ and observe that $\|[\nabla_{yy}^2g(x,y) ]^{-1}\|_2^2 \leq \mu_g^{-2} \leq L_q^2$  for all  $x \in \RR^{d_x}$ and $y \in \RR^{d_y} $. For notational convenience, we use $\sigma_g, \sigma_f >0$ to represent the upper bounds of standard deviations such that 
\begin{align*}
\EE [ \| \nabla_x f^k(x,y;\zeta^k) - \nabla_x f^k(x,y) \|^2 ] \leq \sigma_f^2, \EE [ \| \nabla_y f^k(x,y;\zeta^k) - \nabla_y f^k(x,y) \|^2] \leq \sigma_f^2,
\end{align*}
and 
\begin{align*}
    \EE [ \| \nabla_y g^k(x,y;\xi^k) - \nabla_y g^k(x,y) \|^2 ] & \leq \sigma_g^2,  \\
     \EE [ \| \nabla_{xy}^2 g^k(x,y;\xi^k) - \nabla_{xy}^2 g^k(x,y) \|_F^2 ]\leq \sigma_g^2, & \  \EE [ \| \nabla_{yy}^2 g^k(x,y;\xi^k) - \nabla_{yy}^2 g^k(x,y) \|_F^2 ] \leq \sigma_g^2. 
\end{align*}
We also 
adopt constants $L_F, L_y >0$ to quantify the Lipschitz properties, specified in Section~\ref{app:lemmas}. Given $(x,y)$, we use  $\nabla_x f^k(x,y; \zeta^k )$, $\nabla_y f^k(x,y; \zeta^k )$, $\nabla_y g^k(x, y; \xi^k )$, $\nabla_{xy}^2 g^k( x, y; \xi^k)$, and $\nabla_{yy}^2 g^k( x, y; \xi^k) $ to represent the independent stochastic information sampled in round $t$ by agent $k$.  Such independent samples can be obtained by querying the $\cS \cO$ at $(x,y)$ for three times. 

% \subsection{Applications}

% \paragraph{Representation transfer in multi-task learning}

% \paragraph{Min-Max Optimization} Let $g(x,y, \xi)  = - f(x,y, \zeta) : \RR^{d_x} \times \RR^{d_y} \to \RR  $, SBO can be reformulated as the stochastic min-max optimization problem 
% \begin{equation*}
%     \min_{x\in \RR^{d_x}} \max_{y \in \RR^{d_y} } \EE_{\xi} [ f(x,y;\xi)]. 
% \end{equation*}
% Numerous efforts have been made to solve the above stochastic min-max optimization problem because of its  wide applications in GAN \citep{goodfellow2014generative}, robust optimization \citep{ben2001approximate,ben2002robust,bertsimas2003robust,bertsimas2004price}, game theory \citep{myerson1997game,nouiehed2019solving}, etc. 

% \sg{$L(x,\lambda)$}
% \xz{Formulating 4 applications might be too many? Maybe we can instead focus on the two applications we run experiments on, namely hyperparameter optimization and federated policy evaluation. Otherwise readers might get distracted by all of these different formulations. We could move all the existing formulations to the appendix, and point interesting readers there if they are interested in a particular application.}

\subsection{Technical Lemmas for Lipschitz Properties and Hessian Inverse Estimation}\label{app:lemmas}

We first restate Lemmas 2.2 of \citep{ghadimi2018approximation}  to characterize the smoothness properties of  $y^*(x)$ and $\nabla F(x)$. 
%\sg{Need to explicitly write the constants. }

\begin{lemma}\label{lemma:Lip}
Suppose Assumptions \ref{assumption:1} and \ref{assumption:2} hold. Then  there exist constants $L_F, L_y >0$ dependent on the constants within Assumptions \ref{assumption:1} and \ref{assumption:2} such that for any $x, x_1, x_2 \in \RR^{d_x} $, the followings hold. 
\begin{equation*}  
\begin{split}
 \| \nabla F(x_1) - \nabla  F(x_2) \| \leq  L_F \| x_1 - x_2\| \mbox{ and }  \| y^{\star}(x_1) - y^{\star}(x_2)\| \leq  L_y  \| x_1 - x_2\|  
    . 
\end{split}
\end{equation*}

\end{lemma}
In this supplementary material, we  adopt $L_F$ and $L_y$ to quantity the Lipschitz  properties of the above functions. 

% We then restate Lemma 3.2 of \cite{ghadimi2018approximation} to analyze the bias of $q_t^k$ (Algorithm~\ref{alg:1} Step 9).
% \begin{lemma}\label{lemma:biased_q}
% Let  $\kappa_g = L_g/\mu_g$ be the condition number of $g^k$ and $q_t^k$ be the estimator of $\nabla_{yy}^2 g(x_t^k  ,y _t^k  )^{-1}$ returned by Algorithm \ref{alg:1} Step 9, then we have 
% \begin{equation*} 
% \begin{split}
%  \|  \nabla_{yy}^2 g^k(x_t^k ,y_t^k  )^{-1}  - \EE[q_t^k \mid x_t^k, y_t^k ]\|  \leq \tfrac{1}{\mu_g} (\tfrac{\kappa_g - 1}{\kappa_g})^{b_t} \text{ and }
%   \EE [ \|   q_t^k \|^2 ]  \leq  \sigma_q^2, \text{ for some } \sigma_q >0.
% \end{split}
% \end{equation*}
% \end{lemma}

\begin{lemma}\label{lemma:inverse_diff}
 Let $A $ be a positive definite matrix such that $ \delta I \succeq A \succ \bf{0}$ for some $0 < \delta < 1$, and $A_1,\cdots, A_k$ be $k$ matrices such that 
 $\EE[ \| \prod_{i=j}^k A_i\|_2^2] \leq \delta^{2(k-j+1)}$ for $1\leq j \leq k$. Let $   Q_k = I + A_k +A_{k-1}  A_k  + \cdots + A_1 A_2 \cdots A_k$, then the following holds. 
 \begin{equation*}
    \begin{split}
     \EE [ \|  (I-A)^{-1}   - Q_k \|_2^2 ] 
  \leq \frac{1}{1-\delta} \Big ( \sum_{1 \leq j \leq k+1 }  \delta^{k-j} \EE[ \| A_j - A \|_2^2]  \Big ) + \frac{ \delta^{k+1} }{(1-\delta)^3}.
    \end{split} 
\end{equation*}
\end{lemma}
\textit{Proof:}
Recall that for any positive definite matrix $A$ such that $\delta I  \succ A \succeq 0 $ for some $0<\delta <1$, we have 
\begin{equation*}
    (I- A)^{-1} = \sum_{i=0}^\infty A^i = I + A + A^2 + \cdots  .
\end{equation*}
Letting $Q = (I- A)^{-1}$, we have 
\begin{equation*}
    Q = I + AQ \text{ and } \|Q \|_2^2  = \| (I- A)^{-1} \|_2^2 \leq \frac{1}{(1-\delta)^2}.
\end{equation*}
We define an auxiliary sequence $Q_1 = I + A_1$, $Q_2 = I + A_2  Q_1,\cdots$, and  $Q_{k-1} = I + A_{k-1} Q_{k-2}$,  and note that $Q_k = I + A_k Q_{k-1}$. Consider $ (I-A)^{-1} - Q_k$, by utilizing the above  sequence, we obtain 
\begin{equation*}
\begin{split}
   (I-A)^{-1} - Q_k & = A Q  - A_k Q_{k-1} = (A- A_k) Q + A_k (Q - Q_{k-1})  + A^{k+1}Q.
%   \\
%   & = (A- A_k) Q + A_k ( A - A_{k-1} )Q + A_k A_{k-1} ( Q - Q_{k-2})  + A^{k+1}Q.
   \end{split}
\end{equation*}
We note that $Q - Q_{k-1} = AQ -A_{k-1}Q_{k-2} = (A- A_{k-1})Q + A_{k-1}(Q - Q_{k-2})$. 
By using such induction relationship, we can quantify the estimation error by 
\begin{equation*}
\begin{split}
  \|  (I-A)^{-1} - Q_k \|_2  
   & \leq \| A- A_k \|_2 \| Q \|_2 + \| A_k \|_2  \| A - A_{k-1} \|_2 \| Q\|_2  + \cdots 
   \\
   & \quad + \| A_k A_{k-1} \cdots A_2 \|_2 \| A - A_1 \|_2  \| Q \|_2  + \|  A^{k+1}Q \|_2.
   \end{split}
\end{equation*}
 Letting $a_i = \|A - A_i \|_2$ and $b_i = \|  A_{i+1} \cdots A_k\|_2 \| Q\|_2$, and taking expectations on both sides of the above inequality, we obtain $\EE[ \|b_i \|_2^2] \leq \delta^{2(k-i)} \| Q \|_2$ and $\EE[\| A^{k+1}\|_2^2] \leq \delta^{2(k+1)}$. By using the fact that $\|AB \|_2 \leq \| A\|_2 \| B\|_2 \leq \frac{\|A \|_2^2}{2} + \frac{\|B \|_2^2}{2}$, we  further have that 
\begin{equation*}
    \begin{split}
      & \EE [ \|  (I-A)^{-1}   - Z_k \|_2^2 ] 
       \\
     & \leq \sum_{i=1}^k \EE[ \|  b_i a_i \|_2^2 ] +  \sum_{1 \leq i < j\leq k} 2\EE[ \| b_i  b_j a_i a_j \|_2]
 +   \sum_{1\leq i \leq k} 2 \EE [ \| a_ib_i A^{k+1}Q \|_2]  +  \EE [ \|  A^{k+1}Q  \|_2^2 ]
      \\
      & \leq  \sum_{i=1}^k \EE[ \|  b_i \|_2^2] \EE[ \| a_i \|_2^2 ] +  \sum_{1 \leq i < j\leq k} \EE[ \| b_i  b_j \|_2 ] \EE [ \| a_i \|_2^2 + \| a_j \|^2  ] 
      \\
      & \quad +   \sum_{1\leq i \leq k} 2 \delta^{2k+1-i} \EE[ \|  Q \|_2  \| a_i\|_2 ]
      +  \delta^{2(k+1)} \| Q  \|_2^2 
      \\
      & \leq \sum_{i=1}^k  \delta^{2(k - i)} \EE[\| a_i\|_2^2] +  \sum_{1 \leq i < j\leq k} \delta^{2k - i-j} \EE [ \| a_i \|_2^2 + \| a_j \|_2^2  ] 
      \\
      & \quad +   \sum_{1\leq i \leq k}  \delta^{2k+1-i} \EE[ \|  Q \|_2^2 +  \| a_i\|_2^2 ] + \delta^{2(k+1)} \| Q\|_2^2
      \\
      & =( \sum_{1  \leq  i\leq k } \delta^{k-i} + \delta^{k+1})  \Big ( \sum_{1 \leq j \leq k }  \delta^{k-j} \EE[ \| a_j\|_2^2]  \Big ) + ( \delta^{2(k+1)} +  \sum_{1\leq i \leq k}  \delta^{2k+1-i} )  \| Q \|_2^2 
      \\
      & \leq \frac{1}{1-\delta}\Big ( \sum_{1 \leq j \leq k }  \delta^{k-j} \EE[ \| a_j\|_2^2]  \Big ) + \frac{ \delta^{k+1}}{1-\delta}\| Q \|_2^2,
    \end{split} 
\end{equation*}
where the last inequality uses the fact that  $\sum_{i=0}^\infty \delta^i =\frac{1}{1-\delta}$. 
The desired inequality can be acquired by using $\| Q \|_2^2 \leq \frac{1}{(1-\delta)^2}$.

\QED

%\sg{How to formally describe the sequence $A_1,\cdots, A_k$ such that $\EE[ \| A_1 A_2 \cdots A_k\|^2] \leq \EE[\| A_1\|] \EE[\| A_2\|]\cdots \EE[\| A_k\|]$.}
% Consequently, we have 
% \begin{equation}
% \begin{split}
%   \EE [ \|  (I-A)^{-1} - Z_k \|^2 ]
%   & \leq  k \EE [ \| A- A_k \|^2 ] \| Z \|^2 + k \mu \EE [ \| A - A_{k-1} \|^2 ]  \| Z\|^2  + \cdots 
%   \\
%   & \quad + k \mu^k \EE [ \| A - A_1 \|^2 ]  + \EE [  \| A^{k+1} Z \|^2 ] 
%   \end{split}
% \end{equation}

We provide the following result to characterize the estimation error $\| [\nabla_{yy}^2 g(\bar x_t, \bar y_t) ]^{-1} - q_{t}^k\|_2^2$ induced by Algorithm~\ref{alg:1}.
\begin{lemma} \label{lemma:inverse_Hessian}
Suppose Assumptions~\ref{assumption:1}, \ref{assumption:2}, and \ref{assumption:W} hold, then we have 
  \begin{equation*}
    \EE[ \| [\nabla_{yy}^2 g(\bar x_t, y_t^* ) ]^{-1} - q_{t}^k\|_2^2 ] 
    \leq \frac{1}{L_g^4 \kappa_g  }   \Big ( \sum_{1 \leq j \leq  b}  (1-\kappa_g)^{b - j } \EE[ \| g(\bar x_t,  y_t^* ) -  v_{t,i}^k \|_F^2]  \Big ) + \frac{(1-\kappa_g)^{b+1}}{ L_g^2 \kappa_g^3}.
\end{equation*}  
\end{lemma}
\textit{Proof:}
Recall that each $v_{t,i}^k $ is the convex combination of $\mu_g I$ and sampled hessian $\nabla_{yy}^2 g^k(x,y;\xi^k)$, under Assumption~\ref{assumption:2} (v) that $\EE[\|  I - \frac{1}{L_g} \nabla_{yy}^2 g^k(x,y;\xi^k) \|_2^2]\leq (1- \kappa_g)^2$, we have  $ \EE[ \| I - \frac{1}{L_g}  v_{t,i}^k \|_2^2] \leq (1- \kappa_g )^2$.
By applying Lemma~\ref{lemma:inverse_diff} with $A = I - \frac{1}{L_g}  \nabla_{yy}^2 g(\bar x_t, y_t^* )$, $A_i = I - \frac{v_{t,i}^k }{L_g} $, and $\delta = 1-\kappa_g$, we obtain that 
 \begin{equation*}
    \begin{split}
   &   \EE [ \|  L_g [\nabla_{yy}^2 g(\bar x_t,  y_t^* )  ]^{-1} - Q_{t,b}^k \|_2^2 ] 
     \\
  & \leq \frac{1}{\kappa_g } \Big ( \sum_{1 \leq j \leq b}  \frac{ (1- \kappa_g)^{b-j}  }{L_g^2}\EE[ \|  \nabla_{yy}^2 g(\bar x_t,  y_t^* )  - v_{t,j}^k  \|_2^2]  \Big ) + \frac{(1- \kappa_g)^{b+1} }{\kappa_g^3}
  \\
  & \leq \frac{1}{\kappa_g } \Big ( \sum_{1 \leq j \leq b }  \frac{ (1- \kappa_g)^{b-j}  }{L_g^2}\EE[ \|  \nabla_{yy}^2 g(\bar x_t,  y_t^* )  - v_{t,j}^k  \|_F^2]  \Big ) + \frac{ (1- \kappa_g)^{b+1} }{\kappa_g^3}.
    \end{split} 
\end{equation*}
We obtain the desired result through dividing both sides of the above inequality  by $L_g^2$ and using the fact that $q_t^k = Q_{t,b}^k/L_g$.

\QED

%%%%%%%%%%%%%%%%%%%%
%%%%%%%%%%%%%%%%%%%%%
%%%%%%%%%%%%%%%%%%%%%

\section{Proof of Results for Nonconvex Objectives} \label{app:B}
Throughout this section, we assume Assumptions \ref{assumption:1} and \ref{assumption:2} hold and the step-sizes follow  \eqref{eq:stepsize} that
\begin{equation*}
\alpha_t = C_0 \sqrt{\tfrac{K}{T}}, \   \beta = \gamma_t = \sqrt{\tfrac{K}{T}},  \ \text{ and } b = 3 \lceil \log_{\frac{1}{1-\kappa_g}} T \rceil ,    \text{ for all }t=0,1,\cdots, T,
\end{equation*} 
where $C_0>0$ is a small constant and the number of iterations $T $ is large   such that $\beta_t ,\gamma_t \leq 1$ and 
\begin{equation*}
  \Upsilon(C_0,T) =  1 -\frac{C_0L_F}{\sqrt{T}}- \frac{7C_0^2  (L_f^2(m_1 + m_2)  + \widetilde L_g^2 (m_3 + m_4) )(1 + L_y^2 )}{2 }  - \frac{3 m_5 L_y^2 C_0^2 }{ \mu_g^2} \geq 0  , 
\end{equation*}
with 
\begin{equation}\label{eq:m}
\begin{split}
 m_1 =  4, \ \ m_2  & = 12L_g^2 L_q^2 ,  \ \ m_3 = 12 C_f^2 L_q^2,
 \\
 m_4 = 12 C_f^2L_g^{-2} \kappa_g^{-2}  , \ \text{ and } m_5 &  = 12 \big  ( L_f^2(m_1 + m_2) + \widetilde  L_g^2(m_3 + m_4)  \big ).    
 \end{split}
\end{equation}

% $$C_0 =  \left [\Big ( \tfrac{7 (1 + L_y^2 )}{2 }    +  \frac{36 L_y^2}{ \mu_g^2  } \Big ) \Big ( 24C_g^2 C_f^2 L_g^2   + 14 C_g^2 L_q^2 L_f^2 + 8L_f^2  + 24 C_f^2 L_q^2 L_g^2 \Big ) \right ]^{-1/2} > 0 $$ is a small constant 
% % $\tilde C =  \tfrac{7 (1 + L_y^2 )}{2 }    +  \frac{36 L_y^2}{ \mu_g^2  }$.
% % % By setting $C_0 =  \left [\tilde C \Big ( 24C_g^2 C_f^2 L_g^2   + 14 C_g^2 L_q^2 L_f^2 + 8L_f^2  + 24 C_f^2 L_q^2 L_g^2 \Big ) \right ]^{-1/2}$
%  and . 

% \begin{lemma}
% Let  $\kappa_g = L_g/\mu_g$ be the condition number of $g$ and $q=\text{HIA}(x,y, M ) $ be the estimator of $\nabla_{yy}^2 g(x,y)^{-1}$ returned by Algorithm \ref{alg:HIA}, then we have 
% \begin{equation*} 
% \begin{split}
%  \|  \nabla_{yy}^2 g(x,y)^{-1}  - \EE[q ]\|  \leq \tfrac{1}{\mu_g} (\tfrac{\kappa_g - 1}{\kappa_g})^M \text{ and }
%   \EE [ \|   q \|^2 ]  \leq  \sigma_q^2, \text{ for some } \sigma_q >0.
% \end{split}
% \end{equation*}
% \end{lemma}

\subsection{Lemma~\ref{lemma:boundedness} and Its Proof}

 \begin{lemma}\label{lemma:boundedness}
%\sg{add a high probability bound for this} 
Suppose Assumption \ref{assumption:1}, \ref{assumption:2}, and \ref{assumption:W} hold, then we have the followings.
\begin{equation}
\begin{split}
& \EE[ \| s_t^k\|^2] \leq C_f^2, \ \EE[\| h_{t}^k\|^2] \leq C_f^2, \EE[ \| u_{t}^k \|^2] \leq  L_g^2,  \EE[ \| q_t^k \|^2] \leq L_q^2, 
\\
& \EE[ \| v_{t,j}^k \|_F^2] \leq  L_g^2, \forall 1\leq j \leq b, 
    \text{ and } \ \EE[\| z_t^k\|^2]  \leq 2 C_f^2 + 2C_f^2 L_q^2 L_g^2. 
 \end{split}
\end{equation}
\end{lemma}

\textit{Proof:} 
We first observe that  $s_{t}^k, h_{t}^k,  u_{t}^k$, $v_{t,j}^k$ are convex combinations of past sampled stochastic information $ \nabla_x f^k(x_t^k, y_t^k ; \zeta_t^k )$, $ \nabla_y f^k(x_t^k, y_t^k ; \zeta_t^k )$, $ \nabla_{xy}^2 g^k(x_t^k, y_t^k ; \xi_t^k )$,  $ \nabla_{yy}^2 g^k(x_t^k, y_t^k ; \xi_t^k )$, respectively. Therefore, under Assumption \ref{assumption:1}, for all $t \leq T$, for all $ 1\leq j \leq b$, we have 
$$
\EE[ \| u_{t}^k \|^2] \leq  L_g^2, \ \EE[ \|  v_{t,j}^k \|^2] \leq  L_g^2,  \EE[ \| s_t^k\|^2] \leq C_f^2, \text{ and } \EE[\| h_{t}^k\|^2] \leq C_f^2.
$$
%Moreover, $u_t^k v_t^k h_t^k $ can also be treated as the convex combination of $\nabla_{xy}^2 g(x_j^s, y_j^s,\xi_i^t) q_t^k \nabla_y f(x_i^k, y_i^t, \zeta_i^k)$ for $0 \leq i,j \leq t-1$ and $1\leq s,k \leq K$. 
Recall that $q_t^k = \frac{1}{L_g}\sum_{i=0}^b \prod_{j=1}^i (  I - \tfrac{v_{t,j}^k}{L_g}  ) $, we further obtain that 
\begin{equation}
\begin{split}
 \EE[ \| q_t^k \|^2] & =  \frac{1}{L_g^2} \sum_{ 0 \leq i \leq b} \EE \left (  \prod_{j=1}^i(  I - \tfrac{v_{t,j}^k}{L_g}  )  \cdot  \sum_{ 0 \leq s \leq b} \prod_{j=1}^s(  I - \tfrac{v_{t,j}^k}{L_g}  )  \right ) 
\\
& \leq   \frac{1}{L_g^2} \sum_{ 0 \leq i \leq b}  \frac{ (1-\kappa_g)^{i} }{\kappa_g}   \leq \frac{1}{\kappa_g^2L_g^2} = L_q^2. 
\end{split}
\end{equation}
By using the conditional independence of the sampled stochastic information, we have $\EE[ \| u_t^k q_t^k h_t^k \|^2] \leq  C_f^2 L_q^2L_g^2  $, further implying that 
$$
\EE[\| z_t^k\|^2] = \EE[ \| s_t^k - u_t^k q_t^k h_t^k   \|^2 ] \leq 2 C_f^2 + 2C_f^2 L_q^2 L_g^2 .
$$
% Further, we observe that $\bar s_{t+1} = (1-\beta_t) \bar s_t + \beta_t \delta_t$ where $\delta_t =  \frac{1}{K} \sum_{k \in \cK} \nabla_y f^k(x_t^k,y_t^k;\xi_t^k)$.
% Here, $\bar s_t$ can  be viewed as the convex combination of $\delta_s$'s with weight being $\beta_0(1-\beta_0)^{t-s}$ and $\bar s_0 = \textbf{0}$ with weight being $(1-\beta_0)^t$. Because $\text{Var}(\delta_s) \leq \frac{C_f^2}{K}$, by using the fact that $\beta_0 = \sqrt{\frac{K}{T}}$, we have with  probability at least $1-\delta$, $\| \bar s_t \|^2 \leq $ is bounded such that \sg{ fill the bound. }

This completes the proof. 
\QED

%%%%%%%%%%%%%%
%%%%%%%%%%%%%%
%%%%%%%%%%%%%%%

\subsection{Lemma~\ref{lemma:consensus} and Its Proof} \label{app:consensus}
We  quantify the convergence behavior of consensus errors under the choices of step-sizes \eqref{eq:stepsize} and \eqref{eq:stepsize_PL} as follows. 
\begin{lemma}\label{lemma:consensus}
Suppose Assumption \ref{assumption:1}, \ref{assumption:2}, and \ref{assumption:W} hold and   the step-sizes satisfy $\beta_t\leq 1$ and one of the followings:
\begin{enumerate}
    \item[(i)] $\alpha_t = \alpha_0$, $\beta_t = \beta_0$, and $\gamma_t = \gamma_0$, for $0 \leq t \leq T$.
    \item[(ii)] $\lim_{t\to \infty} (\alpha_t + \beta_t + \gamma_t ) = 0$, $\lim_{t \to \infty} \frac{\alpha_{t-1}}{\alpha_t} = 1$, $\lim_{t \to \infty} \frac{\beta_{t-1}}{\beta_t} = 1$, and  $\lim_{t \to \infty} \frac{\gamma_{t-1}}{\gamma_t} = 1$. 
\end{enumerate}
Then we have for  all $1 \leq j \leq b$, 
\begin{equation*}
\begin{split}
 & \sum_{ k \in \cK }  \EE [ \| x_t^k - \bar x_t    \|^2  ] \leq \cO \left ( \frac{ K \alpha_t^2}{(1-\rho)^2} \right ),  \ 
  \sum_{ k \in \cK }\EE[ \| y_{t}^k - \bar y_t\|^2] \leq \cO \left ( \frac{ K \gamma_t^2}{(1-\rho)^2} \right ), 
  \\
 \text{ and } &\sum_{  k \in \cK }  \EE \left [ \| s_t^k - \bar s_t\|^2 +  \| u_t^k - \bar u_t\|^2    +   \| h_t^k - \bar h_t\|_F^2  +  \| v_{t,j}^k - \bar v_{t,j}\|_F^2  \right ]\leq \cO \left ( \frac{ K \beta_t^2}{(1-\rho)^2} \right ). 
\end{split}
\end{equation*}
\end{lemma}

\textit{Proof:}
Recall the update rule that 
 \begin{equation*}
 X_{t+1} = X_{t}W - \alpha_{t} Z_{t} \text{ and } \bar X_{t+1} =\bar X_{t} - \alpha_{t} \bar Z_{t},
 \end{equation*}
 by using the fact that $\bar X_t = X_t W^\infty$, 
 we have 
 \begin{equation*}
 X_{t+1} - \bar X_{t+1} =  X_{t}( W -W^\infty )  - \alpha_t (Z_t - \bar Z_t). 
 \end{equation*}
Under Assumption~\ref{assumption:W} that  $\| W - W^\infty \|_2 =   \sqrt{\rho}$, we have 
 \begin{equation*}
 \begin{split}
\| X_t  W- \bar X_t \|_F  & = \| (X_t - \bar X_t)(W - W^\infty)\|_F =   \| (W - W^\infty)^\top(X_t - \bar X_t)^\top\|_F 
\\
& \leq \| W - W^\infty\|_2 \| X_t - \bar X_t\|_F \leq \sqrt{\rho} \| X_t - \bar X_t\|_F,
\end{split}
\end{equation*}
where the first equality uses the fact that $\bar X_t W = \bar X_t = X_t W^\infty$. 
Consequently, by using the fact that $\| A + B\|_F^2 \leq (1+\eta) \| A\|_F^2 + (1 + \frac{1}{\eta}) \| B\|_F^2$ for $\eta > 0$, we have 
 \begin{equation*}
 \begin{split}
 \|  X_{t+1} - \bar X_{t+1}  \|_F^2 & \leq (1+ \eta)  \delta \| X_{t} - \bar X_{t} \|_F^2 + (1 + \tfrac{1}{\eta})  \alpha_t^2  \| Z_t - \bar Z_t\|_F^2.
 \end{split}
 \end{equation*}
By setting $\eta = \frac{1-\rho}{2\rho}$, we obtain
 \begin{equation*}
 \|  X_{t+1} - \bar X_{t+1}  \|_F^2 \leq  \frac{1+\rho}{2} \| X_{t} - \bar X_{t} \|_F^2 + \frac{(1+\rho) \alpha_t^2}{1-\rho}  \| Z_t - \bar Z_t\|_F^2.
 \end{equation*}
%  \begin{equation}
% \|  X_{t+1} - \bar X_{t+1}  \|_F \leq \sqrt{\rho}\| X_{t} - \bar X_{t} \|_F + \alpha_t \sqrt{\rho}  \| Z_t - \bar Z_t\|_F.
% \end{equation}
 Taking expectations on both sides of the above inequality and using Lemma~\ref{lemma:boundedness}  that $\| Z_t - \bar Z_t\|_F^2 \leq 4K(C_f^2 + C_f^2 L_q^2 C_g^2) $ and $1+\delta \leq 2$, we further have 
  \begin{equation*}
  \begin{split}
 \EE[ \|  X_{t+1} - \bar X_{t+1}  \|_F^2] &  \leq  \frac{1+\rho}{2} \EE[ \| X_{t} - \bar X_{t} \|_F^2 ] + \frac{2 \alpha_t^2}{1-\rho}\EE[ \| Z_t - \bar Z_t\|_F^2 ]
 \\
 & \leq  \frac{1+\rho}{2} \EE[ \| X_{t} - \bar X_{t} \|_F^2 ] + \frac{8 \alpha_t^2 K (C_f^2 + C_f^2 L_q^2 L_g^2) }{1-\rho}.
 \end{split}
 \end{equation*}
 We then use an induction argument to prove the result. Suppose $\EE[ \| X_{t} - \bar X_{t} \|_F^2 ]  \leq \hat C K \alpha_{t-1}^2$, then we have 
   \begin{equation*}
  \begin{split}
 \EE[ \|  X_{t+1} - \bar X_{t+1}  \|_F^2] & \leq  \frac{(1+\rho) \hat C K \alpha_{t-1}^2}{2}  + \frac{8 \alpha_t^2 K (C_f^2 + C_f^2 L_q^2 L_g^2) }{1-\rho} 
 \\
 & = \alpha_t^2 \Big ( \frac{(1+\rho) \hat C  K \alpha_{t-1}^2 }{2 \alpha_t^2}   +  \frac{8  K (C_f^2 + C_f^2 L_q^2 L_g^2)}{1-\rho}  \Big ) .
 \end{split}
 \end{equation*}
 We observe that $ \frac{(1+\rho)   \alpha_{t-1}^2 }{2 \alpha_t^2} =  \frac{1+\rho}{2}$ under condition (i). Under condition (ii) where $\lim_{t\to\infty} \frac{\alpha_{t-1}}{\alpha_t} = 1$, we can see that  $ \frac{(1+\rho)   \alpha_{t-1}^2 }{2 \alpha_t^2} \leq \frac{3+ \rho}{4}$  for $t$ sufficiently large.  Combining both scenarios, we observe that $ \Big ( \frac{(1+\rho) \hat C   \alpha_{t-1}^2 }{2 \alpha_t^2}   +  \frac{8   (C_f^2 + C_f^2 L_q^2 L_g^2)}{1-\rho}  \Big ) \leq \hat C$ for $\hat C =   \frac{32   (C_f^2 + C_f^2 L_q^2 L_g^2)}{(1-\rho)^2}   $. We then obtain that 
 \begin{equation*}
     \sum_{ k \in \cK }  \EE [ \| x_t^k - \bar x_t    \|^2  ] \leq \cO \left ( \frac{ K \alpha_t^2}{(1-\rho)^2} \right ). 
 \end{equation*}
 The analysis for   $  \sum_{ k \in \cK }\EE[ \| y_{t}^k - \bar y_t\|^2] $ is similar. To quantify  $  \sum_{ k \in \cK }\EE[ \| s_{t}^k - \bar s_t\|^2] $, we observe that a weight $1-\beta_t \leq 1$ is assigned to the prior value $s_t^k$, yielding that 
   \begin{equation*}
  \begin{split}
\sum_{ k \in \cK }\EE[ \| s_{t+1}^k - \bar s_{t+1}\|^2]
 & \leq  \frac{(1+\rho)(1-\beta_t)^2}{2 }  \sum_{ k \in \cK }\EE[ \| s_{t}^k - \bar s_t\|^2] + \frac{4 \beta_t^2 K C_f^2}{1-\rho}.
 \end{split}
 \end{equation*}
We acquire the desired result by following the analysis of quantifying $\sum_{ k \in \cK }  \EE [ \| x_t^k - \bar x_t    \|^2  ] $. 
\QED

\subsection{Lemma~\ref{lemma:nonconvex_main} and Its Proof}
We recall Algorithm~\ref{alg:1} Step 4 that $x_{t+1}^k = \sum_{j \in \cN_k}w_{k,j} x_t^j  - \alpha_t  \left ( s_t^k -  u_t^k  q_t^k   h_t^k \right ) $ and express $\bar x_{t+1}$ as follows. 
\begin{equation}\label{eq:bar_x_update}
\bar x_{t+1} = \bar x_t - \alpha_t \bar z_t, \text{ where }\bar z_t =\frac{1}{K} \sum_{k \in \cK} \left ( s_t^k -  u_t^k q_t^k   h_t^k \right ).
\end{equation}

   \begin{lemma}\label{lemma:nonconvex_main}
   Suppose Assumptions \ref{assumption:1}, \ref{assumption:2}, and \ref{assumption:W} hold.
   %and let $\{ (\bar x_t, y_t^*)\}$ be the sequence generated by Algorithm \ref{alg:1} with step-sizes defined in \eqref{eq:stepsize}. 
   We have that 
   \begin{equation}\label{eq:nonconvex_main}
\begin{split}  
& \EE[ \| \nabla F(\bar x_t)\|^2] 
\\
& \leq  \frac{2}{\alpha_t } \Big ( \EE[ F(\bar{x}_t)] -  \E[F(\bar{x}_{t+1}) ] \Big ) 
  -  (1 - \alpha_t L_F ) \EE[ \|  \bar z_t  
\|^2  ]  
 + 4  \EE [ \|  \nabla_x f( \bar x_t ,   y_t^*   ) - \bar s_t \|^2 ]  
 \\
 & \quad + 12 C_g^2 L_q^2  \EE[  \| \nabla_y f(  \bar x_t  ,   y^*_t    ) -  \bar h_t \|^2 ]  
 +12 C_f^2 L_q^2  \EE[ \|  \nabla_{xy}^2 g(  \bar x_t  ,   y_t^*   ) - \bar u_t \|_F^2 ]  
\\
& \quad  
+  \frac{12 C_f^2 }{L_g^2 \kappa_g  }   \sum_{1 \leq j \leq  b}  (1-\kappa_g)^{b - j } \EE[ \| \nabla_{yy}^2 g(\bar x_t, y_t^*  ) -  \bar v_{t,i} )\|_F^2] + \cO \left ( \frac{\beta_t^2}{(1-  \rho  )^2}\right )
. 
  \end{split}
\end{equation}

\end{lemma}

    \textit{Proof:}
    We start from the $L_F$-smoothness of $F(x)$ provided by Lemma~\ref{lemma:Lip}:
\begin{equation*}
\begin{aligned}
F(\bar{x}_{t+1})-F(\bar{x}_t) & \leq  \lv \nabla F(\bar{x}_t),  \bar x_{t+1} - \bar x_t \rv +    \frac{\alpha_t^2L_F}{2} \|\bar{z}_t\|^2 
\\
& \leq    -\alpha_t \lv \nabla F(\bar{x}_t), \bar{z}_t\rv +    \frac{\alpha_t^2L_F}{2} \|\bar{z}_t\|^2 .
\end{aligned}
\end{equation*}
By using the fact that $- 2\lv a, b \rv = - \|a\|^2 - \| b\|^2 + \| a-b\|^2$, we further obtain 
\begin{equation}\label{eq:nonconvex_1}
\begin{aligned}
&  F(\bar{x}_{t+1})  -F(\bar{x}_{t})\\
& \leq   - \frac{ \alpha_t}{2} \| \nabla F(\bar x_t)\|^2 - \frac{\alpha_t}{2}\|  \bar z_t  
\|^2  +  \frac{\alpha_t }{2} \| \nabla F(\bar x_t) - \bar z_t    \|^2  + \frac{\alpha_t^2 L_F}{2} \|\bar{z}_t\|^2.
\end{aligned}
\end{equation}
Consider the term $ \| \nabla F(\bar x_t) - \bar z_t     \|^2$, we denote by $y_t^* = y^*(\bar x_t)$ and obtain 
\begin{equation*}
\begin{split}
&  \| \nabla F(\bar x_t) - \bar z_t     \|^2
  =  \| \nabla F(\bar x_t)   - \frac{1}{K} \sum_{k \in \cK }   z_t^k \|^2
 \\
&  \leq \frac{2}{K} \sum_{ k \in \cK } \Big ( \|  \nabla_x f( \bar x_t , y_t^* ) - s_t^k \|^2 + \|  \nabla_{xy}^2 g(  \bar x_t  ,   y_t^*  ) [\nabla_{yy}^2 g( \bar x_t  ,   y_t^*   ) ]^{-1}\nabla_y f(  \bar x_t  ,  y_t^* ) - u_t^k [ v_t^k]^{-1}  h_t^k \|^2  \Big ) 
\\
&  \leq \frac{2}{K} \sum_{ k \in \cK } \Big ( \|  \nabla_x f( \bar x_t , y_t^*   ) - s_t^k \|^2 +3 \| \nabla_y f(  \bar x_t  ,  \bar y_t  ) \|^2 \| [\nabla_{yy}^2 g( \bar x_t  ,    y_t^*  ) ]^{-1} \|^2 \|  \nabla_{xy}^2 g(  \bar x_t  ,   y_t^*  ) - u_t^k \|^2 \Big ) 
\\ 
& \quad +  \frac{6}{K} \sum_{ k \in \cK } \big (  \| u_t^k\|^2  \| \nabla_y f(  \bar x_t  ,  y_t^*    ) \|^2 \| [\nabla_{yy}^2 g( \bar x_t  ,  y_t^*    ) ]^{-1} - q_t^k \|^2 \big ) 
\\
& \quad +  \frac{6}{K} \sum_{ k \in \cK } \big (    \| u_t^k\|^2 \| q_t^k \|^2 \| \nabla_y f(  \bar x_t  , y_t^*   ) -  h_t^k \|^2  \big ).
\end{split}
\end{equation*}
Taking expectations on both sides of the above inequality and applying Lemma~\ref{lemma:boundedness}, we have that 
\begin{equation*}
\begin{split}
& \EE[ \| \nabla F(\bar x_t) - \bar z_t     \|^2]
\\
&  \leq  \frac{2}{K} \sum_{ k \in \cK }  \Big (\EE[  \|  \nabla_x f( \bar x_t , y_t^*   ) - s_t^k \|^2] +3 C_f^2 L_q^2 \EE[  \|  \nabla_{xy}^2 g(  \bar x_t  ,   y_t^*  ) - u_t^k \|^2] \Big ) 
\\ 
& \quad +  \frac{6}{K} \sum_{ k \in \cK }  \Big ( L_g^2 C_f^2 \EE[ \| [\nabla_{yy}^2 g( \bar x_t  ,  y_t^*    ) ]^{-1} - q_t^k \|^2 ] + L_g^2 L_q^2 \EE[ \| \nabla_y f(  \bar x_t  , y_t^*   ) -  h_t^k \|^2]  \Big ).
\end{split}
\end{equation*}
Taking expectations on both sides of \eqref{eq:nonconvex_1}, combining with the above inequality, dividing both sides by $\frac{\alpha_t}{2}$, and rearranging the terms, we obtain 
\begin{equation}\label{eq:L_F_smooth}
\begin{aligned}
&   \EE [ \| \nabla F(\bar x_t)\|^2 ] \
\\
& \leq   \frac{2}{\alpha_t} \Big ( \EE[ F(\bar{x}_{t+1}) ] - \EE [ F(\bar{x}_{t}) ] \Big ) - (1-\alpha_t L_F) \EE [ \|  \bar z_t  
\|^2  ] \\
& \quad + \frac{2}{K} \sum_{ k \in \cK }  \Big ( \EE [ \|  \nabla_x f( \bar x_t ,   y_t^*   ) - s_t^k \|^2 ] +3 C_f^2 L_q^2  \EE[ \|  \nabla_{xy}^2 g(  \bar x_t  ,   y_t^*  ) - u_t^k \|^2 ] \Big ) 
\\ 
& \quad +  \frac{6}{K} \sum_{ k \in \cK }  \Big ( L_g^2 C_f^2  \EE [ \| [\nabla_{yy}^2 g( \bar x_t  ,    y_t^*   ) ]^{-1} - q_t^k \|^2 ]+  L_g^2 L_q^2 \EE[  \| \nabla_y f(  \bar x_t  ,   y_t^*  ) -  h_t^k \|^2 ] \Big )  . 
\end{aligned}
\end{equation}

%$\|  \nabla_x f( \bar x_t ,   y_t^*   ) - s_t^k \|^2 \leq 2 \|  \nabla_x f( \bar x_t ,   y_t^*   ) - \bar s_t \|^2 +2 \| \bar s_t^k - \bar s_t\|^2$,  
By applying Lemma~\ref{lemma:inverse_Hessian} with $(1-\kappa_g)^{b+1} \leq \cO(\frac{1}{T^3})$ for $b = 3 \lceil  \log_{\frac{1}{1-\kappa_g}}(T) \rceil $,
we conclude
\begin{equation}\label{eq:nonconvex_03}
\begin{split}
& \EE[ \| \nabla F(\bar x_t)\|^2] 
\\
& \leq  \frac{2}{\alpha_t } \Big ( \EE[ F(\bar{x}_t)] -  \E[F(\bar{x}_{t+1}) ] \Big ) 
  -  (1 - \alpha_t L_F ) \EE[ \|  \bar z_t   
\|^2  ]  + \tfrac{ 2}{K} \sum_{k \in \cK}  \EE [ \|  \nabla_x f( \bar x_t ,   y_t^*   ) -  s_t^k \|^2 ]  
\\
& \quad +\frac{6}{K} \sum_{k \in \cK}  \Big  (   L_g^2 L_q^2  \EE[  \| \nabla_y f(  \bar x_t  ,   y_t^*   ) -   h_t^k \|^2 ]  +  C_f^2 L_q^2  \EE[ \|  \nabla_{xy}^2 g(  \bar x_t  ,   y_t^*   ) -  u_t^k \|^2 ]  \Big )
\\ 
& \quad + \frac{6}{K} \sum_{k \in \cK} \Big [  \frac{L_g^2 C_f^2 }{L_g^4 \kappa_g  }   \Big ( \sum_{1 \leq j \leq  b}  (1-\kappa_g)^{b - j } \EE[ \| \nabla_{yy}^2 g(\bar x_t, y_t^*  ) -  v_{t,i}^k )\|_F^2]  \Big )  + \cO \Big (\frac{1}{T^3} \Big) \Big ] 
\\
& \leq \frac{2}{\alpha_t } \Big ( \EE[ F(\bar{x}_t)] -  \E[F(\bar{x}_{t+1}) ] \Big ) 
  -  (1 - \alpha_t L_F ) \EE[ \|  \bar z_t  
\|^2  ]  
 + 4  \EE [ \|  \nabla_x f( \bar x_t ,   y_t^*   ) - \bar s_t \|^2 ]  
 \\
 & \quad + 12 L_g^2 L_q^2  \EE[  \| \nabla_y f(  \bar x_t  ,   y^*_t    ) -  \bar h_t \|^2 ]  
 +12 C_f^2 L_q^2  \EE[ \|  \nabla_{xy}^2 g(  \bar x_t  ,   y_t^*   ) - \bar u_t \|_F^2 ]  
\\
& \quad  
+  \frac{12 C_f^2 }{L_g^2 \kappa_g  }   \sum_{1 \leq j \leq  b}  (1-\kappa_g)^{b - j } \EE[ \| \nabla_{yy}^2 g(\bar x_t, y_t^*  ) -  \bar v_{t,i} )\|_F^2] + \cO \left ( \frac{\beta_t^2}{(1-  \rho  )^2}\right ),
  \end{split}
\end{equation}
where the last inequality uses the facts that $\| a+b\| \leq 2\| a\|^2 + 2 \| b\|^2$ and $\| A \|_2 \leq \| A\|_F$ for any matrix $A$, and applies the convergence of consensus errors characterized by Lemma~\ref{lemma:consensus} that 
$$\sum_{k \in \cK} \EE \Big ( \| s_t^k - \bar s_t\|^2 + \| u_t^k - \bar  u_t\|_F^2 + \| v_{t,j}^k - \bar v_{t,j}\|_F^2 + \| h_t^k - \bar  h_t\|^2  \Big ) \leq \cO \Big (\frac{K\beta_t^2}{(1-\rho)^2} \Big). $$ 
This completes the proof. 
\QED

%%%%%%%%%%%%%%
%%%%%%%%%%%%%%
%%%%%%%%%%%%%%%

%%%%%%%%%%%%%%%
%%%%%%%%%%%%%%%%
%%%%%%%%%%%%%%%%

\subsection{Lemma~\ref{lemma:y_t} and Its Proof}
Recall Algorithm~\ref{alg:1} Step 5 that $y_{t+1}^k = \sum_{ j \in \cN_k}y_t^j - \gamma_t \nabla_y g^k (x_t^k,y_t^k; \xi_t^k )$, we may express $\bar y_{t+1}$ as 
$$
 \bar y_{t+1} = \bar y_t -  \frac{\gamma_t}{K}\sum_{k \in \cK } \nabla_y g^k (x_t^k,y_t^k; \xi_t^k ) . 
$$
\begin{lemma}\label{lemma:y_t}
Suppose Assumptions~\ref{assumption:1}, \ref{assumption:2}, and \ref{assumption:W} hold
%Let $\{ (\bar x_t,\bar y_t) \}$ be the solution generated by Alg.~\ref{alg:1} with step-sizes defined in \eqref{eq:stepsize}. 
and $T$ is sufficiently large, we have 
\begin{equation}\label{eq:yt_recursion}
\begin{split}
& \EE [ \| \bar y_{t+1} - y^*(\bar x_{t+1})\|^2 ] 
\\
&
\leq ( 1 - \gamma_t \mu_g)  \EE [ \| \bar y_{t} -  y^*(\bar x_{t})\|^2 ]  + \cO\left (\frac{ \gamma_t \alpha_t^2 }{(1- \rho )^2 } \right ) + \frac{2\gamma_t^2 \sigma_g^2 }{K}  + \frac{3L_y^2\alpha_t^2 }{\gamma_t \mu_g } \EE [\| \bar z_t \|^2 ]  .
\end{split}
\end{equation}
\end{lemma}
\textit{Proof:} We first decompose the estimation error  $\| \bar y_{t} - y^*(\bar x_t) \|^2$ as 
\begin{equation}\label{eq:yk_00}
\| \bar y_{t+1} -  y^*(\bar x_{t+1})\|^2 \leq \left ( 1+ \frac{\gamma_t \mu_g}{2} \right ) \| \bar y_{t+1} -  y^*(\bar x_{t})\|^2+ \left (1 + \frac{2}{\gamma_t \mu_g } \right )  \| y^*(\bar x_{t}) - y^*(\bar x_{t+1}) \|^2.
\end{equation}

Recall that  $\bar y_{t+1} = \bar y_t -  \frac{\gamma_t}{K}\sum_{k \in \cK } \nabla_y g^k (x_t^k,y_t^k;\xi_t^k )  $. Letting $\delta_t = \nabla_y g(\bar x_t, \bar y_t) - \frac{1}{K}\sum_{k \in \cK } \nabla_y g^k (x_t^k,y_t^k;\xi_t^k )  $, we obtain 
\begin{equation}\label{eq:yk_01}
\begin{split}
& \| \bar y_{t+1} - y^*(\bar x_{t}) \|^2 =  \|  \bar y_t -  \gamma_t \nabla_y g(\bar x_t, \bar y_t) - y^*(\bar x_{t}) + \gamma_t \delta_t  \|^2 
\\
& = \|  \bar y_t -  \gamma_t \nabla_y g(\bar x_t, \bar y_t) - y^*(\bar x_{t})  \|^2  + \gamma_t  \lv \bar y_t -  \gamma_t \nabla_y g(\bar x_t, \bar y_t) - y^*(\bar x_{t}) ,  \delta_t  \rv + \gamma_t^2   \| \delta_t \|^2. 
\end{split}
\end{equation}
We then provide bounds for the above terms. First, consider $ \|  \bar y_t -  \gamma_t \nabla_y g(\bar x_t, \bar y_t) - y^*(\bar x_{t})  \|^2$, by using the $\mu_g$-strong convexity of $g(\bar x_t,y)$ in $y$ under Assumption~\ref{assumption:2} (i), 
we have 
\begin{equation}\label{eq:yk_02}
\begin{split}
& \|  \bar y_t -  \gamma_t \nabla_y g(\bar x_t, \bar y_t) - y^*(\bar x_{t})  \|^2 
\\
 & \leq \| \bar y_t - y^*(x_t)\|^2  - 2\gamma_t  \lv \bar y_t - y^*(x_t), \nabla_y g(\bar x_t, \bar y_t)  \rv + \gamma_t^2 \|  \nabla_y g(\bar x_t, \bar y_t)  \|^2
 \\
 & \leq (1- 2\gamma_t  \mu_g)\| \bar y_t - y^*(x_t)\|^2 + \gamma_t^2 C_g^2. 
 \end{split}
\end{equation}
Next, consider $ \lv \bar y_t -  \gamma_t \nabla_y g(\bar x_t, \bar y_t) - y^*(\bar x_{t}) ,  \delta_t \rv  $, we can see that 
\begin{equation}\label{eq:yk_03}
\begin{split}
&  \EE \left [  \lv \bar y_t -  \gamma_t \nabla_y g(\bar x_t, \bar y_t) - y^*(\bar x_{t}) ,  \delta_t  \rv   \right  ]
\\
& =  \EE \left [  \Big (  \bar y_t -  \gamma_t \nabla_y g(\bar x_t, \bar y_t) - y^*(\bar x_{t})  \Big )^\top  \Big (  \nabla_y g(\bar x_t, \bar y_t) - \frac{1}{K}\sum_{k=1}^k \nabla_y g^k (x_t^k,y_t^k)  \Big ) \right   ]
\\
& \leq   \frac{ \mu_g}{2} \EE[ \|  \bar y_t -  \gamma_t \nabla_y g(\bar x_t, \bar y_t) - y^*(\bar x_{t})   \|^2 ]
+ \frac{ \Delta_t}{2\mu_g} ,
 \end{split}
\end{equation}
where $\Delta_t  = \EE[ \| \nabla_y g(\bar x_t, \bar y_t) - \frac{1}{K}\sum_{k \in \cK} \nabla_y g^k (x_t^k,y_t^k)  \|^2]$ and the last inequality comes from the fact that $\lv a, b \rv \leq \frac{\| a\|^2}{2} + \frac{\| b \|^2}{2}$. 
Further, for  $\| \delta_t^k\|^2$, we have 
\begin{equation}\label{eq:yk_04}
\begin{split}
& \EE[ \|  \delta_t^k\|^2 ]  = \EE \left [ \Big \| \frac{1}{K} \sum_{k \in \cK} \big  ( \nabla_y g^k (\bar x_t, \bar y_t) - \nabla_y g^k (x_t^k,y_t^k;\xi_t^k )  \big )   \Big \|^2 \right ]
\\
& \leq 2 \EE \left [ \Big \| \frac{1}{K} \sum_{k \in \cK} \big  ( \nabla_y g^k (x_t^k , y_t^k ) - \nabla_y g^k (x_t^k,y_t^k;\xi_t^k )  \big )   \Big \|^2 \right ] + 2\Delta_t^k \leq \frac{2 \sigma_g^2}{K} + 2 \Delta_t^k,
 %+ \frac{2}{K} \sum_{k=1}^K L_g^2 \big ( \| x_t^k - \bar x_t\|^2 + \| y_t^k - \bar y_t\|^2 \big )
\end{split}
\end{equation}
where the last inequality uses the fact that $\{ \nabla_y g^k (x_t^k , y_t^k ) - \nabla_y g^k (x_t^k,y_t^k;\xi_t^k )  \}$'s are conditionally mean-zero and independent such that  for $k \neq s$, 
$$
\EE \left [ \lv \nabla_y g^k (\bar x_t, \bar y_t) - \nabla_y g^k (x_t^k,y_t^k;\xi_t^k ),  \nabla_y g^s (\bar x_t, \bar y_t) - \nabla_y g^s (x_t^s,y_t^s;\xi_t^s )  \rv \right ] = 0. 
$$ 
Taking expectations on both sides of \eqref{eq:yk_01} and combining with \eqref{eq:yk_02}, \eqref{eq:yk_03}, and \eqref{eq:yk_04}, we have 
\begin{equation}\label{eq:yk_05}
\begin{split}
 & \EE [ \| \bar y_{t+1} - y^*(\bar x_{t}) \|^2 ]
 \\
& \leq (1  +  \frac{\gamma_t \mu_g}{2}) \Big (  (1- 2\gamma_t  \mu_g)\EE[ \| \bar y_t - y^*(x_t)\|^2 ] + \gamma_t^2 C_g^2 \Big )
+ \left ( \frac{ \gamma_k}{2\mu_g} +  2\gamma_t^2 \right ) \Delta_t  + \frac{2\gamma_t^2 \sigma_g^2}{K}
\\
& \leq  (1- \frac{3\gamma_t \mu_g}{2})\EE[\| \bar y_t - y^*(x_t)\|^2] + (1 + \gamma_t \mu_g)\gamma_t^2 C_g^2 
\\
& \quad + \left ( \frac{ \gamma_k}{2\mu_g} +  2\gamma_t^2 \right ) \frac{1}{K} \sum_{k \in \cK} L_g^2 \EE \big ( \| x_t^k - \bar x_t\|^2 + \| y_t^k - \bar y_t\|^2 \big ) + \frac{2\gamma_t^2 \sigma_g^2}{K},
\end{split}
\end{equation}
where the second inequality uses the $L_g$-smoothness of $\nabla_y g(x,y)$ in both $x$ and $y$ such that
\begin{equation}\label{eq:Delta_t}
    \begin{split}
    \Delta_t  & = \EE[ \| \nabla_y g(\bar x_t, \bar y_t) - \frac{1}{K}\sum_{k \in \cK}  \nabla_y g^k (x_t^k,y_t^k)  \|^2] 
    \\
    & \leq \frac{1}{K} \sum_{k \in \cK} \EE[ \| \nabla_y g^k  (\bar x_t, \bar y_t) - \nabla_y g^k  (x_t^k,y_t^k) \|^2]
    \\
 & 
 \leq \frac{1}{K} \sum_{k \in \cK}  L_g^2 \big ( \EE[ \| x_t^k - \bar x_t\|^2] + \EE[ \| y_t^k - \bar y_t\|^2 ]\big ).    
    \end{split}
\end{equation}

Moreover, by using the $L_y$-smoothness of $y^*(\bullet)$ characterized by Lemma~\ref{lemma:Lip}, we have 
\begin{equation}\label{eq:yk_06}
\EE [ \| y^*(\bar x_{t}) - y^*(\bar x_{t+1}) \|^2 ]\leq L_y^2 \EE [\|  \bar x_{t} - \bar x_{t+1} \|^2 ].
\end{equation}
Finally, by substituting \eqref{eq:yk_05} and \eqref{eq:yk_06} into \eqref{eq:yk_00} and applying the bounds of the consensus errors provided by Lemma~\ref{lemma:consensus}, we conclude that 
\begin{equation*}
\begin{split}
& \EE [ \| \bar y_{t+1} -  y^*(\bar x_{t+1})\|^2  ]
 \\
& \leq  ( 1 - \gamma_t \mu_g) \EE[  \| \bar y_{t} -  y^*(\bar x_{t})\|^2 ]
+ 
(1 + \gamma_t\mu_g/2)  (1 + \gamma_t \mu_g)\gamma_t^2 C_g^2 
\\
& \quad + \left  ( \frac{ \gamma_k}{2\mu_g} +  2\gamma_t^2 \right ) \frac{1}{K} \sum_{k \in \cK} L_g^2 \big ( \EE[ \| x_t^k - \bar x_t\|^2] +  \EE [ \| y_t^k - \bar y_t\|^2] \big ) \\
& \quad + \frac{2\gamma_t^2 \sigma_g^2}{K}
 + \left (1 + \frac{2}{\gamma_t \mu_g } \right ) L_y^2 \EE [\|  \bar x_{t} - \bar x_{t+1} \|^2 ] 
\\
& \leq  ( 1 - \gamma_t \mu_g)  \EE [ \| \bar y_{t} -  y^*(\bar x_{t})\|^2 ]  + \cO(\gamma_t^3 ) + \cO\left (\frac{ \gamma_t \alpha_t^2 }{(1- \rho)^2 } \right ) + \frac{2\gamma_t^2 \sigma_g^2}{K} + \frac{3L_y^2\alpha_t^2 }{\gamma_t \mu_g } \EE [\| \bar z_t \|^2 ] ,
 \end{split}
\end{equation*}
where the last inequality uses the facts that $\bar x_{t+1} = \bar x_t - \alpha_t \bar z_t$ and $\gamma_t\mu_g \leq 1$ for large $T$. 
This completes the proof. 

\QED

%%%%%%%%%%%%%%
%%%%%%%%%%%%%%%

%%%%%%%%%%%%%%%%%%%
%%%%%%%%%%%%%%%%%
\subsection{Lemma \ref{lemma:s_t} and Its Proof}
By recalling Algorithm \ref{alg:1} Step 6 that $s_{t+1}^k   =(1-\beta_t) \sum_{j \in \cN_k} w_{kj} s_t^j + \beta_t  \nabla_x f^k(x_t^k, y_t^k ; \zeta_t^k )$, we express the update rule of $\bar s_t = \frac{1}{K} \sum_{k \in \cK} s_t^k$ as 
\begin{equation}\label{def:update_bar_st} 
\bar s_{t+1} =(1-\beta_t) \bar s_t + \frac{\beta_t}{K} \sum_{k \in \cK}  \nabla_x f^k(x_t^k, y_t^k ; \zeta_t^k ).
\end{equation}

\begin{lemma}\label{lemma:s_t}
Suppose Assumptions \ref{assumption:1},  \ref{assumption:2} , and  \ref{assumption:W} hold and $T$ is sufficiently large, then  we have 
\begin{equation}\label{eq:s_t}
    \begin{split}
    &    \EE[ \|  \bar s_{t+1} - \nabla_x f( \bar x_{t+1}, y^*(\bar x_{t+1}) ) \|^2]  
        \\
              & \leq (1-\beta_t)\EE [ \| \bar s_t - \nabla f (   \bar x_{t}, y_{t}^* ) \|^2] + \frac{2\beta_t^2   C_f^2}{K }    + \frac{7\alpha_t^2  L_f^2 (1 + L_y^2 )}{2\beta_t }  \EE[  \| \bar z_t \|^2  ] 
              \\
              & \quad +    \cO \left ( \frac{ K\beta_t ( \alpha_t^2 +  \beta_t^2) }{(1- \rho)^2 } \right )  + 12\beta_t L_f^2  \EE[ \| \bar y_t - y_t^*\|^2 ]. 
        \end{split}
\end{equation}
\end{lemma}

\textit{Proof:}
We denote by $y_t^* = y^*(\bar x_t)$
%$w_t^k = (x_t^k, y_t^k)$ and $\bar w_t = \frac{1}{K} \sum_{ k \in \cK} w_t^k$, $ \nabla_x f^k(w_t^k ; \zeta_t^k )= \nabla_x f^k(x_t^k, y_t^k ; \zeta_t^k ) $, and $\nabla_x f^k(w_t^k  )= \nabla_x f^k(x_t^k, y_t^k  ) $ 
for notational convenience. 
Consider the update rule of $\bar s_{t+1}$, we have 
\begin{equation*}
\begin{split}
 \bar s_{t+1} - \nabla_x f(\bar x_{t+1}, y_{t+1}^*)  
 = (1-\beta_t) [\bar s_t  -\nabla_x f(\bar x_{t+1}, y_{t+1}^*)   ]  + \beta_t \Delta_{f,t},
\end{split}
\end{equation*}
where 
$$
\Delta_{f,t}   =  \frac{1}{K}\sum_{k \in \cK}  \nabla_x f^k(x_t^k, y_t^k ; \zeta_t^k )  - \nabla_x f(\bar x_{t+1},  y_{t+1}^* ) .
$$
We can see that 
\begin{equation}\label{eq:st_1}
    \begin{split}
       & \EE[ \|   \bar s_{t+1} - \nabla_x f(\bar x_{t+1}, y_{t+1}^*)   \|^2] 
       \\
 & = (1-\beta_t)^2 \EE [ \| \bar s_t -  \nabla_x f(\bar x_{t+1}, y_{t+1}^* )  \|^2] + \beta_t^2 \EE [ \|  \Delta_{f,t}  \|^2] \\
       & \quad + 2(1-\beta_t)\beta_t \EE \left  [ (\bar s_t - \nabla_x f(\bar x_{t+1}, y_{t+1}^*  )  )^\top  \Delta_{f,t}   \right ]
    \\
   &  = (1-\beta_t)^2 \EE [ \| \bar s_t -  \nabla_x f(\bar x_{t+1}, y_{t+1}^* )  \|^2] + \beta_t^2 \EE [ \|  \Delta_{f,t}  \|^2] \\
       & \quad + 2(1-\beta_t)\beta_t \EE \left  [ (\bar s_t - \nabla_x f(\bar x_{t+1}, y_{t+1}^*  )  )^\top  \Big  ( \frac{1}{K}\sum_{k \in \cK}  \nabla_x f^k(x_t^k, y_t^k )  - \nabla_x f(\bar x_{t+1},  y_{t+1}^* ) \Big )  \right ]
    \\
    & \leq (1-\beta_t)^2 \EE [ \| \bar s_t -  \nabla_x f( \bar x_{t+1}, y_{t+1}^* )  \|^2  ] + \beta_t^2 \EE [ \|  \Delta_{f,t}     \|^2] 
    \\
       & \quad + \frac{(1-\beta_t)\beta_t }{2} \Big ( \EE[ \| \bar s_t - \nabla_x f(\bar x_{t+1}, y_{t+1}^*  )   \|^2 +4  \EE[ \|    \tfrac{1}{K}\sum_{k \in \cK}  \nabla_x f^k(x_t^k, y_t^k)  - \nabla_x f(\bar x_{t+1}, y_{t+1}^*  )   \|^2 ]  \Big)
       \\
       & \leq (1-\beta_t) \left (1- \frac{\beta_t}{2} \right ) \EE[ \| \bar s_t - \nabla_x f(\bar x_{t+1}, y_{t+1}^*  )   \|^2 ] + \beta_t^2  \EE [ \|  \Delta_{f,t}  \|^2]  \\
       & \quad  + \frac{2\beta_t(1-\beta_t)  L_f^2 }{K}  \sum_{k \in \cK } \left ( \EE[ \|   x_{t}^k - \bar x_{t+1} \|^2 ] + \EE[ \|   y_{t}^k -  y_{t+1}^*  \|^2 ]  \right ) ,
    \end{split}
\end{equation}
where the first equality uses the conditional independence between $\{ \nabla_x f^k(x_t^k, y_t^k ; \zeta_t^k )  - \nabla_x f^k(x_t^k, y_t^k ) \} $ and $\bar x_{t+1},\bar y_{t+1}, \bar s_t$,  the first inequality uses the fact that $2\lv a,b\rv \leq \frac{\|a\|^2}{2} + 2 \| b\|^2 $, and the last inequality applies similar analysis as \eqref{eq:Delta_t} under the $L_f$-smoothness of $\nabla_x f^k$. 
%second inequality uses the conditional independence between $\{ \nabla_x f^k(x_t^k, y_t^k ; \zeta_t^k )  - \nabla_x f^k(x_t^k, y_t^k ) \} $ and $\bar x_{t+1},\bar y_{t+1}, \bar s_t$. 
We observe that 
\begin{equation*}
\begin{split}
&  \| \Delta_{f,t} \|^2 
\\
&  =  \Big \| 
\frac{1}{K}\sum_{k \in \cK}  \left ( \nabla_x f^k(x_t^k, y_t^k ; \zeta_t^k )  - \nabla_x f^k(x_t^k, y_t^k)  \right )
 +  \frac{1}{K} \sum_{k\in \cK}( \nabla_x f^k ( x_t^k, y_t^k) - \nabla_x f^k(\bar x_{t+1}, y_{t+1}^* ) 
 )  \Big \|^2 
 \\
 & \leq  2 \Big \|  \frac{1}{K}\sum_{k \in \cK}  \left ( \nabla_x f^k(x_t^k, y_t^k ; \zeta_t^k )  - \nabla_x f^k(x_t^k, y_t^k)  \right )  \Big \|^2 
 \\
 & \quad +  2\Big \|  \frac{1}{K} \sum_{k\in \cK}( \nabla_x f^k ( x_t^k, y_t^k) - \nabla_x f^k(\bar x_{t+1},  y_{t+1}^* ) 
 )  \Big \|^2. 
 \end{split}
\end{equation*}
By noting that $\nabla_x f^k(x_t^k, y_t^k ;  \zeta_t^k )  - \nabla_x f^k(x_t^k, y_t^k)$ is conditionally mean-zero and 
$$\EE \Big [ \big (  \nabla_x f^i(x_t^i, y_t^i ; \zeta_t^i )  - \nabla_x f^i(x_t^i, y_t^i)  \big )^\top \big ( \nabla_x f^j( x_t^j, y_t^j ; \zeta_t^j)  - \nabla_x f ^j(x_t^i, y_t^i) \big ) \Big ] = 0, \text{ for } 1\leq i \neq j \leq K, 
$$
and taking expectations on both sides of the above inequality, we obtain 
\begin{equation}\label{eq:st_2}
\begin{split}
\EE[ \| \Delta_{f,t} \|^2 ]
 & \leq    \frac{2}{K^2}  \sum_{k \in \cK} \EE[ \| \nabla_x f^k(x_t^k, y_t^k ; \zeta_t^k )  - \nabla_x f^k(x_t^k, y_t^k)  \|^2 ] 
 \\
 & \quad +  \frac{2}{K} \sum_{k \in \cK} \EE[ \|  \nabla_x f^k ( x_t^k, y_t^k) - \nabla_x f^k(\bar x_{t+1},  y_{t+1}^* ) 
 \|^2 ] 
 \\
 & \leq \frac{2\sigma_f^2}{K} +  \frac{2 L_f^2 }{K} \sum_{k\in \cK}  \left ( \EE[ \| x_{t}^k -  \bar x_{t+1}  \|^2 ] 
 + \EE[ \| y_{t}^k -  y_{t+1}^*   \|^2 ] \right ),
 \end{split}
\end{equation}
where the last inequality uses the $L_f$-smoothness of $\nabla_x f^k(x,y)$ in both $x$ and $y$, similar as  \eqref{eq:Delta_t}.

Further, we have 
\begin{equation*}
\begin{split}
 & \| \bar s_t - \nabla_x f(\bar x_{t+1},  y_{t+1}^*  )   \|^2
 \\
 & \leq \left ( 1 +  \frac{\beta_t}{3} \right )  \| \bar s_t - \nabla_x f(\bar x_{t}, y_{t}^* )   \|^2  + \left (1 + \frac{3}{\beta_t}  \right ) \| \nabla_x f(\bar x_{t},  y_{t}^* )   -\nabla_x f(\bar x_{t+1}, y_{t+1}^* ) \|^2 
     \\
  & \leq  \left ( 1 +  \frac{\beta_t}{3} \right ) \| \bar s_t - \nabla_x f ( \bar x_{t}, y_{t}^*  )  \|^2 + \left (1 + \frac{3}{\beta_t}  \right ) L_f^2 (  \| \bar x_t - \bar x_{t+1}\|^2  +  \| y_t^*  -  y_{t+1}^* \|^2 )
  \\
  & =  \left ( 1 +  \frac{\beta_t}{3} \right ) \|\bar s_t - \nabla_x f ( \bar x_{t}, y_{t}^*   )  \|^2 + \left  (1 + \frac{3}{\beta_t}  \right )\alpha_t^2 L_f^2 (1+ L_y^2)\| \bar z_t \|^2,
    \end{split}
\end{equation*}
where the first inequality uses the $L_f$-smoothness of $\nabla_x f(x,y)$ and the second inequality uses the $L_y$-Lipschitz continuity of $y^*(x)$ such that $\| y^*(\bar x_t) - y^*(\bar x_{t+1}) \|  \leq L_y \|\bar x_t - \bar x_{t+1} \|$ and the fact that $ \alpha_t \| \bar z_t\| = \| \bar x_t - \bar x_{t+1} \|$.

By substituting  the above inequality and \eqref{eq:st_2} into \eqref{eq:st_1}, we obtain 
\begin{equation*}
    \begin{split}
       &  \EE[ \|  \bar s_{t+1} - \nabla f_x( \bar x_{t+1}, y_{t+1}^* ) \|^2]  
       \\
              & \leq (1-\beta_t)\EE [ \| \bar s_t - \nabla f (   \bar x_{t}, y_{t}^* ) \|^2] + \frac{2\beta_t^2   C_f^2}{K }    + \frac{7\alpha_t^2  L_f^2 (1 + L_y^2 )}{2\beta_t }  \EE[  \| \bar z_t \|^2  ] 
              \\
              & \quad + \frac{2\beta_t \sigma_g^2 }{K} \sum_{k \in \cK} \left ( \| x_{t}^k -  \bar x_{t+1}  \|^2   + \| y_{t}^k -  y_{t+1}^*   \|^2  \right )  ,
        \end{split}
\end{equation*}
where the last inequality uses the fact that $(1- \frac{\beta_t}{2}) ( 1+ \frac{\beta_t}{3})\leq 1 $ and $(1-\beta_t)(1- \frac{\beta_t}{2}) ( 1 +  \frac{3}{\beta_t } ) \leq \frac{7}{2\beta_t}$ when $T$ is sufficiently large.

Further, we can see 
\begin{equation}\label{eq:x_y_consensus}
\begin{split}
& 
\sum_{k \in \cK} \left ( \EE[ \| x_{t}^k -  \bar x_{t+1}  \|^2 ]  + \EE[ \| y_{t}^k -  y_{t+1}^*   \|^2 ] \right ) 
\\
& \leq \sum_{k \in \cK}  \EE \left [ 2\| x_{t}^k -  \bar x_t\|^2 +2 \|  \bar x_t- \bar x_{t+1}  \|^2  +3 \| y_{t}^k -  \bar y_t \|^2 + 3\| \bar y_t  - y_t^* \|^2  +  3\| y_t^* - y_{t+1}^*   \|^2   \right ]
\\
&  \leq \cO \left ( \frac{ K( \alpha_t^2 +  \beta_t^2) }{(1- \rho)^2 } \right ) + 2K(1+ L_y^2)\EE[ \|  \bar x_t- \bar x_{t+1}  \|^2]  + 3\sum_{k\in \cK} \EE[ \| \bar y_t - y_t^*\|^2 ]
\\
& \leq  \cO \left ( \frac{ K( \alpha_t^2 +  \beta_t^2) }{(1- \rho)^2 } \right )  + 2K (1 + L_y^2)  \alpha_t^2 \EE[ \|\bar z_t  \|^2] +  3 K \EE[ \| \bar y_t - y_t^*\|^2 ],
\end{split}
\end{equation}
implying that 
\begin{equation*}
    \begin{split}
       &  \EE[ \|  \bar s_{t+1} - \nabla f_x( \bar x_{t+1}, y_{t+1}^* ) \|^2]  
       \\
              & \leq (1-\beta_t)\EE [ \| \bar s_t - \nabla f (   \bar x_{t}, y_{t}^* ) \|^2] + \frac{2\beta_t^2   \sigma_f^2}{K }    + \frac{7\alpha_t^2  L_f^2 (1 + L_y^2 )}{2\beta_t }  \EE[  \| \bar z_t \|^2  ] 
              \\
              & \quad +    \cO \left ( \frac{ \beta_t ( \alpha_t^2 +  \beta_t^2) }{(1- \rho)^2 } \right )  + 12\beta_t L_f^2  \EE[ \| \bar y_t - y_t^*\|^2 ]. 
        \end{split}
\end{equation*}
This completes the proof.

\QED

%%%%%%%%%%%
%%%%%%%%%%%%
%%%%%%%%%%%%%

\subsection{Lemmas~\ref{lemma:h_t} and Its Proof} \label{app:lemma_ht}
% \begin{equation}
% h_{t+1}^k  =(1-\beta_j) \sum_{j \in \cN_k} w_{kj} h_t^j + \beta_j  \nabla_y f(x_t^k, y_t^k ; \zeta_t^k ).  \label{def:update_ht}
% \end{equation}
By following the analysis in Lemma~\ref{lemma:s_t}, we may obtain the following result. 
\begin{lemma}\label{lemma:h_t}
Suppose Assumptions \ref{assumption:1}, \ref{assumption:1}, and \ref{assumption:W}  hold and $T $ is sufficiently large, then we have 
\\
\noindent 
(a)\begin{equation}\label{eq:h_t}
    \begin{split}
    &    \EE[ \|  \bar h_{t+1} - \nabla_y f( \bar x_{t+1}, y^*(\bar x_{t+1}) ) \|^2]  
        \\
              & \leq (1-\beta_t)\EE [ \| \bar h_t - \nabla_y f (   \bar x_{t}, y^*(x_{t}) ) \|^2]    + \frac{7\alpha_t^2  L_f^2 (1 + L_y^2 )}{2\beta_t }  \EE[  \| \bar z_t \|^2  ] 
              \\
              & \quad  + \frac{2\beta_t^2   \sigma_f^2}{K }  +  \cO \left ( \frac{  \beta_t ( \alpha_t^2 +  \beta_t^2) }{(1- \rho)^2} \right ) + 12\beta_t L_f^2  \EE[ \| \bar y_t - y^*(x_{t})\|^2 ].
        \end{split}
\end{equation}
\noindent 
(b)
\begin{equation}\label{eq:u_t}
    \begin{split}
    &    \EE[ \|  \bar u_{t+1} - \nabla_{xy}^2 g( \bar x_{t+1}, y^*(\bar x_{t+1}) ) \|_F^2]  
        \\
              & \leq (1-\beta_t)\EE [ \| \bar u_t - \nabla_{xy}^2  g (   \bar x_{t}, y^*(x_{t}) ) \|_F^2]    + \frac{7\alpha_t^2  \widetilde L_g^2 (1 + L_y^2 )}{2\beta_t }  \EE[  \| \bar z_t \|^2  ] 
              \\
              & \quad  + \frac{2\beta_t^2   \sigma_g^2}{K }  +  \cO \left ( \frac{  \beta_t ( \alpha_t^2 +  \beta_t^2)  }{(1- \rho)^2 } \right )  + 12\beta_t \widetilde L_g^2  \EE[ \| \bar y_t - y^*(x_{t})\|^2 ].
        \end{split}
\end{equation}
\\
\noindent
(c)
\begin{equation}\label{eq:v_t}
    \begin{split}
    &    \EE[ \|  \bar v_{t+1} - \nabla_{yy}^2 g( \bar x_{t+1}, y^*(\bar x_{t+1}) ) \|_F^2]  
        \\
              & \leq (1-\beta_t)\EE [ \| \bar v_t - \nabla_{yy}^2 g (   \bar x_{t}, y^*(x_{t}) ) \|_F^2]    + \frac{7\alpha_t^2  \widetilde L_g^2 (1 + L_y^2 )}{2\beta_t }  \EE[  \| \bar z_t \|^2  ] 
              \\
              & \quad  + \frac{2\beta_t^2   \sigma_g^2}{K }  +  \cO \left ( \frac{  \beta_t ( \alpha_t^2 +  \beta_t^2)  }{(1- \rho)^2 } \right )  + 12\beta_t \widetilde L_g^2  \EE[ \| \bar y_t - y^*(x_{t})\|^2 ].
        \end{split}
\end{equation}

\end{lemma}
\textit{Proof:} Part (a) can be derived by following the analysis of Lemma~\ref{lemma:s_t}. 
The key difference between parts (b)-(c) and Lemma~\ref{lemma:s_t} is that here we establish stochastic recursions in terms of the Frobenius norm of  the estimation error matrices  $\nabla_{xy}^2 g(\bar x_t, \bar y_t) - \bar u_t \in \RR^{d_x \times d_y}$ and $\nabla_{yy}^2 g(\bar x_t, \bar y_t) - \bar v_{t,i} \in \RR^{d_y \times d_y}$, while Lemma~\ref{lemma:s_t} considers the Euclidean norm of the estimation error vector $\nabla_x f(\bar x_t, \bar y_t) -\bar s_t \in \RR^{d_x}$. 

(b) Here we provide the detailed proof of part (b) for completeness. 
By recalling Algorithm \ref{alg:1} Step 8 that $u_{t+1}^k   =(1-\beta_t) \sum_{j \in \cN_k} w_{kj} u_t^j + \beta_t  \nabla_{xy}^2 g^k(x_t^k, y_t^k ; \xi_t^k )$, we express the update rule of $\bar u_t = \frac{1}{K} \sum_{k \in \cK} u_t^k$ as 
\begin{equation*}
\bar u_{t+1} =(1-\beta_t) \bar u_t + \frac{\beta_t}{K} \sum_{k \in \cK}  \nabla_{xy}^2 g^k(x_t^k, y_t^k ; \xi_t^k ).
\end{equation*} 
Equivalently, we have 
\begin{equation*}
\begin{split}
 \bar u_{t+1} - \nabla_{yx}^2 g(\bar x_{t+1}, y_{t+1}^*)  
 = (1-\beta_t) [\bar u_t  -\nabla_{xy}^2 g(\bar x_{t+1}, y_{t+1}^*)   ]  + \beta_t \Delta_{g,t},
\end{split}
\end{equation*}
where 
$$
\Delta_{g,t}   =  \frac{1}{K}\sum_{k \in \cK}  \nabla_{yx}^2 g^k(x_t^k, y_t^k ; \xi_t^k )  - \nabla_{yx}^2 g(\bar x_{t+1},  y_{t+1}^* ) .
$$
Following \eqref{eq:st_1}, 
we  can see that 
\begin{equation}\label{eq:ut_1}
    \begin{split}
       & \EE[ \|   \bar u_{t+1} - \nabla_{xy}^2 g(\bar x_{t+1}, y_{t+1}^*)     \|_F^2] 
       \\
   &  = (1-\beta_t)^2 \EE [ \| \bar u_t  -\nabla_{xy}^2 g(\bar x_{t+1}, y_{t+1}^*)  \|_F^2] + \beta_t^2 \EE [ \|  \Delta_{g,t}  \|_F^2] \\
       & \quad + \frac{ 2(1-\beta_t)\beta_t}{K} \EE \left  [ \lv \bar s_t - \nabla_x f(\bar x_{t+1}, y_{t+1}^*  )  ,   \sum_{k \in \cK}  \nabla_x f^k(x_t^k, y_t^k )  - \nabla_x f(\bar x_{t+1},  y_{t+1}^* ) \rv_F  \right ]
    \\
    & \leq (1-\beta_t) \left (1- \frac{\beta_t}{2} \right ) \EE [ \| \bar u_t  -\nabla_{xy}^2 g(\bar x_{t+1}, y_{t+1}^*)  \|_F^2  ] + \beta_t^2 \EE [ \|  \Delta_{g,t}     \|_F^2] 
    \\
       & \quad + 2(1-\beta_t)\beta_t   \EE[ \|    \tfrac{1}{K}\sum_{k \in \cK}  \nabla_x f^k(x_t^k, y_t^k)  - \nabla_x f(\bar x_{t+1}, y_{t+1}^*  )    \|_F^2 ]  
       \\
       & \leq (1-\beta_t) \left (1- \frac{\beta_t}{2} \right ) \EE[ \| \bar u_t  -\nabla_{xy}^2 g(\bar x_{t+1}, y_{t+1}^*)  \|_F^2  ] + \beta_t^2  \EE [ \|  \Delta_{g,t}  \|_F^2]  \\
       & \quad  + \frac{2\beta_t(1-\beta_t)  L_g^2 }{K}  \sum_{k \in \cK } \left ( \EE[ \|   x_{t}^k - \bar x_{t+1} \|^2 ] + \EE[ \|   y_{t}^k -  y_{t+1}^*  \|^2 ]  \right ) ,
    \end{split}
\end{equation}
where $\lv \cdot , \cdot  \rv_F$ is the Frobenius inner product. 
By following \eqref{eq:st_2}, we may obtain 
\begin{equation}\label{eq:ut_2}
\begin{split}
\EE[ \| \Delta_{g,t} \|_F^2 ]
 & \leq    \frac{2}{K^2}  \sum_{k \in \cK} \EE[ \| \nabla_{xy}^2 g^k(x_t^k, y_t^k ; \xi_t^k )  - \nabla_{xy}^2 g^k(x_t^k, y_t^k)  \|_F^2 ] 
 \\
 & \quad +  \frac{2}{K} \sum_{k \in \cK} \EE[ \|  \nabla_{xy}^2 g^k ( x_t^k, y_t^k) - \nabla_{xy}^2 g^k(\bar x_{t+1},  y_{t+1}^* ) 
 \|_F^2 ] 
 \\
 & \leq \frac{2\sigma_g^2}{K} +  \frac{2 \widetilde L_g^2 }{K} \sum_{k\in \cK}  \left ( \EE[ \| x_{t}^k -  \bar x_{t+1}  \|^2 ] 
 + \EE[ \| y_{t}^k -  y_{t+1}^*   \|^2 ] \right ).
 \end{split}
\end{equation}

Using the fact that $\| A+B\|_F^2 \leq (1+ \frac{\beta_t}{3}) \|A \|_F^2 + (1+ \frac{3}{\beta_t}) \|B \|_F^2$ and Lemma~\ref{lemma:Lip} that $\| y_{t+1}^* - y_t^* \| \leq L_y \| x_{t+1} - x_t\|$, we have 
\begin{equation*}
\begin{split}
 & \| \bar u_t - \nabla_{xy}^2 g(\bar x_{t+1}, y_{t+1}^*)  \|^2
 \\
 & \leq \left ( 1 +  \frac{\beta_t}{3} \right )  \| \bar u_t  -\nabla_{xy}^2 g(\bar x_{t+1}, y_{t+1}^*)    \|^2  + \left (1 + \frac{3}{\beta_t}  \right ) \| \nabla_{xy}^2 g(\bar x_{t},  y_{t}^* )   -\nabla_{xy}^2 g(\bar x_{t+1}, y_{t+1}^* ) \|_F^2 
     \\
  & \leq  \left ( 1 +  \frac{\beta_t}{3} \right ) \| \bar u_t  -\nabla_{xy}^2 g(\bar x_{t+1}, y_{t+1}^*)   \|^2 + \left (1 + \frac{3}{\beta_t}  \right ) \widetilde L_g^2 (  \| \bar x_t - \bar x_{t+1}\|^2  +  \| y_t^*  -  y_{t+1}^* \|^2 )
  \\
  & =  \left ( 1 +  \frac{\beta_t}{3} \right ) \|\bar u_t  -\nabla_{xy}^2 g(\bar x_{t+1}, y_{t+1}^*)    \|^2 + \left  (1 + \frac{3}{\beta_t}  \right )\alpha_t^2 \widetilde L_g^2 (1+ L_y^2)\| \bar z_t \|^2. 
    \end{split}
\end{equation*}
By substituting  the above inequality and \eqref{eq:ut_2} into \eqref{eq:ut_1}, we obtain 
\begin{equation*}
    \begin{split}
       &  \EE[ \|   \bar u_{t+1} - \nabla_{xy}^2 g(\bar x_{t+1}, y_{t+1}^*)     \|_F^2 ]  
       \\
              & \leq (1-\beta_t)\EE [ \| \bar u_t  -\nabla_{xy}^2 g(\bar x_{t+1}, y_{t+1}^*)  \|_F^2  ] + \frac{2\beta_t^2   \sigma_g^2}{K }    + \frac{7\alpha_t^2  \widetilde L_g^2 (1 + L_y^2 )}{2\beta_t }  \EE[  \| \bar z_t \|^2  ] 
              \\
              & \quad + \frac{2\beta_t \widetilde L_g^2 }{K} \sum_{k \in \cK} \left ( \| x_{t}^k -  \bar x_{t+1}  \|^2   + \| y_{t}^k -  y_{t+1}^*   \|^2  \right ) .
        \end{split}
\end{equation*}
The desired result can be acquired by applying \eqref{eq:x_y_consensus}. This completes the proof. 

\QED

\subsection{Proof of Theorem \ref{thm:nonconvex}}\label{app:proof_nonconvex}
\textit{Proof:}
We start our analysis by considering the term $ \| \bar y_{t} -  y^*(\bar x_{t})\|^2$. By rearranging  \eqref{eq:yt_recursion}, we have 
\begin{equation}\label{eq:nonconvex_yt}
\begin{split}
  \EE [  \| \bar y_{t} -  y^*(\bar x_{t})\|^2]& \leq \frac{1}{\gamma_t \mu_g} \EE \left (  \| \bar y_{t} - y^*(\bar x_{t})\|^2 -  \| \bar y_{t+1} - y^*(\bar x_{t+1})\|^2 \right )  
   \\
   & \quad + \cO \left (\frac{ \alpha_t^2}{(1- \rho)^2} \right ) + \frac{2\gamma_t  \sigma_g^2}{\mu_g  K}  + \frac{3L_y^2 \alpha_t^2 }{\gamma_t^2 \mu_g^2 } \EE [\| \bar z_t \|^2 ].
   \end{split}
\end{equation}
Letting $m_1 = 4, m_2 = 12L_g^2 L_q^2 ,m_3 =  12 C_f^2 L_q^2 , m_{4,j} = 12 C_f^2 (1-\kappa_g)^{b - j } L_g^{-2} \kappa_g^{-1}   $, and $m_5 = 12 \big  ( L_f^2(m_1 + m_2) +  \widetilde L_g^2(m_3 + m_4)  \big )  $ be the constants defined within \eqref{eq:m},
we define a random variable
\begin{equation*}
\begin{split}
    P_t  & =   \tfrac{2}{\alpha_t}F(\bar x_t)  + \tfrac{m_1}{\beta_t}\| \bar s_t - \nabla_x f (   \bar x_{t}, y_{t}^* ) \|^2 
    +  \tfrac{m_2}{\beta_t}\| \bar h_t - \nabla_y f (   \bar x_{t}, y^*(x_{t}) ) \|^2
    \\
    & \quad + \tfrac{m_3}{\beta_t} \| \bar u_t - \nabla_{xy}^2g (   \bar x_{t}, y^*(x_{t}) ) \|_F^2 + \sum_{j=1}^b \tfrac{m_{4,j}}{\beta_t}\| \bar v_{t,j} - \nabla_{yy}^2  g (   \bar x_{t}, y^*(x_{t}) ) \|_F^2.
    \end{split}
\end{equation*}
Here we observe that $P_0 \leq \cO \left (\sqrt{ \frac{T}{K}}\right)$ and $P_t \geq \frac{2}{\alpha_t} F(x^*) = 2 \sqrt{ \frac{T}{K}} F(x^*) $.
By multiplying $m_1 \beta_t^{-1}$, $m_2\beta_t^{-1} $, $m_3 \beta_t^{-1}$, and $m_{4,j}\beta_t^{-1}$ to both sides of \eqref{eq:s_t}, \eqref{eq:h_t}, \eqref{eq:u_t}, and \eqref{eq:v_t}, respectively, and  combining with \eqref{eq:nonconvex_main}, we obtain that 
   \begin{equation*}
\begin{split}  
& \EE[ \| \nabla F(\bar x_t)\|^2]  + \EE [ P_{t+1} ]
\\
& \leq \EE[P_t]
 + \cO \left ( \frac{ \alpha_t^2 + \beta_t^2 }{(1- \rho)^2}\right ) +  12 \big  ( L_f^2(m_1 + m_2) +  L_g^2(m_3 + \sum_{j=1}^b m_{4,j})  \big )  \EE[ \| \bar y_t - y_t^*\|^2 ] 
 \\
 & \quad + \cO \Big (  \frac{\beta_t   }{K }   \Big )
- \left ( 1 - \alpha_t L_F   - \frac{7\alpha_t^2  \big (L_f^2(m_1 + m_2)  + \widetilde L_g^2 (m_3 + \sum_{j=1}^b m_{4,j} ) \big )(1 + L_y^2 )}{2\beta_t^2 }  \right ) \EE[  \|\bar z_t \|^2]. 
  \end{split}
\end{equation*}
Letting $m_4 = 12 C_f^2L_g^{-2} \kappa_g^{-2} $, we note that $m_4 \geq \sum_{j=1}^b m_{4,j}$, and express the above inequality as 
   \begin{equation*}
\begin{split}  
& \EE[ \| \nabla F(\bar x_t)\|^2]  + \EE [ P_{t+1} ]
\\
& \leq \EE[P_t]
 + \cO \left ( \frac{ \alpha_t^2 +  \beta_t^2 }{(1- \rho)^2}\right ) +  12 \big  ( L_f^2(m_1 + m_2) +  \widetilde L_g^2(m_3 + m_4 )  \big )  \EE[ \| \bar y_t - y_t^*\|^2 ] 
 \\
 & \quad + \cO \Big (  \frac{\beta_t    }{K }   \Big )
- \left ( 1 - \alpha_t L_F   - \frac{7\alpha_t^2  \big (L_f^2(m_1 + m_2)  + \widetilde L_g^2 (m_3 + m_4 ) \big )(1 + L_y^2 )}{2\beta_t^2 }  \right ) \EE[  \|\bar z_t \|^2]. 
  \end{split}
\end{equation*}
By multiplying $m_5$ to both sides of \eqref{eq:nonconvex_yt}, combining the above inequality with \eqref{eq:nonconvex_yt}, and recalling that $\alpha_t = C_0\sqrt { \frac{K}{T} } $ and $\beta_t = \gamma_t = \sqrt { \frac{K}{T} } $, we further obtain 
\begin{equation*}
    \begin{split}
    &     \EE[ \| \nabla F(\bar x_t)\|^2]  + \EE [ P_{t+1} ] + \frac{ m_{5} }{\mu_g } \sqrt{ \frac{T}{K}}   \EE [\| \bar  y_{t+1}  - y_{t+1}^*\|^2 ] 
        \\
& \leq \EE[P_t] +\frac{ m_{5} }{\mu_g } \sqrt{ \frac{T}{K}}    \EE [\| \bar  y_{t}  - y_{t}^*\|^2 ] 
  + \cO \left ( \frac{K}{T (1- \rho)^2}\right ) +\cO \Big (   \frac{1   }{\sqrt{KT}  }  \Big ) 
\\
& \quad  - \underbrace{ \left (  1 -\frac{C_0L_F}{\sqrt{T}}- \frac{7C_0^2  (L_f^2(m_1 + m_2)  + \widetilde L_g^2 (m_3 + m_4) )(1 + L_y^2 )}{2 }  - \frac{3 m_5 L_y^2 C_0^2 }{ \mu_g^2} \right ) }_{\Upsilon(C_0,T)}\EE[  \|\bar z_t \|^2].
  \end{split}
\end{equation*}
% where $M_t =\tfrac{ 12C_g^2( C_f^2 +  L_q^2)}{\beta_t}   \frac{12L_f^2 }{\gamma_t \mu_g} +  \tfrac{4 + 12 C_f^2 L_q^2}{\beta_t}  \frac{12L_g^2}{\gamma_t \mu_g}$ and $\tilde C =  \tfrac{7 (1 + L_y^2 )}{2 }    +  \frac{36 L_y^2}{ \mu_g^2  }$. 
We then observe that for large $T$, there exists a small constant $\tilde C_0 >0$ such that $\Upsilon(C_0,T) \geq0$ for all $C_0 \leq \tilde C_0$.
% %By setting $C_0 =  \left [\tilde C \Big ( 24C_g^2 C_f^2 L_g^2   + 14 C_g^2 L_q^2 L_f^2 + 8L_f^2  + 24 C_f^2 L_q^2 L_g^2 \Big ) \right ]^{-1/2}$, 
% $$
%  1 -\frac{C_0L_F}{\sqrt{T}}- \frac{7C_0^2  (L_f^2(m_1 + m_2)  + L_g^2 (m_3 + m_4) )(1 + L_y^2 )}{2 }  - \frac{3 m_5 L_y^2 C_0^2 }{ \mu_g^2}  \geq 0$$ for $T $ sufficiently large.
 In such scenario, we sum the above inequality over $t=0,1,\cdots, T-1$ and conclude that 
  \begin{equation*}
\begin{split}
\frac{1}{T}\sum_{t=0}^{T-1} \EE[ \| \nabla F(\bar x_t)\|^2] & \leq \frac{P_0 + \frac{ m_{5} }{\gamma_t \mu_g} \| \bar y_0 - y_0^*\|^2 - \EE[P_{T}] }{T}  +  \cO \Big ( \frac{1}{\sqrt{ TK} } \Big )   + \cO \left ( \frac{1}{T (1- \rho)^2 }\right ) 
\\
 & \leq  \cO \Big ( \frac{1}{\sqrt{ TK} } \Big )   + \cO \left ( \frac{K}{T (1- \rho)^2 }\right ),
  \end{split}
\end{equation*}
where the last inequality applies the facts $P_0 +  \frac{ m_{5} }{\mu_g } \sqrt{ \frac{T}{K}}  \| \bar y_0 - y_0^*\|^2\leq \cO(\sqrt{ \frac{T}{K} })$ and $P_T \geq 2 \sqrt{ \frac{T}{K} }  F(x^*)$. 
This completes the proof. 
\QED

\section{Proof of Results for $\mu$-PL Objectives} \label{app:C}
Throughout this subsection, we assume Assumptions~\ref{assumption:1}, \ref{assumption:2}, \ref{assumption:W}, and \ref{assumption:PL} hold. We set $b = 3 \lceil \log_{\frac{1}{1-\kappa_g}} T \rceil $, consider the scenario where step-sizes follow \eqref{eq:stepsize_PL} such that 
\begin{equation*}
\alpha_t = \frac{2}{\mu(C_1 + t)},    \text{ and } \beta_t =  \gamma_t = \frac{C_1}{ C_1 + t},  \ \ \text{ for } 1\leq t \leq T,
\end{equation*}
where $C_1 >0$ is  a large constant making 
$$
\Psi(C_1) =  \frac{1}{2} -  \frac{ 2L_F}{\mu C_1  }
  - \frac{7\alpha_t  (L_f^2(z_1 + z_3)  + \widetilde L_g^2 (z_3 + z_4) )(1 + L_y^2 )}{ \mu C_1  }  -  \frac{6z_5 L_y^2 }{ \mu_g \mu C_1 } > 0 
$$
% $$C_1  =  \frac{28 (L_f^2(z_1 + z_3)  + L_g^2 (z_3 + z_4) )(1 + L_y^2 )}{\mu}  +  \frac{24z_5 L_y^2  }{ \mu_g \mu } >0 
% $$ 
with 
\begin{equation}\label{def:z}
    \begin{split}
   z_1  = \frac{4}{\mu( C_1 - 2 ) } , \ \  & z_2  = \frac{12 L_g^2 L_q^2 }{\mu( C_1 -2 )},   z_3   = \frac{12 C_f^2 L_q^2 }{ \mu( C_1   - 2 ) }, 
 \\
 z_4  =\frac{12   C_f^2  }{ \mu( C_1   - 2) L_g^2  \kappa_g^{2} }  , \ \ \text{ and } & z_5 =  \frac{ 12 C_1 \big  ( L_f^2(z_1 + z_2) +   \widetilde L_g^2(z_3 + z_4)  \big ) }{\mu_g (C_1 - 2/\mu)}.      
    \end{split}
\end{equation}

%%%%%%%%%%%%
%%%%%%%%%%%%
%%%%%%%%%%%%
\subsection{Lemma \ref{lemma:PL_01} and Its Proof} \label{app:main_PL}

\begin{lemma}\label{lemma:PL_01}
Suppose Assumption \ref{assumption:1}, \ref{assumption:2}, \ref{assumption:W}, and \ref{assumption:PL}  hold and the objective $F$ satisfies $\mu$-PL Assumption \ref{assumption:PL} in addition. We have 
   \begin{equation*}
\begin{split} 
	  & \EE[F(\bar{x}_{t+1})]  - F^* 
	  \\
	  &  \leq \big (1   - \alpha_t \mu  \big)\E[ F(\bar x_t) -F^*]  
  -  \frac{\alpha_t}{2}(1 - \alpha_t L_F ) \EE[ \|  \bar z_t  
\|^2  ]   + 2\alpha_t  \EE [ \|  \nabla_x f( \bar x_t ,   y_t^*   ) - \bar s_t \|^2 ]   
\\
& \quad+ 6\alpha_t L_g^2 L_q^2  \EE[  \| \nabla_y f(  \bar x_t  ,   y^*_t    ) -  \bar h_t \|^2 ]   +6\alpha_t C_f^2 L_q^2  \EE[ \|  \nabla_{xy}^2 g(  \bar x_t  ,   y_t^*   ) - \bar u_t \|_F^2 ] 
\\
& \quad  
+  \frac{6 C_f^2}{L_g^2 \kappa_g}  \sum_{j=1}^b (1-\kappa_g)^{b-j}  \EE [ \| \nabla_{yy}^2 g( \bar x_t  ,    y_t^*   )  -  \bar v_{t,j} \|_F^2 ]   + \cO \left ( \frac{\alpha_t \beta_t^2 }{ (1- \rho)^2 }\right )
. 
  \end{split}
\end{equation*}

\end{lemma}

\textit{Proof:}
Suppose the objective function $F$ satisfies the $\mu$-PL condition \eqref{def:PL}, by combining it with \eqref{eq:L_F_smooth}, we have 
\begin{equation*}
\begin{split}
& \EE[ F(\bar{x}_{t+1}) ]  - \EE [ F(\bar{x}_{t}) ]
\\
& \leq   - \alpha_t \mu \EE  [ F(\bar x_t ) - F^*]   - \frac{\alpha_t}{2}(1-\alpha_t L_F) \EE [ \|  \bar z_t  
\|^2 ]  
\\
& \quad + \frac{\alpha_t}{K} \sum_{k=1}^K \Big ( \EE [ \|  \nabla_x f( \bar x_t ,   y_t^*   ) - s_t^k \|^2 ] +3 C_f^2 L_q^2  \EE[ \|  \nabla_{xy}^2 g(  \bar x_t  ,   y_t^*  ) - u_t^k \|^2 ] \Big ) 
\\ 
& \quad +  \frac{3\alpha_t}{K} \sum_{k=1}^K \Big ( L_g^2 C_f^2  \EE [ \| [\nabla_{yy}^2 g( \bar x_t  ,    y_t^*   ) ]^{-1} - q_t^k \|^2 ]+  C_g^2 L_q^2 \EE[  \| \nabla_y f(  \bar x_t  ,   y_t^*  ) -  h_t^k \|^2 ] \Big ).
\end{split}
\end{equation*}
By following \eqref{eq:nonconvex_03}, we further express the above inequality as  
\begin{equation*} 
\begin{split}
& \EE[ F(\bar{x}_{t+1}) ]  - \EE [ F(\bar{x}_{t}) ]
\\
& \leq   - \alpha_t \mu \EE  [ F(\bar x_t ) - F^*]   - \frac{\alpha_t}{2}(1-\alpha_t L_F) \EE [ \|  \bar z_t  
\|^2 ] + 2\alpha_t \EE [ \|  \nabla_x f( \bar x_t ,   y_t^*   ) - \bar s_t \|^2 ] 
\\
& \quad  + 6\alpha_t  L_g^2 L_q^2 \EE[  \| \nabla_y f(  \bar x_t  ,   y_t^*  ) -   \bar h_t \|^2 ]  +6\alpha_t C_f^2 L_q^2  \EE[ \|  \nabla_{xy}^2 g(  \bar x_t  ,   y_t^*  ) -  \bar u_t \|_F^2 ]
\\ 
& \quad + \frac{6 C_f^2}{L_g^2 \kappa_g}  \sum_{j=1}^b (1-\kappa_g)^{b-j}  \EE [ \| \nabla_{yy}^2 g( \bar x_t  ,    y_t^*   )  -  \bar v_{t,j} \|_F^2 ]    + \cO \left ( \frac{\alpha_t \beta_t^2 }{ (1- \rho)^2 }\right ).
\end{split}
\end{equation*}
We acquire the desired result by substracting $F^*$ on both sides of the above inequality.

\QED

%%%%%%%%%%%%%%%%%
%%%%%%%%%%%%%%%%%%
%%%%%%%%%%%%%%%%%%

%%%%%%%%%%%%
%%%%%%%%%%%%%

\subsection{Proof of Theorem \ref{thm:PL}}\label{app:proof_of_thm_PL}
Before establishing the convergence rate for $\mu$-PL function, we provide a result \citep[Lemma 1]{ghadimi2016accelerated}   to characterize the convergence behavior for a random sequence satisfying a  special form of stochastic recursion as follows.

\begin{lemma}\label{lemma:sequence_alpha}
 Letting  $b_k = \frac{2}{k+1}$  and $\Gamma_k = \frac{2}{k(k+1)}$ for $k \geq 1$ be two nonnegative sequences.
%  \begin{equation*}
% \Gamma_k =
%  \begin{cases}
%  1,  & k=1\\
%  (1-b_k) \Gamma_{k-1}. & k \geq 2.
% \end{cases}
% \end{equation*} 
For any nonnegative sequences $\{ A_k \}$ and $\{ B_k \}$  satisfying
$$
A_k \leq (1-b_k) A_{k-1} + B_k , \ \ \text{ for } k \geq 1,
$$
we have $\Gamma_k =  \Gamma_{s} \prod_{j=s+1}^k (1-b_j)$ and 
\begin{equation*}
    A_k \leq  \frac{\Gamma_k}{\Gamma_s} A_{s} + \sum_{i=s+1}^k \frac{   \Gamma_k B_i }{\Gamma_i}.
\end{equation*}
\end{lemma}

We then derive the convergence rate of $\{ \bar x_t\}$ for $\mu$-PL objectives as follows.

\textit{Proof of Theorem \ref{thm:PL}:}
First of all, under the choice of step-sizes that  $\alpha_t  = \frac{2}{\mu(C_1 + t)}$ and $\beta_t = \gamma_t = \frac{C_1}{ C_1 + t}$, we have  $\lim_{t\to \infty} \alpha_t  = 0$, $\lim_{t\to \infty} \beta_t = 0$, and  $\lim_{t\to \infty} \gamma_t = 0$. By following the analysis of Lemmas~\ref{lemma:y_t},~\ref{lemma:s_t}, and  \ref{lemma:h_t} and applying the convergence rates of consensus errors in Lemma~\ref{lemma:consensus_adaptive},  we obtain that 
hold for all $t  \geq 1$. 

Next, we define a random variable 
\begin{align*}
J_k & =  F(\bar x_t)  -F^* +  z_1   \|  \nabla_x f( \bar x_t , y^*_t    ) - \bar s_t \|^2   + z_2  \| \nabla_y f(  \bar x_t  ,   y^*_t    ) -  \bar h_t \|^2   
\\
& \quad + z_3   \|  \nabla_{xy}^2 g(  \bar x_t  ,  y^*_t    ) - \bar u_t \|_F^2      + \sum_{1\leq j \leq b}z_{4,j} \| \nabla_{yy}^2 g( \bar x_t  ,    y^*_t    )  - \bar v_{t,j} \|_F^2,
\end{align*}
where 
\begin{align*}
 z_1 = \frac{2\alpha_t}{\beta_t  - \alpha_t \mu } = \frac{4}{\mu( C_1 - 2 ) } , \ \  & z_2 = \frac{6\alpha_t L_g^2 L_q^2 }{\beta_t  - \alpha_t \mu } = \frac{12 L_g^2 L_q^2 }{\mu( C_1 -2 )}, \\
 z_3  = \frac{6\alpha_t C_f^2 L_q^2 }{\beta_t  - \alpha_t \mu } = \frac{12 C_f^2 L_q^2 }{ \mu( C_1   - 2 ) }, \ \ \text{ and } & z_{4,j} = \frac{6\alpha_t L_g^2 C_f^2(1-\kappa_g)^{b-j}  }{(\beta_t  - \alpha_t \mu ) L_g^4  \kappa_g }    =\frac{12  L_g^2 C_f^2 (1-\kappa_g)^{b-j}  }{ \mu( C_1   - 2) L_g^4  \kappa_g },
\end{align*}
are all constants defined in \eqref{def:z}. By multiplying $z_1, z_2 , z_3$, and $z_{4,j}$ to  both sides of \eqref{eq:s_t}, \eqref{eq:h_t}, \eqref{eq:u_t}, and \eqref{eq:v_t}, respectively,  and combining them with Lemma~\ref{lemma:PL_01}, we obtain 
\begin{equation*}
	\begin{split}
	 \EE[ J_{t+1}	]	
	& \leq  (1   - \alpha_t \mu ) \EE[ J_t] +  \cO \Big (  \frac{\beta_t^2 }{K }   \Big )    +  \cO \left ( \frac{1}{t^3 (1-\rho)^2}\right ) 
  \\
  & \quad   - \alpha_t \left ( \frac{1-\alpha_t L_F}{2}
  - \frac{7\alpha_t  \big (L_f^2(z_1 + z_3)  + L_g^2 (z_3 + \sum_{1\leq j \leq b}z_{4,j} )  \big )(1 + L_y^2 )}{2\beta_t }  \right ) \EE[  \|\bar z_t \|^2]
  \\
  & \quad 
  + 12 \beta_t  \big  ( L_f^2(z_1 + z_2) +  L_g^2(z_3 + \sum_{1\leq j \leq b}z_{4,j})  \big )  \EE[ \| \bar y_t - y_t^*\|^2 ] .
\end{split}
\end{equation*}
By noting that $\sum_{1\leq j \leq b}z_{4,j} \leq L_g^{-4} \kappa_g^{-2} = z_4$ in \eqref{def:z}, we further express the above inequality as 
\begin{equation*}
	\begin{split}
	 \EE[ J_{t+1}	]	
	& \leq  (1   - \alpha_t \mu ) \EE[ J_t] +  \cO \Big (  \frac{\beta_t^2 }{K }   \Big )    +  \cO \left ( \frac{1}{t^3 (1-\rho)^2}\right ) 
  \\
  & \quad   - \alpha_t \left ( \frac{1-\alpha_t L_F}{2}
  - \frac{7\alpha_t  \big (L_f^2(z_1 + z_3)  + L_g^2 (z_3 + z_4)  \big )(1 + L_y^2 )}{2\beta_t }  \right ) \EE[  \|\bar z_t \|^2]
  \\
  & \quad 
  + 12 \beta_t  \big  ( L_f^2(z_1 + z_2) +  L_g^2(z_3 + z_4)  \big )  \EE[ \| \bar y_t - y_t^*\|^2 ] .
\end{split}
\end{equation*}

By recalling that $\alpha_t = \frac{2}{\mu(C_1 + t)}$ and $\beta_t = \gamma_t =  \frac{C_1}{C_1 + t}$, we have $\alpha_t/\beta_t =  \alpha_t/\gamma_t  =  \frac{2}{\mu C_1}$.
Further, letting 
$$z_5 = \frac{ 12 \beta_t  \big  ( L_f^2(z_1 + z_2) +  L_g^2(z_3 + z_4)  \big ) }{\mu_g (\gamma_t - \alpha_t)} =  \frac{ 12 C_1 \big  ( L_f^2(z_1 + z_2) +  L_g^2(z_3 + z_4)  \big ) }{\mu_g (C_1 - 2/\mu)},$$
by multiplying $z_5 $ to both sides of \eqref{eq:yt_recursion} and combining with the above inequality, we have 
\begin{equation*}
	\begin{split}
	 & \EE[ J_{t+1}] 	 + z_5 \EE[ \| \bar y_{t+1} - y_{t+1}^*\|^2]
	 \\
	& \leq  \left (1   - \tfrac{2}{C_1 + t+1}  \right ) \Big ( \EE[ J_t ]  + z_5\EE[ \| \bar y_{t} - y_{t}^*\|^2] \Big )  +  \cO \Big (  \frac{\beta_t^2 }{K }   \Big )    +  \cO \left ( \frac{1}{t^3 (1-\rho)^2 }\right ) 
  \\
  & \quad   - \alpha_t \underbrace{ \left ( \frac{1}{2} -  \frac{ 2L_F}{\mu C_1 }
  - \frac{7   (L_f^2(z_1 + z_3)  + L_g^2 (z_3 + z_4) )(1 + L_y^2 )}{ \mu C_1  }  -  \frac{6z_5 L_y^2 }{ \mu_g \mu C_1 } \right ) }_{\Psi(C_1)}\EE[  \|\bar z_t \|^2]
 .
\end{split}
\end{equation*}
We note that $\Psi(C_1)$ is increasing in $C_1$ with $\lim_{C_1 \to \infty} \Psi(C_1) = \frac{1}{2}$. 
Clearly, there exists a constant $\tilde C_1>0$ such that  $\Psi(C_1) \geq 0$ for $C_1 \geq \tilde C_1$.  
% which further implies that $\delta_t \geq 0$ when $\alpha_t L_F \leq \frac{1}{2}$ and 
% $$
% C_1 = \frac{28 (L_f^2(z_1 + z_3)  + L_g^2 (z_3 + z_4) )(1 + L_y^2 )}{\mu}  +  \frac{24z_5 L_y^2 }{ \mu_g \mu } >0 .
% $$
%Finally, we denote by $\Gamma_1 = 1$ and $\Gamma_{k} = (1- \frac{2}{k+1}) \Gamma_{k-1} $  for $k \geq 2$. Basic algebra yields that $\Gamma_k = \frac{2}{k(k+1)}$.  
% Letting $t_1 =  \max  \{ t_0, \tilde t \}$, we observe that $t_1$ is finite and there exists a constant $D_{t_1} \geq 0$ such that $ \EE[ J_{t_1}]	  + z_5 \EE[ \| \bar y_{t_1} - y_{t_1}^*\|^2] \leq D_{t_1}$ is bounded. 
Finally, we conclude that  for $t \geq 0$, 
\begin{equation*}
	\begin{split}
	&  \EE[ J_{t+1}	]  + z_5 \EE[ \| \bar y_{t+1} - y_{t+1}^*\|^2] 
	 \\
	& \leq  \Big (1   - \tfrac{2}{C_1 + t+1} \Big ) \Big ( \EE[ J_t ] + z_5 \EE[ \| \bar y_{t} - y_{t}^*\|^2] \Big ) + \cO \left (\frac{\beta_t^2}{K} \right )  + \cO \left (\frac{1}{t^3(1-\rho)^2} \right )\\
	& \leq \frac{\Gamma_{C_1 + t}}{\Gamma_{C_1 }}[ J_0  + z_5 \| \bar y_{0} - y_{0}^*\|^2 ]  + \sum_{j=C_1}^{C_1 + t} \cO \Big ( \frac{\beta_j^2 \Gamma_{C_1 + t}}{K \Gamma_j} \Big)   + \sum_{j = C_1 }^{t+ C_1 } \cO \left (\frac{\Gamma_{t + C_1}}{j^3(1-\rho)^2 \Gamma_j} \right ) 
	\\
	& = \frac{C_1 (C_1 + 1) J_0 }{(C_1 + t)(C_1+t+1)} + \frac{2}{(C_1 + t )(C_1 + t+1)} \left (\sum_{j=C_1}^{C_1 + t}  \cO \Big ( \frac{ j(j+1)}{j^2 K}  \Big )  + \sum_{j=C_1}^{C_1 + t} \cO \left (\frac{j (j+1)}{j^3(1-\rho)^2 } \right )   \right ) 
\\
& \leq  \frac{C_1 (C_1 + 1) J_0 }{(C_1 + t)(C_1+t+1)}  + \cO \Big (  \frac{t}{(C_1 +t)(C_1 + t+1) K}  \Big )  + \cO \left ( \frac{2\ln ( C_1 + t) }{(C_1 +t)(C_1 + t+1)(1-\rho)^2}  \right ) 
\\
&
\leq \cO \Big ( \frac{1}{(t+1) K} \Big ) + \cO \left ( \frac{\ln t}{t^2 (1-\rho)^2 } \right ) ,
	\end{split}
\end{equation*}
where the second inequality uses Lemma~\ref{lemma:sequence_alpha}. This completes the proof. 

\QED

%%%%%%%%%%%%
%%%%%%%%%%%%
%%%%%%%%%%%%

\section{Result for Single-center-multi-user-federated Stochastic Bilevel Optimization}
We first provide the following Federated Sochastic Bilevel Optimization algorithm to tackle federated SBO over a Single-center-multi-user star network. Our algorithm utilizes the star network structure that a central server connects all agents and synchronously  passes the same solution $(x_t,y_t)$ to all agents within each learning round. Then, each agent would query the stochastic information from its own data and pass it to the central server. After collecting all stochastic information, the central server would employ a similar stochastic approximation scheme as DSBO to update its estimation of the gradients and hessians within \eqref{eq:gradient_BO}. The details are provided in Algorithm~\ref{alg:2}. 
\begin{algorithm}[t!]
\caption{Federated Stochastic Bilevel Optimization over Star Network}\label{alg:2}
\begin{algorithmic}[1]
\REQUIRE Step-sizes $\{ \alpha_t \}$,  $ \{ \beta_t \}$, $\{ \gamma_t\}$, total iterations $T$.
\\
$x_0^k = \bf{0}$, $y_0^k = \bf{0}$, $u_0^k = \bf{0}$, $v_0^k = \mu_g \bf{I}$, $s_0^k = \bf{0}$, $h_0^k = \bf{0}$
\FOR{$t=0, 1, \cdots, T-1$}
	\STATE \textcolor{blue}{Communication:} Pass $(x_t,y_t)$ to each agent $k \in \cK$.
	\FOR{$k=1,\cdots, K$}
	\STATE \textcolor{blue}{Local sampling}: Query $\cS\cO$ at $(x_t,y_t)$ to obtain $\nabla_x f^k(x_t,y_t;\zeta_t^k)$, 
	$\nabla_y f^k (x_t,y_t;\zeta_t^k)$, 
	\\
			\hspace{2.3cm}  $\nabla_{xy}^2 g^k(x_t,y_t;\xi_t^k)$,   and  $\nabla_{yy}^2  g^k(x_t,y_t;\xi_{t}^k) $.
%  		\STATE \textcolor{blue}{Gradient update:}   $z_{t+1}^k =  s_{t+1}^k -  u_{t+1}^k v_{t+1}^k h_{t+1}^k $.

 	\ENDFOR
 	
 	\STATE \textcolor{blue}{Communication:} Receive  stochastic samples from each agent $k \in cK$. 
 
	\STATE \textcolor{blue}{Outer loop update}: $x_{t+1} =  x_t  - \alpha_t  \left ( s_t -  u_t [v_t ]^{-1} h_t \right ) $.
	\STATE 
	%Query the $\cS\cO$ at $(x_t^k,y_t^k)$ to obtain $\nabla_y g^k(x_t^k,y_t^k;\xi_t^k)$ and $\nabla_{xy}^2 g^k(x_t^k,y_t^k;\xi_t^k)$.
	%Query the $\cS\cO$ at $(x_t^k,y_t^k)$ to obtain stochastic gradient $\nabla_y g^k (x_t^k,y_t^k;\xi_t^k )$. 
	\textcolor{blue}{Inner loop update}:  
$
	y_{t+1}  =  y_t   - \tfrac{ \gamma_t }{K}  \sum_{k\in \cK}\nabla_y g^k (x_t^k,y_t^k;\xi_t^k ).
$
\STATE 
\textcolor{blue}{Estimate $\nabla_x f(x_t,y_t)$:}
$
s_{t+1}   =(1-\beta_t) s_t + \tfrac{\beta_t}{K}  \sum_{k\in \cK} \nabla_x f^k(x_t, y_t ; \zeta_t^k )$.
\STATE 
\textcolor{blue}{Estimate $\nabla_y f(x_t,y_t)$:}
$
h_{t+1}  =(1-\beta_t)  h_t + \frac{\beta_t}{K}  \sum_{k\in \cK} \nabla_y f^k(x_t, y_t ;  \zeta_t^k).  
$

\STATE \textcolor{blue}{Estimate $\nabla_{xy}^2 g(x_t,y_t)$:} 
$
u_{t+1}   =(1-\beta_t) u_t + \frac{ \beta_t }{K} \sum_{k\in \cK} \nabla_{xy}^2 g^k(x_t,  y_t; \xi_t^k).
$

\STATE 
\textcolor{blue}{Estimate $[\nabla_{yy}^2  g(x_t,y_t)]^{-1}$:} 
Set 
$Q_{t+1} =  \mathbf{I}$. 
\FOR{ $i= 1,\cdots,b$}
\STATE $v_{t+1,i}  =(1-\beta_t) v_{t,i} + \tfrac{\beta_t}{K} \sum_{k \in \cK} \nabla_{yy}^2 g^k(x_t^k,y_t^k;\xi_{t,i}^k),$
\STATE $Q_{t+1,i} = \mathbf{I} + ( \mathbf{I} - \frac{1}{L_g} v_{t+1,i} )Q_{t+1,i-1}$
\ENDFOR
\STATE Set $q_{t+1} = \frac{1}{L_g} Q_{t+1,b}$
% \STATE \textcolor{blue}{Estimate $\nabla_{xy}^2 g(x_t,y_t)$:} 
% $
% v_{t+1}  =(1-\beta_t)  v_t + \tfrac{\beta_t}{K} \sum_{k\in \cK}  \nabla_{yy}^2  g^k(x_t,y_t;\xi_{t}^k) .
% $
 \ENDFOR
 \ENSURE $\{ x_t \}_{t=0}^{T-1}$.
\end{algorithmic}
\end{algorithm}

\begin{theorem}[Single-center-multi-user-federated SBO]\label{thm:single_center}
Suppose Assumptions~\ref{assumption:1}, \ref{assumption:2}, and \ref{assumption:W} hold. Letting $\{ x_t\}$ be the sequence generated by Algorithm~\ref{alg:2} with step-sizes defined in \eqref{eq:m}, we have 
\end{theorem}
\textit{Proof:} We observe that the key difference between Alg.~\ref{alg:2} and  Alg.~\ref{alg:1} is that Alg.~\ref{alg:2} considers an environment where each agent 
synchronously receives a common solution $(x_t,y_t)$ from the central agent so that the effect of consensus disappears. Consequently, with a slight abuse of notation that $(x_t, y_t)  = (\bar x_t, \bar y_t)$. Further, we observe that all consensus error terms within \eqref{eq:nonconvex_main}, \eqref{eq:yt_recursion}, \eqref{eq:s_t}, \eqref{eq:h_t}, \eqref{eq:u_t}, and \eqref{eq:v_t} would disappear, leading to  tightened characterizations of the overall estimation errors under the star-network. For instance, \eqref{eq:nonconvex_main} could be tightened as
 \begin{equation*}
\begin{split}  
& \EE[ \| \nabla F(\bar x_t)\|^2] 
\\
& \leq  \frac{2}{\alpha_t } \Big ( \EE[ F(\bar{x}_t)] -  \E[F(\bar{x}_{t+1}) ] \Big ) 
  -  (1 - \alpha_t L_F ) \EE[ \|  \bar z_t  
\|^2  ]  
 + 4  \EE [ \|  \nabla_x f( \bar x_t ,   y_t^*   ) - \bar s_t \|^2 ]  
 \\
 & \quad + 12 L_g^2 L_q^2  \EE[  \| \nabla_y f(  \bar x_t  ,   y^*_t    ) -  \bar h_t \|^2 ]  
 +12 C_f^2 L_q^2  \EE[ \|  \nabla_{xy}^2 g(  \bar x_t  ,   y_t^*   ) - \bar u_t \|_F^2 ]  
\\
& \quad  
+  \frac{12 C_f^2 }{L_g^2 \kappa_g  }   \sum_{1 \leq j \leq  b}  (1-\kappa_g)^{b - j } \EE[ \| \nabla_{yy}^2 g(\bar x_t, y_t^*  ) -  \bar v_{t,i} )\|_F^2] 
. 
  \end{split}
\end{equation*}
By following the analysis of Theorem~\ref{thm:nonconvex} and employing the step-sizes in \eqref{eq:m}, we obtain that 
\begin{equation*}
    \frac{1}{T} \sum_{t=0}^{T-1} \| \nabla F(x_t) \|^2 \leq \cO \left (\frac{1}{\sqrt{KT}} \right ).
\end{equation*}
This completes the proof. 

\section{Additional Numerical Experiments}\label{app:numerics}

\subsection{Hyper-parameter Optimization} \label{app:hyper_opt}

The baseline algorithm DBSA conducts the followings. 
At the outer solution  $x_t^k$, DBSA obtains an estimator $y_t^k$ of $y^*(x_k)$ via conducting $t$ gossip stochastic gradient descent steps
$\tilde y_{t,i+1}^k = \sum_{j \in \cN_k}w_{k,j}\tilde y_{t,i}^{j} - \eta_{t,i} \nabla_y g(x_t^k, \tilde y_{t,i}^k;\xi_i^k) $ for $i=0,1,\cdots, t$ with $y_t^k = \tilde y_{t,t}^k$, and then update the main solution $x_t^k$ by one stochastic gradient descent step that $x_{t+1}^k = \sum_{j \in \cN_k}w_{k,j}x_t^{j} - \alpha_t \nabla_x  f(x_t^k,y_t^k;\zeta_t^k)$.  We summarize the details in Algorithm~\ref{alg:DBSA}. 

\begin{algorithm}[t!]
\caption{Decentralized Bilevel Stochastic Approximation}\label{alg:DBSA}
\begin{algorithmic}[1]
\REQUIRE Step-sizes $\{ \alpha_t \}$,  $\{ \eta_{t,i} \}$, number of total iterations $T$.
\\
$x_0^k = \bf{0}$, $y_0^k = \bf{0}$
\FOR{$t=0, 1, \cdots, T-1$}
	\STATE 	\textcolor{blue}{Inner loop update}:  
	\FOR{ $i = 0,1,\cdots, t$}
	\FOR{$k = 1,2,\cdots, K$}
		\STATE \textcolor{blue}{Local sampling}: Query $\cS\cO$ at $(x_t^k, \tilde y_{t,i}^k)$ to obtain $\nabla_y g(x_t^k, \tilde y_{t,i}^k;\xi_{t,i}^k)  $.
	\STATE \textcolor{blue}{Estimate:} $\tilde y_{t,i+1}^k = \sum_{j \in \cN_k}w_{k,j}\tilde y_{t,i}^{j} - \eta_{t,i} \nabla_y g(x_t^k, \tilde y_{t,i}^k;\xi_{t,i}^k) $.
 	\ENDFOR
 	\ENDFOR 
 
	\STATE \textcolor{blue}{Outer loop update}: $x_{t+1}^k = \sum_{j \in \cN_k}w_{k,j}x_t^{j} - \alpha_t \nabla_x  f(x_t^k,y_t^k;\zeta_t^k)$.
	\STATE 
	%Query the $\cS\cO$ at $(x_t^k,y_t^k)$ to obtain $\nabla_y g^k(x_t^k,y_t^k;\xi_t^k)$ and $\nabla_{xy}^2 g^k(x_t^k,y_t^k;\xi_t^k)$.
	%Query the $\cS\cO$ at $(x_t^k,y_t^k)$ to obtain stochastic gradient $\nabla_y g^k (x_t^k,y_t^k;\xi_t^k )$. 
 \ENDFOR
 \ENSURE $ \bar x_{T} = \frac{1}{K} \sum_{k\in \cK} x_T^k$.
\end{algorithmic}
\end{algorithm}

\subsection{Distributed policy evaluation for reinforcement learning} \label{app:MDP}

\textbf{Simulation environment:} In our experiments, for each state $s \in \cS$, we generate its feature $\phi_s \sim \text{Unif}[0,1]^m$;
we uniformly generate the transition probabilities $p_{s,s'}$ and standardize them such that $\sum_{s' \in \cS} p_{s,s'} = 1$; we sample the mean of rewards $\bar r_{s,s'}^k \sim \text{Unif}[0,1]$ for all $s \in \cS$ and each agent $k \in [K]$. We set the regularizer parameter $\lambda = 1$ and .

In each simulation, we set $|\cS| = 100$ and update the solution $(x^k,y^k)$ for each agent in a parallel manner as follows:
At iteration $t$, for each state $s \in \cS$, we simulate a random transition to another state $s' \in \cS$ using the transition probability $p_{s,s'}$'s, generate a random reward $r_{s,s'}^k \sim \cN(\bar r_{s,s'}^k, 1)$, and update $x_t^k$ using step-sizes $\alpha_t = \min \{ 0.01,\frac{2}{\lambda t} \}$ and $\beta_t = \gamma_t = \min \{ 0.5, \frac{50}{t}\}$.

\begin{algorithm}[t!]
\caption{Decentralized Stochastic Gradient Desecent}\label{alg:DSGD}
\begin{algorithmic}[1]
\REQUIRE Step-sizes $\{ \alpha_t \}$, weights  $ \{ \eta_{t,i}\}$, number of total iterations $T$.
\\
$x_0^k = \bf{0}$, $y_0^k = \bf{0}$
\FOR{$t=0, 1, \cdots, T-1$}
	\STATE 	\textcolor{blue}{Inner value update}:   Set $\tilde y_{t,0}^k = 0$.
	\FOR{ $i = 0,1,\cdots, t-1$}
	\FOR{$k = 1,\cdots, K$}
		\STATE \textcolor{blue}{Local sampling}: Query $\cS\cO$ at $x_t^k$ to obtain $  g^k(x_t^k;\xi_{t,i}^k)  $.
	\STATE \textcolor{blue}{Estimate:} $ \tilde y_{t,i+1}^k = (1-\eta_{t,i})\sum_{j \in \cN_k}w_{k,j}\tilde y_{t,i}^{j}  +
 \eta_{t,i} g^k(x_t^k;\xi_{t,i}^k) $.
 	\ENDFOR
 	\ENDFOR 
 	Set $y_{t}^k = \tilde y_{t,t}^k$. 
	\STATE \textcolor{blue}{Outer loop update}: $x_{t+1}^k = \sum_{j \in \cN_k}w_{k,j}x_t^{j} - \alpha_t \nabla g(x_t^k;\xi_t^k) \nabla  f(y_t^k;\zeta_t^k)$.
	\STATE 
	%Query the $\cS\cO$ at $(x_t^k,y_t^k)$ to obtain $\nabla_y g^k(x_t^k,y_t^k;\xi_t^k)$ and $\nabla_{xy}^2 g^k(x_t^k,y_t^k;\xi_t^k)$.
	%Query the $\cS\cO$ at $(x_t^k,y_t^k)$ to obtain stochastic gradient $\nabla_y g^k (x_t^k,y_t^k;\xi_t^k )$. 
 \ENDFOR
 \ENSURE $ \bar x_{T} = \frac{1}{K} \sum_{k\in \cK} x_T^k$.
\end{algorithmic}
\end{algorithm}

\begin{figure}[t]
\begin{minipage}{0.32\textwidth}
    \centering
   \includegraphics[width=1\linewidth]{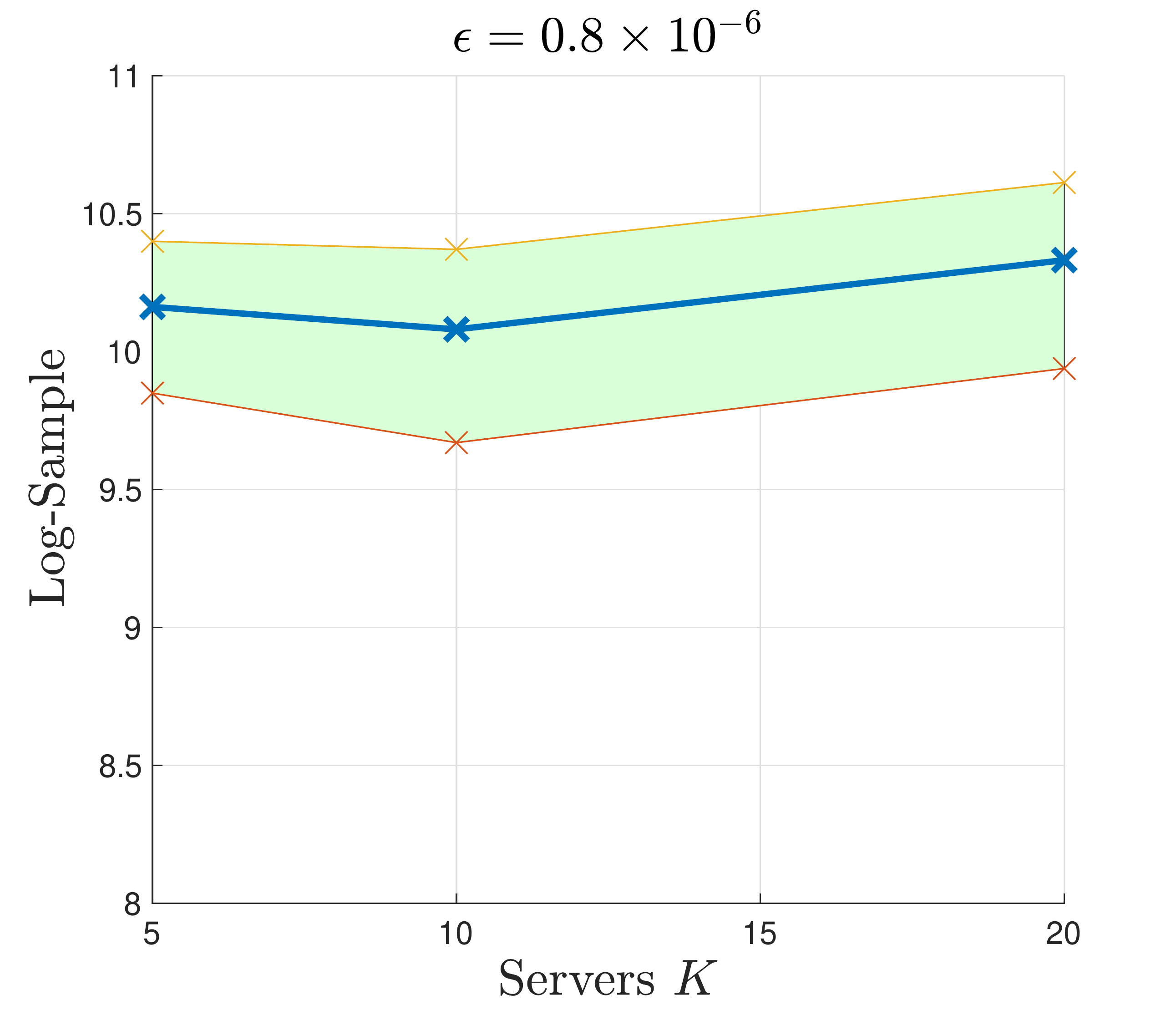}
        (a)
        \end{minipage}  
        \begin{minipage}{0.32\textwidth}
        \centering
   \includegraphics[width=1\linewidth]{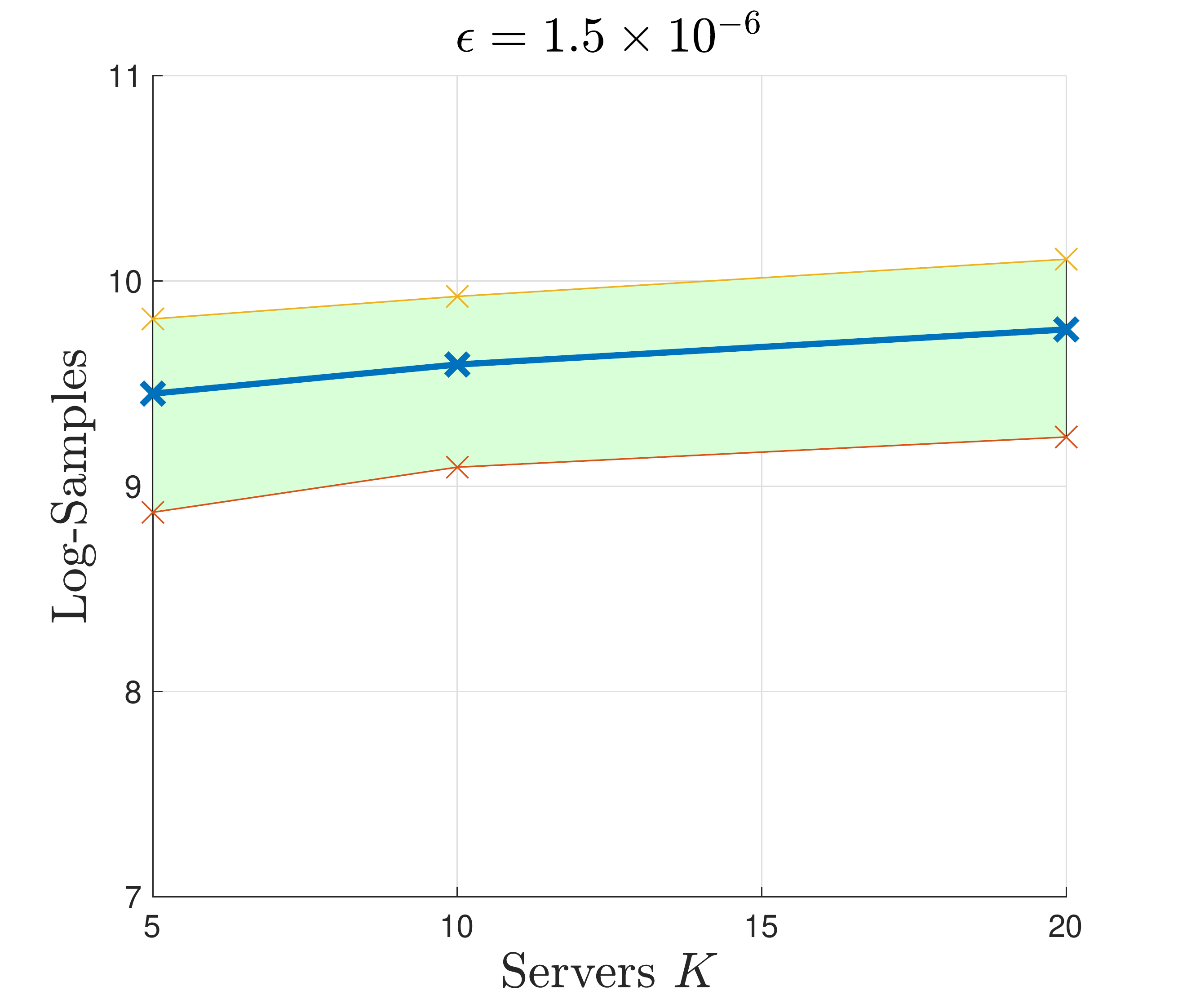}
   (b)
   \end{minipage}
 \begin{minipage}{0.32\textwidth}
        \centering
   \includegraphics[width=1\linewidth]{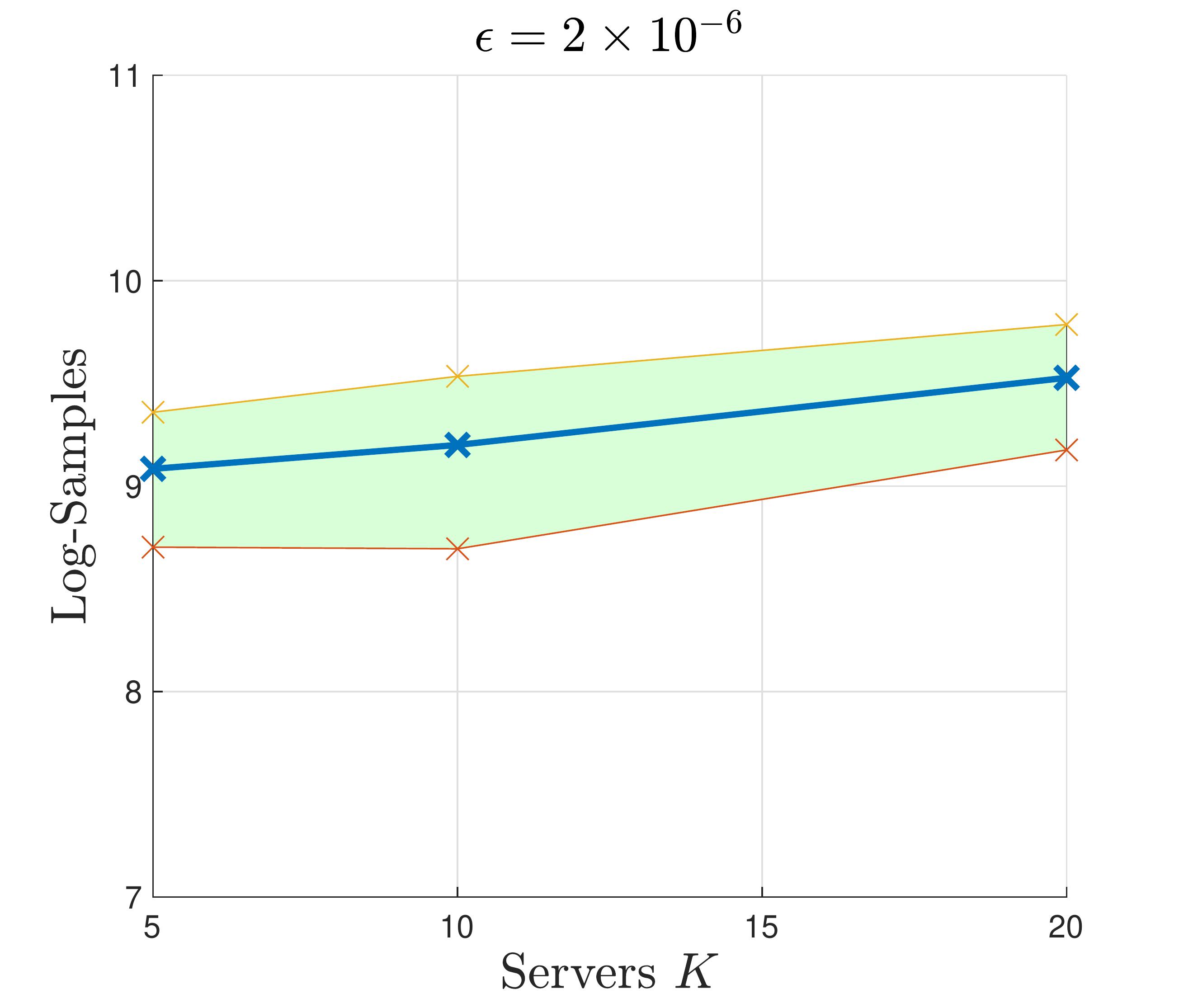}
   (c)
   \end{minipage}
    \caption{75\% confidence region of log- total samples for achieving $\|\bar x_t - x^* \|^2 \leq \epsilon$ with varying network sizes $K = 5,10,20$ for $\epsilon = 0.8\times 10^{-6}, 1.5\times 10^{-6}, 2 \times 10^{-6}$. 
    %All figures are generated through 10 independent simulations. 
    }
    %\mw{log-sample confusing. total or per-agent?}}
    \label{fig:MDP2}
\end{figure}

We provide the details of the baseline algorithm DSGD in Algorithm~\ref{alg:DSGD}.

To further study the linear speedup effect under various accuracy levels, we  compute the total generated samples for finding an $\epsilon$-optimal solution $\|\bar x_t - x^* \|^2 \leq \epsilon $ and plot the 75\% confidence region of log-sample against number of agents $K = 5,10,20$ for various $\epsilon$'s  in Figure~\ref{fig:MDP2}. Similar as in Figure~\ref{fig:1}, we observe for all accuracy levels $\epsilon = 0.8\times 10^{-6}, 1.5\times 10^{-6}, 2 \times 10^{-6}$, the required samples for finding an $\epsilon$-optimal solution by $K$ agents are roughly the same,
 further demonstrating the linear speedup effect.

\end{document}